\documentclass[5p,authoryear]{elsarticle}
\usepackage{graphicx}
\usepackage{amsmath, amssymb}

\usepackage{caption, subcaption}
\usepackage{float}
\usepackage{multirow}
\usepackage[colorlinks=true,linkcolor=black,citecolor=black,urlcolor=black]{hyperref}
\usepackage{pgf-pie}
\begin{document}
\begin{frontmatter}
%%%%%%%%%%%%%%%%%%%%%%%%%%
\title{A Comprehensive Survey for Hyperspectral Image Classification: The Evolution from Conventional to Transformers and Mamba Models}
%%%%%%%%%%%%%%%%%%%%%%%%%%
\author[1]{Muhammad Ahmad}
%\ead{dr.ahmad@nu.edu.pk}
\author[2]{Salvatore Distefano}
%\ead{sdistefano@unime.it}
\author[3]{Adil Mehmood Khan}
%\ead{a.m.khan@hull.ac.uk}
\author[4]{Manuel Mazzara}
%\ead{m.mazzara@innopolis.ru}
\author[5]{Chenyu Li}
%\ead{lichenyu@seu.edu.cn}
\author[6]{Hao Li}
%\ead{hao\_bgd.li@tum.de}
\author[7]{Jagannath Aryal}
%\ead{jagannath.aryal@unimelb.edu}
\author[8]{Yao Ding}
%\ead{yuchengguo@cardc.cn}
\author[9]{Gemine Vivone}
%\ead{gemine.vivone@imaa.cnr.it}
\author[5,10]{Danfeng Hong\corref{cor}}
\ead{hongdf@aircas.ac.cn}

\address[1]{Department of Computer Science, National University of Computer and Emerging Sciences, Islamabad, Pakistan. dr.ahmad@nu.edu.pk}
\address[2]{Dipartimento di Matematica e Informatica-MIFT, University of Messina, 98121 Messina, Italy. sdistefano@unime.it}
\address[3]{School of Computer Science, University of Hull, Hull HU6 7RX, UK. a.m.khan@hull.ac.uk}
\address[4]{Institute of Software Development and Engineering, Innopolis University, 420500 Innopolis, Russia. m.mazzara@innopolis.ru}
\address[5]{Aerospace Information Research Institute, Chinese Academy of Sciences, Beijing, 100094, China. lichenyu@seu.edu.cn}
\address[6]{Big Geospatial Data Management, Technical University of Munich, Munich 85521, Germany. hao\_bgd.li@tum.de}
\address[7]{Department of Infrastructure Engineering, University of Melbourne, Parkville, VIC, Australia. jagannath.aryal@unimelb.edu}
\address[8]{Intelligent Control Laboratory, PLA Rocket Force University of Engineering, 710025 Xi'an, China. yuchengguo@cardc.cn}
\address[9]{Institute of Methodologies for Environmental Analysis, National Research Council, 85050 Tito, Italy. gemine.vivone@imaa.cnr.it}
\address[10]{School of Electronic, Electrical and Communication Engineering, University of Chinese Academy of Sciences, 100049 Beijing, China. hongdf@aircas.ac.cn}
\cortext[cor]{Corresponding author}
%%%%%%%%%%%%%%%%%%%%%%%%%%
\begin{abstract}
Hyperspectral Image Classification (HSC) presents significant challenges owing to the high dimensionality and intricate nature of Hyperspectral (HS) data. While traditional Machine Learning (TML) approaches have demonstrated effectiveness, they often encounter substantial obstacles in real-world applications, including the variability of optimal feature sets, subjectivity in human-driven design, inherent biases, and methodological limitations. Specifically, TML suffers from the curse of dimensionality, difficulties in feature selection and extraction, insufficient consideration of spatial information, limited robustness against noise, scalability issues, and inadequate adaptability to complex data distributions. In recent years, Deep Learning (DL) techniques have emerged as robust solutions to address these challenges. This survey offers a comprehensive overview of current trends and future prospects in HSC, emphasizing advancements from DL models to the increasing adoption of Transformer and Mamba Model architectures. We systematically review key concepts, methodologies, and state-of-the-art approaches in DL for HSC. Furthermore, we investigate the potential of Transformer-based models and the Mamba Model in HSC, detailing their advantages and challenges. Emerging trends in HSC are explored, including in-depth discussions on Explainable AI and Interoperability concepts, alongside Diffusion Models for image denoising, feature extraction, and image fusion. To substantiate the efficacy of various conventional DL models, Transformers, and the Mamba Model, comprehensive experimental results were conducted on three HS datasets, yielding notable accuracies: 99.94\% on Pavia University, 99.41\% on Indian Pines, and 99.97\% on the University of Houston dataset. Additionally, we identify several open challenges and pertinent research questions in the field of HSC. Finally, we outline future research directions and potential applications aimed at enhancing the accuracy and efficiency of HSC. The Source code is available at \url{https://github.com/mahmad00/Conventional-to-Transformer-for-Hyperspectral-Image-Classification-Survey-2024}.
\end{abstract}
%%%%%%%%%%%%%%%%%%%%%%%%%%
%%%%%%%%%%%%%%%%%%%%%%%%%%
\end{frontmatter}
%%%%%%%%%%%%%%%%%%%%%%%%%%
\section{\textbf{Introduction}}
\label{Intr}

Hyperspectral (HS) Sensors capture detailed spectral information across a broad range of electromagnetic wavelengths \citep{ahmad2021hyperspectral}. Unlike traditional methods, Hyperspectral Images (HSIs) provide a continuous spectrum through numerous narrow bands, enabling precise material characterization and valuable Earth surface information extraction \citep{hong2024spectralgpt, li2024learning}. Its significance in remote sensing lies in overcoming the limitations of multispectral imaging, allowing discrimination of subtle differences in material properties, and facilitating accurate land cover classification \citep{10415455, neupane2021deep}.

%%%%%%%%%%%%%%%%%%%%%%%%%%
\begin{figure*}[!t]
    \centering
    \includegraphics[width=0.99\textwidth]{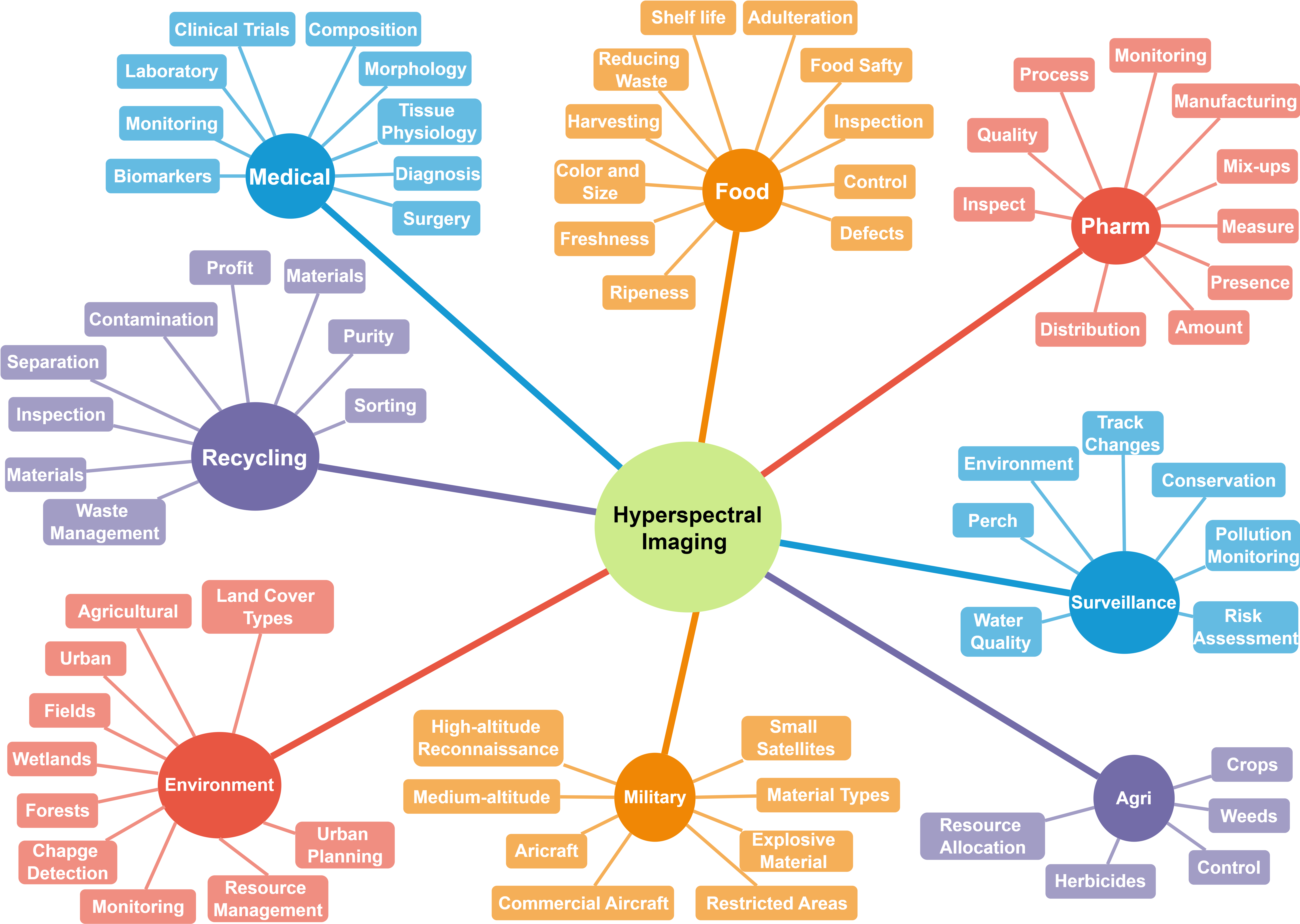}
    \caption{\textbf{Various Real-World Applications of Hyperspectral Imaging (HSI):} \textbf{Agriculture:} HSI is utilized to monitor crop health through stress detection and nutrient assessment. It can identify diseases at an early stage by analyzing spectral changes in plant reflectance. Additionally, HSI helps assess soil properties, including moisture content and organic matter levels, which are crucial for optimizing crop yields. \textbf{Environmental Monitoring:} In this domain, HSI plays a vital role in tracking pollution levels by analyzing spectral signatures of pollutants in air and water. It facilitates ecosystem mapping by providing detailed information about vegetation types and health. HSI also effectively assesses water quality by detecting harmful algal blooms and monitoring sediment levels. \textbf{Mineralogy:} HSI is employed in mineral identification by differentiating between mineral types based on spectral characteristics. This capability is invaluable in mining exploration, as it aids in locating valuable mineral deposits and optimizing extraction processes. \textbf{Healthcare:} In the medical field, HSI assists in disease diagnosis by identifying abnormal tissue characteristics. It is also utilized in monitoring tissue health, particularly in assessing the efficacy of treatments through non-invasive spectral analysis. \textbf{Surveillance and Security:} HSI enhances surveillance capabilities by enabling target detection through the analysis of specific material signatures. It is also used in material identification, allowing for better security measures by distinguishing between benign and hazardous materials.}
    \label{Fig1}
\end{figure*}
%%%%%%%%%%%%%%%%%%%%%%%%%%

HSI's key advantage lies in its distinguishing materials with similar visual appearances (e.g., color, texture, shape, structure, etc.) but distinct spectral properties \citep{10414144}. This specificity is crucial in various applications, such as agriculture \citep{SHUAI2024108577, deng2024rustqnet}, forestry \citep{MA2024111608}, urban planning \citep{KHORAT2024113751, JANECKI2024111302}, environmental monitoring \citep{SONG2024100440, wu2019fourier}, mineral exploration \citep{hong2018augmented}, disaster management, smart cities \cite{aryal2015smart}, bloodstain identification \citep{butt2022fast, zulfiqar2021hyperspectral}, meat processing \citep{ayaz2020myoglobin, ayaz2020hyperspectral}, contamination detection in red chili \citep{khan2021hyperspectral, khan2020hyperspectral}, and bakery products \citep{saleem2020prediction}, further details are presented in Figure \ref{Fig1}.

In agriculture \citep{SHUAI2024108577}, HSI is used to monitor crop health, detect diseases, and estimate vegetation properties like chlorophyll content, optimizing irrigation and pest control for enhanced crop management. In forestry, HSI is used to assess forest health and species composition, and detects pests and diseases early, aiding conservation efforts. In geological studies \citep{KHORAT2024113751, JANECKI2024111302, SONG2024100440} and mineral exploration \citep{10413643, 10399798}, HSI used to identify mineral deposits through unique spectral signatures, supporting resource exploration, mining, and environmental impact assessments. Additionally, HSI is used to assist urban planning by identifying land cover types, analyzing urban heat islands, monitoring urban expansion, and assessing environmental quality for sustainable development and infrastructure planning \citep{Tian21, Kim14}. 

In short, HSI provides detailed spectral information, enhancing understanding of the Earth's surface and supporting land cover classification, environmental assessment, change monitoring, and decision-making \citep{ahmad2020fast}. As technology advances and HS sensors become more accessible, HSI will play a vital role in addressing environmental challenges and facilitating sustainable resource management. This survey focuses on the field of Hyperspectral Image Classification (HSC), which has seen significant advancements, especially with the rise of Deep Learning (DL) models \citep{10433668, 10423094}. The survey aims to comprehensively overview commonly used DL techniques for HSC. It will address primary challenges in HSC, not effectively tackled by Traditional Machine Learning (TML) methods, and emphasize how DL improves HSC performance.

%%%%%%%%%%%%%%%%%%%%%%%%%%
\subsection{\textbf{Limitations of Existing Surveys}}

Existing surveys on HSC offer comprehensive overviews of the field, highlighting various methodologies, techniques, and challenges. These surveys provide insights into the evolution of classification algorithms, including traditional methods and recent DL approaches. They also analyze the effectiveness of different feature extraction and selection techniques, as well as the impact of spatial-spectral fusion methods on classification performance. Additionally, these surveys often explore applications of HSC across diverse domains such as agriculture, environmental monitoring, and remote sensing. Overall, existing surveys serve as valuable resources for researchers and practitioners seeking a deeper understanding of HSC techniques and their applications. 

However, existing surveys \citep{GUERRI2024100062, ahmad2021hyperspectral} on HSC have some shortcomings, including \textbf{Limited Coverage:} Some surveys may focus only on specific aspects of HSC, such as traditional methods or DL approaches, potentially overlooking other important aspects or emerging trends in the field. \textbf{Outdated Information:} Surveys published several years ago may not capture the latest advancements and breakthroughs in HSC techniques, rendering them less relevant for current research and applications. \textbf{Lack of Depth:} Some surveys may provide superficial coverage of the subject matter, lacking in-depth analysis or critical evaluation of different methodologies, techniques, and challenges in HSC. \textbf{Narrow Perspective:} Certain surveys may adopt a narrow perspective, focusing primarily on a single application domain or a particular subset of HSC techniques, thereby limiting their applicability to broader contexts. Addressing these shortcomings could enhance the effectiveness and utility of surveys on HSC, ensuring they provide comprehensive, up-to-date, and accessible resources for researchers and practitioners in the field.

The current survey on HSC offers several advantages compared to existing surveys: \textbf{Comprehensive Coverage:} The survey provides a comprehensive overview of current trends and future prospects in HSC, covering both traditional DL models and the emerging use of transformer-based models. This broad coverage ensures that readers gain insights into the full spectrum of approaches in the field. \textbf{Focus on Emerging Technologies:} By delving into the potential of Transformer-based models, the survey stays up-to-date with the latest advancements in DL for HSC. This focus on emerging technologies ensures that readers are informed about cutting-edge developments and potential future directions in the field. \textbf{Thorough Discussions on Key Concepts:} The survey includes thorough discussions on key concepts, methodologies, and state-of-the-art approaches in DL for HSC (for instance, RML, Convolutional Neural Network (CNN), and its variants, Deep Belief Network (DBN), Recurrent Neural Network (RNN), Autoencoders (AEs), Transformers, Explainable AI and Interoperability, and Diffusion Models). This depth of analysis helps readers understand the underlying principles and techniques used in the field. \textbf{Inclusion of Experimental Results:} The survey presents comprehensive experimental results conducted using three Hyperspectral (HS) datasets, verifying the efficacy of various DL models and transformers. This empirical evidence adds credibility to the survey's findings and helps readers assess the practical implications of different approaches. \textbf{Open Challenges and Future Directions:} The survey addresses several open challenges and research questions pertinent to HSC, providing valuable insights into areas where further research is needed. Additionally, outlining future research directions and potential applications helps guide researchers toward promising avenues for exploration. \textbf{Accessibility of Source Code:} The availability of the survey's source code on \href{https://github.com/mahmad00/Conventional-to-Transformer-for-Hyperspectral-Image-Classification-Survey-2024}{GitHub} enhances transparency and reproducibility, allowing readers to access and validate the experimental results presented in the survey.

Overall, the current survey stands out for its comprehensive coverage, focus on emerging technologies, thorough discussions, inclusion of experimental results, identification of open challenges, and accessibility of source code, making it a valuable resource for researchers and practitioners in the field of HSC.

%%%%%%%%%%%%%%%%%%%%%%%%%%
\subsection{\textbf{Significance of Accurate Classification}}

Accurate classification in HSI analysis is crucial due to the wealth of complex information within the data's numerous spectral bands. HSIs provide unique details about observed objects, making precise classification essential for diverse applications. The significance of accurate classification lies in extracting meaningful information such as land cover types and land-use patterns \citep{XU2024102460}. HSC techniques identify and map different land cover classes, aiding land management, resource planning, and environmental assessments \citep{AHMAD2021166267, AHMAD201786}. This accuracy is crucial for monitoring changes in land cover, providing valuable insights into deforestation, urban expansion, and ecosystem degradation, thereby supporting conservation efforts and sustainable land-use practices \citep{PATEL2024101788}.

Accurate classification is crucial for environmental monitoring and assessment \citep{JANECKI2024111302}. HS imagery detects factors like water quality, pollution sources, and ecosystem health, allowing identification of areas of concern, tracking changes, and evaluating environmental management strategies. This information is vital for ecosystem health, biodiversity conservation, and mitigating environmental risks. In agriculture, precise HSC supports crop monitoring, disease detection, and yield estimation \citep{SHUAI2024108577}. It identifies stressed vegetation, enabling targeted interventions like irrigation or pesticide application. HSC also aids precision agriculture by providing information on soil composition, nutrient levels, and plant health, optimizing resource allocation for improved productivity.

In mineral exploration, precise classification of HS data identifies and maps mineral deposits and geological formations \citep{10413643, 10399798}. Distinct spectral signatures of minerals aid prospecting and resource estimation, crucial for sustainable resource management, reducing exploration costs, and minimizing environmental impacts. Urban planning and infrastructure development benefit from accurate HSC \citep{KHORAT2024113751}. Accurate classification of land cover types and urban features enables assessing urban growth patterns, monitoring changes, and planning for future development. It supports urban heat island analysis, transportation planning, and identifying green spaces, contributing to sustainable and livable cities.

Therefore, accurate HSC is highly significant, improving our understanding of the Earth's surface, supporting informed decision-making, and enabling sustainable management practices. With advancing HSI technology, accurate classification techniques will be crucial in addressing complex challenges and fostering sustainable development.

%%%%%%%%%%%%%%%%%%%%%%%%%%
\subsection{\textbf{Brief Introduction to Deep Learning}}

DL, for instance, CNNs, has revolutionized HSC, outperforming conventional methods by leveraging neural networks and attention mechanisms \citep{ahmad2022hybrid, 10379821,ghorbanzadeh2019evaluation}. CNNs excel in learning hierarchical representations and capturing spatial and spectral dependencies within HS data \citep{ahmad2021artifacts}. Their ability to automatically extract discriminative features makes them well-suited for accurate classification in HSI \citep{ran2023deep, ahmad2022disjoint}. RNNs, designed for sequential data, show promise in HSC for temporal analysis \citep{10416691}. Exploiting the temporal dimension, RNNs capture contextual information for tasks like tracking land cover changes or monitoring dynamic environmental processes. Autoencoders (AEs), a neural network type, find application in HSC \citep{10403871}. Comprising an encoder and decoder network, AEs compress input data into a low-dimensional representation (latent space) and attempt to reconstruct the original input \citep{ahmad2020spatial}. In HSI, AEs are used for unsupervised feature learning and dimensionality reduction \citep{ahmad2019segmented}. Training AEs on unlabeled HS data enables effective feature extraction and classification by capturing the underlying data structure \citep{ahmad2019multi}.

More recently, Transformers have gained significant attention in various domains, including natural language processing and computer vision \citep{10399798, ran2023deep, 10223236, 10415455}. Transformers are based on a self-attention mechanism that allows the model to capture dependencies between different positions within a sequence. In the context of HSC, Transformers have shown promise in learning long-range dependencies and capturing non-local spectral relationships. By modeling the interactions between different spectral bands, transformers can effectively exploit the rich spectral information in HS data. This has led to improved classification performance and the ability to capture complex spectral patterns. Transformers can handle variable-length input sequences, making them suitable for HSIs of different spatial dimensions. Additionally, transformers offer interpretability, as the attention mechanism allows for visualizing the importance of different spectral bands in the classification process. This interpretability can be valuable for understanding the reasoning behind the model's predictions and enhancing domain knowledge. In short, DL models, including CNNs, RNNs, AEs, and Transformers, demonstrate substantial potential in HSC. Leveraging their capacity to learn intricate representations, capture spatial and spectral dependencies, and exploit temporal information, these models contribute to significant advancements in accurate HSC, enhancing our understanding of the Earth's surface across diverse applications.

%%%%%%%%%%%%%%%%%%%%%%%%%%
\subsection{\textbf{Structure of the Survey}}

The following sections are organized as follows: Section \ref{HSI} provides an in-depth description of the HS dataset, covering spectral, spatial, and spectral-spatial representation. In Section \ref{LC}, various learning mechanisms for HSI processing are outlined. Section \ref{Trad} examines TML for HSC along with their limitations. Section \ref{ADL} explores the advantages of employing DL for HSC. Section \ref{CNN} elaborates on CNN models for spectral, spatial, and spectral-spatial models along with Graph CNN models. Section \ref{DBN} outlines DBN for HSC and potential research directions. Section \ref{RNN} discusses RNNs for HSC and potential research directions. Section \ref{AEs} details AEs for HSC and potential research directions. Section \ref{LTDL} draws limitations of traditional DL approaches. In Section \ref{Trans}, an extensive discussion on state-of-the-art deep models, specifically Transformers, includes a comparison with traditional deep models, benefits, and challenges for HSI analysis. Subsequently, Section \ref{Trans} explores emerging trends and advancements in state-of-the-art models for HSI analysis. Section \ref{EXI} discusses Explainable AI and Interoperability in HSC. Section \ref{DFs} briefly describes the Diffusion models employed for HS imaging. Section \ref{SSM} describes State Space Modeling (SSM)--Mamba for HSC, while Section \ref{RQs} outlines challenges and research questions. In Section \ref{Res}, detailed experimental results for traditional DL models and Transformer-based deep models are presented. Finally, Section \ref{Con} concludes the survey and suggests immediate research directions.

%%%%%%%%%%%%%%%%%%%%%%%%%%
\section{\textbf{Hyperspectral Data Representation}}
\label{HSI}

\begin{figure}[!hbt]
    \centering
    \includegraphics[width=0.48\textwidth]{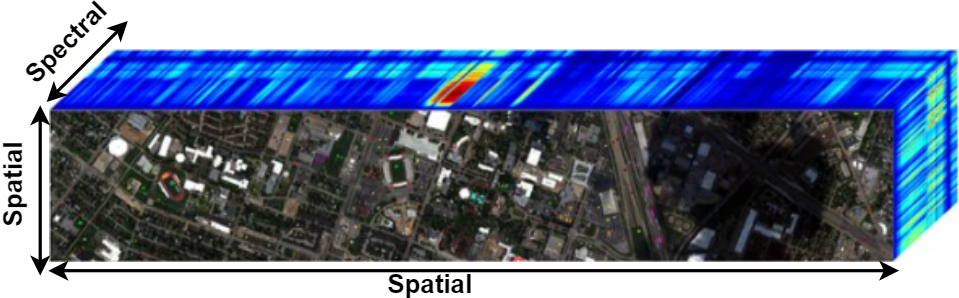}
    \caption{University of Houston: Hyperspectral Image Cube representation.}
    \label{Fig3}
\end{figure}

HS data is typically a 3D hypercube, denoted as $\textbf{X} \in \mathcal{R}^{B \times (N \times M)}$, containing spectral and spatial information of a sample \citep{ahmad2023sharpend}. Here, $B$ is the number of spectral bands, and $N$ and $M$ represent the spatial components (width and height). Figure \ref{Fig3} illustrates an HSI cube from the University of Houston dataset.

%%%%%%%%%%%%%%%%%%%%%%%%%%
%%%%%%%%%%%%%%%%%%%%%%%%%%

In spectral representation, each pixel vector, denoted as $x_i \in \mathcal{R}^B$, is processed independently based on its spectral signature. The value of $B$ represents the spectral channels, which can be the actual number or a reduced set obtained through dimensionality reduction (DR) techniques. A lower-dimensional representation is preferred to reduce redundancy and enhance class separability without significant information loss \citep{10411886}. Unsupervised DR techniques, like Principal Component Analysis (PCA) \citep{10414262}, and Locally Linear Embedding \citep{hong2020joint}, transform HSI data without using class labels. Supervised DR methods, such as linear discriminant analysis (LDA), local Fisher discriminant analysis (LFDA), local discriminant embedding (LDE), and nonparametric weighted feature extraction (NWFE) \citep{4801616}, utilize labeled samples to improve class separability. LDA and LFDA, for instance, enhance class separability by maximizing the distance between classes while minimizing within-class distance. However, spectral mixing challenges differentiation between classes solely based on spectral reflectance. This underscores the need for advanced techniques beyond spectral information for accurate HSI class differentiation.

To overcome spectral representation limitations, spatial information of pixels can be utilized. In this approach, pixels in each spectral band are represented as a matrix, denoted as $x_i \in \mathcal{R}^{N \times M}$. Spatial representation considers neighboring pixels due to high spatial correlation, enhancing class coherence. Neighbors are identified using kernels or pixel-centric windows \citep{10401954}. Various methods extract spatial information from the HSI cube, including morphological profiles (MPs) \citep{10411886}, Gabor filters for texture features \citep{10416200}, gray-level co-occurrence matrix (GLCM) \citep{10398212}, local binary patterns (LBP) \citep{hong2015novel}, and Deep Neural Network (DNN)-based methods. MPs capture geometrical characteristics, with extensions like extended morphological profiles (EMPs) \citep{10411886} and multiple-structure-element morphological profiles \citep{hong2020invariant}. Texture extraction methods like Gabor filters capture textural details, while LBP \citep{10380319} provides rotation-invariant representations. GLCM \citep{10398212, 6410025} determines spatial variability based on relative pixel positions. DNNs can extract spatial information by treating each pixel as an image patch, learning spatial features. Combining multiple methods, as seen in studies like Zhang et al. \citep{zhang2018spatial}, enhances spatial information extraction. In this study, Gabor filters and differential MPs were combined for an RNN-based HSC framework, showcasing the potential of integrating techniques for leveraging spatial information.

Spatial-spectral approaches in HSI process pixel vectors by incorporating both spectral features and spatial-contextual information. Two common strategies exist for the simultaneous use of spectral and spatial representations. The first strategy concatenates spatial details with the spectral vector, demonstrated in studies by Chen et al. \citep{chen2014deep, chen2015spectral}, merging both into a single feature representation. The second strategy processes the three-dimensional HSI cube, preserving its structure and contextual information. Paoletti et al. \citep{paoletti2018deepa} showcase this approach, maintaining the HSI's three-dimensional nature and utilizing DL techniques to extract features capturing both spectral and spatial characteristics. Both strategies aim to enhance HSI analysis and classification tasks by leveraging the complementary nature of spectral and spatial information. While many DNN models focus on spectral representation \citep{7446269}, recognizing its limitations, recent efforts aim to integrate spatial information into the classification process \citep{7446269}. Joint exploitation of spectral and spatial features has gained interest, leading to improved accuracy \citep{6945376}.

%%%%%%%%%%%%%%%%%%%%%%%%%%
\section{\textbf{Learning Categories}}
\label{LC}

\begin{figure}[!hbt]
    \centering
    \includegraphics[width=0.48\textwidth]{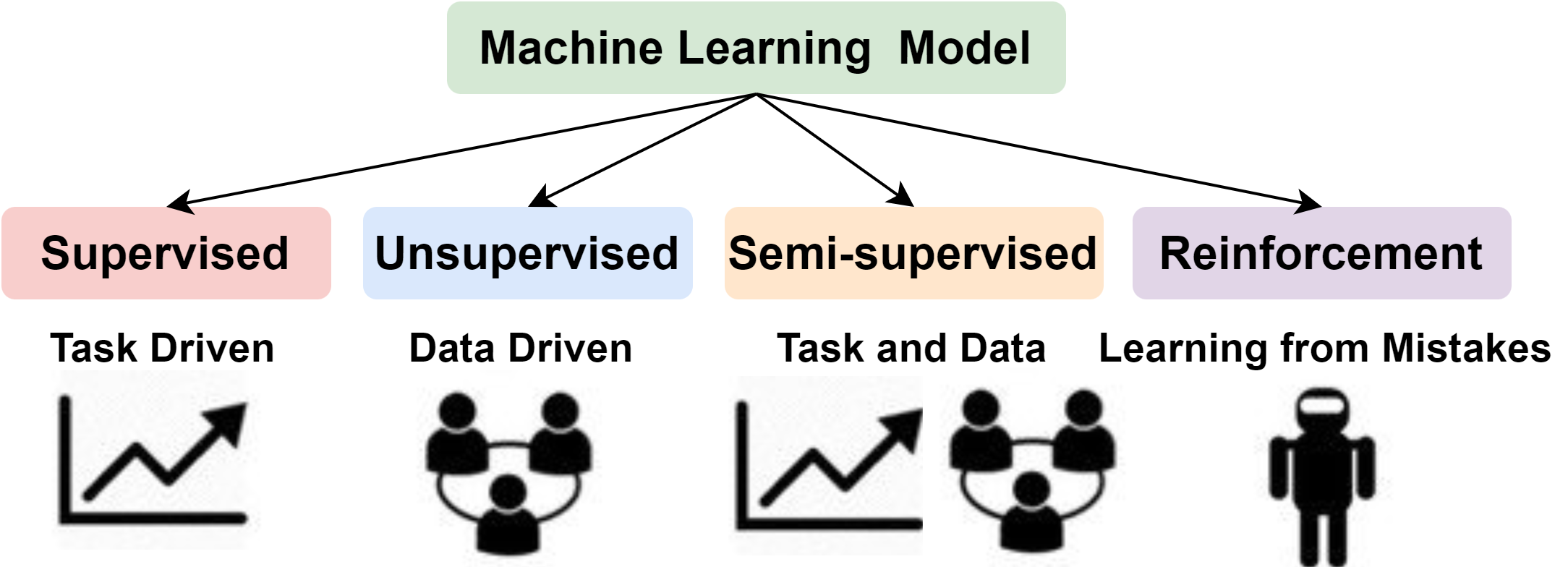}
    \caption{Major Learning Categories.}
    \label{Fig4}
\end{figure}

Machine learning models for HSC can employ different learning strategies, four of which are presented in Figure \ref{Fig4}. 

%%%%%%%%%%%%%%%%%%%%%%%%%%
%%%%%%%%%%%%%%%%%%%%%%%%%%
\subsection{Supervised Learning}

In supervised learning, models are trained on labeled data, associating each example with a target output. The model adjusts its parameters iteratively during training to improve accuracy in predicting outputs. Effective training requires a substantial amount of labeled data, making these models suitable for scenarios with abundant labeled data. DNNs excel in learning complex patterns and making accurate predictions in such scenarios. \textbf{Unsupervised Learning} Unsupervised learning involves training models on unlabeled data to uncover patterns without predefined target labels. Measuring accuracy can be challenging, but it is useful when learning intrinsic structures of datasets with limited labeled training data. Techniques like PCA learn low-dimensional representations by identifying directions of maximum variance. K-means clustering groups data into clusters, revealing patterns or similarities among instances \citep{9945998}. 

\subsection{Unsupervised Learning} 

Unsupervised learning in HSI is a powerful approach for analyzing high-dimensional data without the need for labeled training samples. This method leverages the rich spectral information inherent in HSIs to identify intrinsic patterns and structures within the data. Techniques such as clustering, dimensionality reduction, and feature extraction are commonly employed to uncover underlying data distributions and group similar spectral signatures. For instance, algorithms like k-means clustering or hierarchical clustering can be used to segment regions of interest based on their spectral characteristics, facilitating tasks such as anomaly detection and material identification. Additionally, advanced methods like self-organizing maps, autoencoders, and generative models are increasingly being explored to enhance the interpretability of HSI data, enabling more robust insights into land cover classification and environmental monitoring applications. The ability to extract meaningful information from unlabeled data makes unsupervised learning an essential tool in the HSI domain, particularly in scenarios where obtaining labeled datasets is challenging or costly \citep{10415455, khan2021hyperspectral}.

\subsection{Semi-Supervised Learning}

Semi-supervised learning combines elements of both supervised and unsupervised learning. It utilizes a limited amount of labeled data along with a larger set of unlabeled data for training. The model learns from the labeled data to make predictions on both labeled and unlabeled instances, leveraging the additional unlabeled data for improved performance \citep{10409242, 6423895}. 

\subsection{Reinforcement Learning}

An agent interacts with an environment and learns to take actions to maximize cumulative rewards. The agent receives feedback in the form of rewards or penalties based on its actions, allowing it to learn optimal decision-making policies through trial and error. 

\subsection{Transfer Learning} 

Transfer learning involves leveraging knowledge or representations learned from one task or domain and applying them to a different but related task or domain. This approach can be beneficial when labeled data is scarce or when pre-trained models can provide useful feature representations \citep{9106753, 8913592}. 

\subsection{Online Learning}, 

Online learning also known as, incremental learning, involves updating the model continuously as new data becomes available. The model learns from each instance of data sequentially, adapting to changing patterns or concepts over time \citep{8088348, hong2017learning}. 

\subsection{Self-learning}

Self-learning for HSC leverages the model's ability to iteratively improve its performance without extensive labeled data. This approach utilizes techniques such as self-supervised learning, where the model learns from the inherent structure and properties of the HS data. By generating pseudo-labels and employing consistency regularization, the model refines its understanding of spectral-spatial patterns. Self-learning is particularly beneficial in HSI due to the high dimensionality and limited availability of labeled samples, enabling more efficient and accurate classification by maximizing the utility of available data \citep{10416200, 10443952, 10382626}.

%%%%%%%%%%%%%%%%%%%%%%%%%%
\section{\textbf{Brief Overview of Conventional Methods}}
\label{Trad}

The main goal of HSC is to assign distinct labels to pixel vectors in the HSI cube based on their spectral or spatial characteristics \citep{9307220}. Mathematically represented as $\textbf{X} = [x_1, x_2, x_3, \dots, x_B]^T \in \mathcal{R}^{B \times (N \times M)}$, where $B$ is the total number of spectral bands with $N \times M$ samples per band \cite{ahmad2019spatial}. These samples belong to classes $\textbf{Y}$, each $x_i = [x_{1,i},~x_{2,i},~x_{3,i}, \dots,x_{B,i}]^T$ represents the $i^{th}$ sample with a class label $y_i \in \mathcal{R}^Y$. The classification task is treated as an optimization problem, where a mapping function $f_c(.)$ operates on the input data $\textbf{X}$ \citep{ahmad2021artifacts}, aiming to minimize the disparity between the obtained output and actual labels \citep{paoletti2019deep} as shown in equation \ref{Eq1}.

\begin{equation}
\textbf{Y} = f_c(\textbf{X},\theta)
\label{Eq1}
\end{equation}
where the parameter $\theta$ adjusts transformations applied to the input data $\textbf{X}$ for the mapping function $f_c: \textbf{X} \to \textbf{Y}$. HSC research has seen a trend influenced by computer vision methodologies \citep{ahmad2021hyperspectral}. TML-based HSC relies on handcrafted features like shape, texture, color, spectral, and spatial details, and classification techniques. Common feature extraction methods include texture descriptors (LBPs, HOG \citep{7326462}, PHOG \citep{10043832}), SIFT \citep{8025601}, GIST \citep{8025601} and classifiers include Random Forests \citep{8318892}, kernel-based SVM \citep{8451145}, KNN \citep{5555996}, and ELM \citep{ahmad2019multi}.

Color histograms are simple and effective but lack spatial context, making them sensitive to illumination changes. HOG captures edge orientations effectively \citep{hong2016robust}. SIFT is robust but computationally intensive. GIST provides a global image description based on statistical properties like roughness and openness \citep{8025601}. Texture descriptors like LBPs are common for capturing texture around each pixel \citep{10137555}. Color histograms, GIST, and texture descriptors are global features, capturing statistical characteristics, while HOG and SIFT are local features describing geometric information \citep{8557124}. Combining these features is a common practice in HSC \citep{hong2019cospace}. LBPs are computationally efficient, involving simple binary operations on local pixel neighborhoods. In contrast, SIFT is more complex, requiring convolutions, gradient computations, and orientation assignments, making it computationally intensive \citep{nhat2019feature}.

Hand-crafted features, while effective, face challenges in real-world data due to varying optimal feature sets, subjectivity in human-driven design, biases, and limitations. Traditional HSC approaches encounter the curse of dimensionality, struggle with feature selection and extraction, lack spatial information consideration, exhibit limited robustness to noise, face scalability issues, and may not adapt well to complex data distributions. These challenges, including the curse of dimensionality, reliance on domain expertise, and limited spatial information, have prompted the exploration of DL-based methods, such as CNNs and RNNs, for HSC. DL techniques offer automated feature learning, potentially overcoming the limitations of traditional methods and improving classification accuracy by capturing intricate spatial-spectral relationships in HSIs \citep{hinton2006reducing, hu2015transferring}.

%%%%%%%%%%%%%%%%%%%%%%%%%%
\section{\textbf{Advantages of Deep Learning}}
\label{ADL}

\begin{figure}[!hbt]
    \centering
    \includegraphics[width=0.48\textwidth]{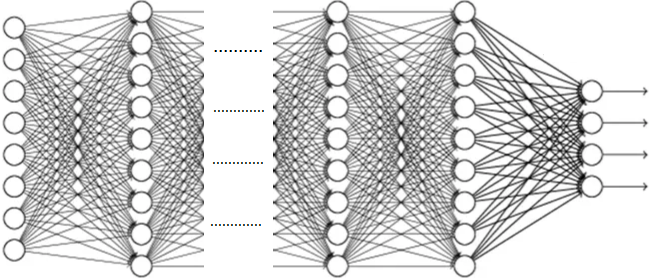}
    \caption{Example of Deep Artificial Neural Network.}
    \label{Fig6}
\end{figure}

Deep Learning (DL) architectures, exemplified in Figure \ref{Fig6}, can discern patterns and features from data without prior knowledge of its statistical distribution \citep{li2024interpretable}. Unlike conventional methods, DL models extract both linear and non-linear features without pre-specified information, making them suitable for handling HS data in spectral and spatial domains, individually or combined \citep{8694781, ahmad2021hyperspectral}. With flexible architectures and varying depths, DL models accommodate different layers, allowing adaptation to diverse machine learning strategies—supervised, semi-supervised, and unsupervised. This adaptability makes DL architectures optimized and tailored for specific classification tasks across various domains and applications \citep{8103149}.

%%%%%%%%%%%%%%%%%%%%%%%%%%
%%%%%%%%%%%%%%%%%%%%%%%%%%

DL enhances HSI data processing, leading to heightened classification accuracy and the ability to capture intricate relationships. DL models automatically learn and extract informative features from high-dimensional HSI data, addressing inherent complexity and variability \citep{chen2014convolutional}. Overall, DL architectures present a promising approach for HSC, learning from data without prior distribution knowledge, and offering architectural and learning strategy flexibility \citep{ahmad2021regularized}. These capabilities make DL models well-suited for overcoming challenges and complexities in HSI data analysis.

\textbf{Automatic feature learning:} DL models autonomously learn relevant features from data, creating hierarchical representations at various abstraction levels \citep{8482250}. Unlike traditional methods relying on hand-crafted features, DL adapts and extracts discriminative features from HSIs, reducing the need for manual engineering and potentially capturing complex spectral-spatial patterns. \textbf{Spatial spectral fusion:} DL models effectively capture spatial and spectral characteristics by integrating both types of information \citep{9207854}. CNNs exploit spatial structure through convolutional layers, discerning local patterns \citep{9078778}. Combining spectral and spatial details enhances accuracy, especially for classes with similar spectra but distinct spatial distributions.

\textbf{End-to-end learning:} DL models are trained end-to-end, optimizing the entire model, including feature extraction and classification, jointly \citep{8950204, 10282329}. This simultaneous learning of features and classifiers can lead to superior performance compared to separate stages in traditional approaches. \textbf{Adaptability to complex data distributions:} DL models, particularly CNNs, with their non-linear transformations, effectively capture complex data distributions and intricate relationships between features and classes \citep{8840914}. This adaptability enables better handling of variability and nonlinearity in HSIs.

\textbf{Transfer learning and pre-training:} DL models benefit from transfer learning, leveraging knowledge from one task or dataset for another \citep{5640675}. Pre-training on large-scale datasets, followed by fine-tuning on smaller HS datasets, enhances classification performance, especially with limited labeled data \citep{ahmad2022disjoint}. \textbf{Scalability:} DL models efficiently handle large-scale datasets, enabling quick classification for real-time applications. Advances in hardware and DL frameworks, like Graphics Processing Units (GPUs), accelerate training and inference processes, making DL applicable to larger and more complex HS datasets \citep{10250909, 10191985}. Despite these advantages, DL models for HSC face challenges, including the need for large labeled datasets, computational requirements, and potential overfitting. However, ongoing developments in DL techniques and increased HS dataset availability mitigate these challenges, establishing DL as a promising approach for accurate and automated HSC.

%%%%%%%%%%%%%%%%%%%%%%%%%%
\section{\textbf{Convolutional Neural Networks for HSC}}
\label{CNN}

The CNN architecture, inspired by Hubel and Wiesel's biological visual system \citep{hubel1962receptive} and Fukushima's Neocognitron model \citep{fukushima1980neocognitron}, comprises two main stages: Feature Extraction (FE) and classification. The FE network, with convolutional and pooling layers, extracts hierarchical representations of input data. The subsequent classification stage utilizes these features, demonstrating CNN's success in tasks like image classification and object detection. The FE network includes stacked convolutional layers, activation, dropout, and pooling—each contributing to meaningful FE. 

The convolutional layer, sharing kernels across the input, captures local patterns efficiently, reducing model complexity and enhancing training ease. In a CNN, convolved results undergo non-linear transformations through an activation layer, extracting crucial non-linear features for capturing complex data patterns. Following activation, pooling reduces feature map resolution, achieving shift-invariance and retaining essential information. Typically, a pooling layer follows each convolutional layer, coupled with activation. This hierarchical process extracts increasingly complex features from input data for subsequent analysis and classification. The classification stage incorporates Fully Connected (FC) layers and a Softmax operator to determine class probabilities. Global average pooling, an FC layer alternative, summarizes spatial information and reduces parameters, mitigating overfitting. Softmax normalizes outputs into a class probability distribution, aiding predictions. While Softmax is common, studies explore SVM as an alternative for classification in CNNs \citep{hong2017learning}, offering different optimization objectives and decision boundaries. Overall, the classification stage utilizes FC layers or global average pooling, followed by a Softmax operator or alternatives like SVM, for predicting class memberships in input patterns. Subsequent sections detail three CNN architectures tailored for HSC: i) Spectral CNN, ii) Spatial CNN, and iii) Spectral-spatial CNN.

%%%%%%%%%%%%%%%%%%%%%%%%%%
\subsection{\textbf{Spectral CNN}}

Spectral CNNs exclusively process 1D spectral data $x_i \in \mathcal{R}^{B}$, where $B$ denotes the original or reduced spectral bands \citep{9361752}. Tailored for HS data, they treat each pixel's spectral information as a 1D vector, capturing spectral intensity at different wavelengths. Unlike conventional 2D CNNs for image classification, these models consider HSIs as 1D sequences, where each element represents spectral intensity. The input is the pixel's spectral profile, either the original full-dimensional vector or a dimensionality-reduced representation, presenting a focused approach to handle spectral dimensionality.

%%%%%%%%%%%%%%%%%%%%%%%%%%
\begin{figure}[!hbt]
    \centering
    \includegraphics[width=0.48\textwidth]{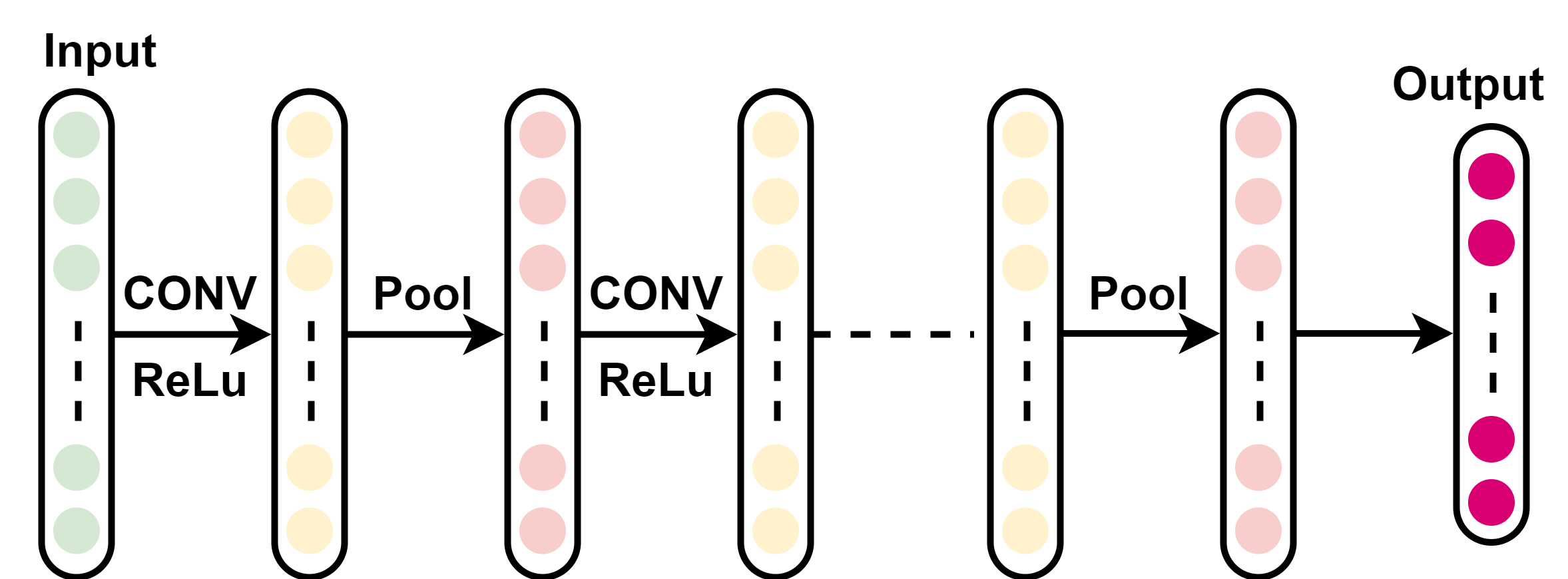}
    \caption{Basic example of Spectral Convolutional Neural Network.}
    \label{Fig8}
\end{figure}
%%%%%%%%%%%%%%%%%%%%%%%%%%

A crucial element in spectral CNNs is the convolutional layer, applying filters to capture local spectral patterns and generate feature maps by element-wise operations. Commonly, pooling layers, such as max or average pooling, downsample and reduce dimensionality. Additional layers like fully connected or recurrent layers can be added for enhanced discriminative power. In \citep{wu2022convolutional}, a CNN addressed overfitting by using $1 \times 1$ convolutional kernels, increased dropout rates, and a global average pooling layer instead of fully connected layers. For handling high correlation among HSI bands, Gao, et al. \citep{gao2023hyperspectral} introduced a CNN architecture transforming the 1D spectral vector into a 2D feature matrix with cascading $1 \times 1$ and $3 \times 3$ convolutional layers. This design allowed feature reuse across different spectral bands. Similarly, Li et al. \citep{li2024hd} employed a global average pooling layer to reduce trainable parameters and extract high-dimensional features.

Wu et al. \citep{wu2017convolutional} proposed a hybrid model combining convolutional and recurrent layers for extracting both position-invariant middle-level features and spectral-contextual details. Similarly, Jin et al. \citep{jin2018classifying} proposed a hybrid architecture that effectively classified healthy and diseased wheat heads by transforming spectral information into a 2D structure for improved analysis. In spectral-based identification of rice seed varieties, Qiu et al. \citep{qiu2018variety} demonstrated CNN's superiority over SVM and KNN, showcasing its effectiveness in accurately identifying rice seed varieties based on spectral characteristics. A similar application in \citep{wu2018discrimination} identified various Chrysanthemum varieties using CNN on spectral data represented by the first five Principal Components (PCs) obtained through PCA, a widely-used dimensionality reduction method. In medical HSI, Huang et al. \citep{huang2018convolutional} used PCA as a preprocessing step, fusing CNN kernels with Gabor kernels for classification, highlighting the efficacy of hybrid architectures and the superiority of CNN over traditional algorithms in spectral-based classification tasks for HSC.

\textbf{Advantages:} Firstly, Spectral CNN captures local spectral patterns effectively through convolutional operations, facilitating discriminative feature extraction \citep{9361752}. Secondly, by integrating pooling operations, spectral CNNs manage high-dimensional HS data, reducing computational complexity. Thirdly, the inclusion of fully connected or recurrent layers enables learning high-level representations and capturing spectral-contextual dependencies, enhancing classification accuracy. 

\textbf{Limitations:} Spectral CNNs primarily focus on spectral features, often neglecting the spatial context. This can result in suboptimal performance when spatial relationships are crucial for accurate classification. HS data have a high dimensionality due to the numerous spectral bands. Spectral CNNs can struggle with the curse of dimensionality, leading to increased computational complexity and memory requirements. Due to the large number of spectral bands and relatively small number of labeled samples, Spectral CNNs are prone to overfitting. This issue is exacerbated in scenarios with limited training data. Variations in spectral signatures caused by different conditions (e.g., illumination, atmospheric effects) can reduce the robustness of Spectral CNNs, making them less effective in diverse real-world scenarios. Manually designing and tuning the spectral feature extraction process can be complex and time-consuming. Spectral CNNs may require extensive preprocessing to achieve optimal results. 

%%%%%%%%%%%%%%%%%%%%%%%%%%
\subsection{\textbf{Spatial CNN}}

\begin{figure}[!t]
    \centering
    \includegraphics[width=0.48\textwidth]{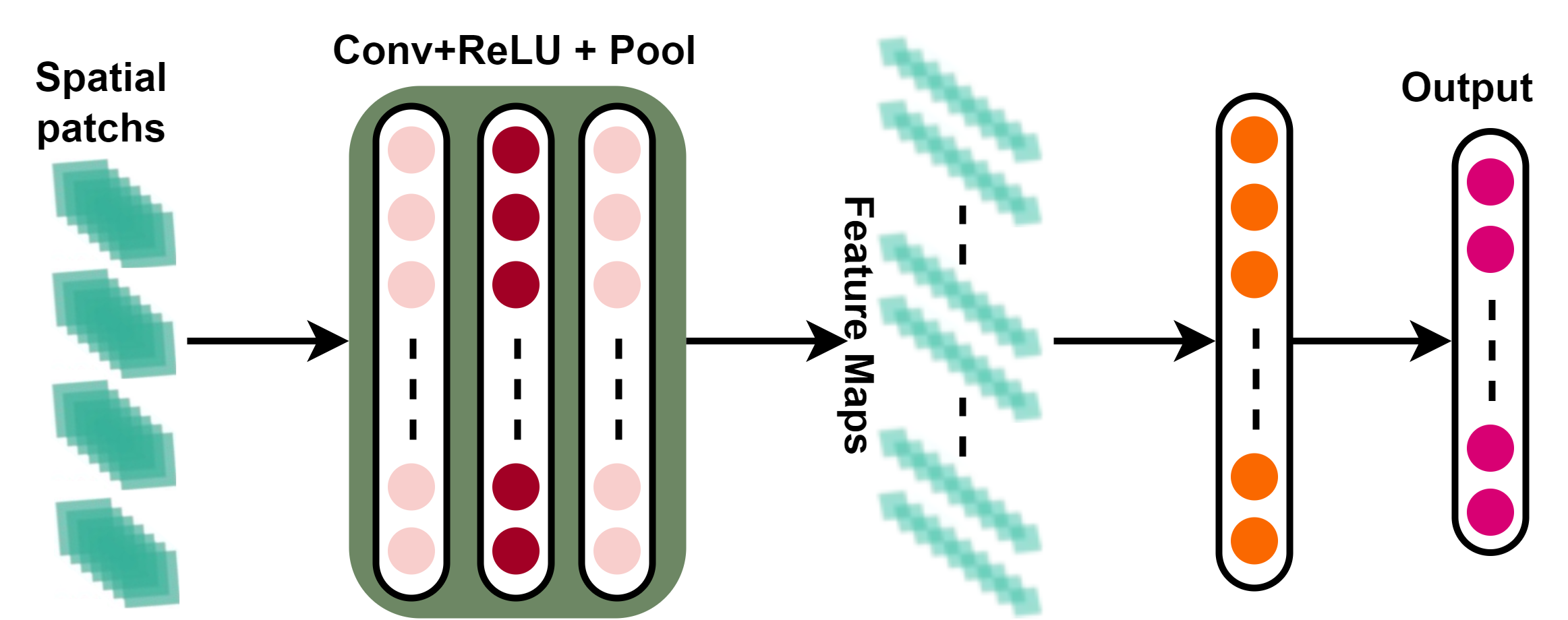}
    \caption{Basic example of Spatial Convolutional Neural Network.}
    \label{Fig9}
\end{figure}

Spatial CNNs, focusing on spatial information in HSIs, address the spatial dimensionality of HS data. Unlike spectral CNNs concentrating on spectral details, spatial CNNs treat the HSI as a 3D volume-width and height as spatial dimensions and the third one represents spectral data at each pixel. The input is a 3D HSI, with convolutional filters capturing local spatial patterns by considering both spectral and neighboring spatial information as shown in Figure \ref{Fig9}.

%%%%%%%%%%%%%%%%%%%%%%%%%%
%%%%%%%%%%%%%%%%%%%%%%%%%%

Pooling layers, like max or average pooling, commonly feature in spatial CNNs to condense feature maps and capture key spatial information. Additional layers, including FC layers, are often integrated to grasp high-level spatial features and enhance classification. In studies by Li et al. \citep{li2018classification}, Haut et al. \citep{haut2019hyperspectral}, and Xu et al. \cite{xu2018hyperspectral}, PCA-derived PCs were employed to infuse refined spatial information into CNN frameworks, demonstrating approaches to bolster DL models for HSC. Wang et al. \citep{wang2021probabilistic} proposed a novel probabilistic neighborhood pooling-based attention network (PNPAN) aiming to enhance model robustness and combat overfitting in HSC. These strategies, including PCA-driven spatial information, random patch networks, and innovative techniques like PNPAN, seek to address challenges linked to limited training samples and overfitting in DL models for HSC.

Ding et al. \citep{ding2017convolutional} extracted patches from 2D input images (representing different spectral bands) to train a 2D-CNN architecture. Data-adaptive kernels were learned to capture pertinent features from HS data. Chen et al. \citep{chen2017hyperspectral} proposed a 2D-CNN augmented with Gabor filtering to address overfitting concerns, effectively preserving spatial details. Zhu et al. \citep{zhu2018deformable} introduced a deformable HSC network using adaptive deformable sampling locations based on spatial features, enhancing the extraction of complex structures. These methods leverage data-adaptive kernels, handcrafted features like Gabor filters, and deformable sampling locations to improve HSC by capturing relevant spatial information and overcoming challenges such as overfitting with limited training samples.

\textbf{Advantages:} Spatial CNN captures vital spatial contextual information for HSC, facilitating the understanding of spatial patterns and relationships. This capability is crucial for handling spatially varying features in HSIs, including structures, textures, and localized objects. Another strength lies in their ability to fuse spatial and spectral features, enhancing classification accuracy by leveraging the complementary aspects of these dimensions. 

\textbf{Limitations:} HSIs often require extensive preprocessing to make them suitable for Spatial CNNs, such as dimensionality reduction or feature selection, adding to the overall complexity and processing time. Spatial CNNs primarily focus on spatial features and may not effectively utilize the rich spectral information available in HS data. This can result in the loss of valuable spectral features necessary for accurate classification. Similar to Spectral CNNs, Spatial CNNs are prone to overfitting, especially when trained on limited labeled data. The high dimensionality and complexity of HSIs exacerbate this issue. Variations in spectral signatures due to different environmental conditions can reduce the robustness of Spatial CNNs, as they are less equipped to handle such variability compared to models designed to exploit spectral information. Models trained on specific spatial patterns in one dataset may not generalize well to other datasets with different spatial characteristics, limiting their applicability across diverse scenarios.

%%%%%%%%%%%%%%%%%%%%%%%%%%
\subsection{\textbf{Spectral-Spatial CNN}}

Spectral-spatial CNNs exploit both spectral and spatial dimensions of HSIs simultaneously, unlike traditional CNNs focusing solely on spatial or spectral aspects. Treating the HSI as a 3D volume, with two dimensions for width and height and a third for spectral information at each pixel, Spectral-Spatial CNNs receive 3D HSI input, where each pixel contains a spectral profile. Comprising three key components: \textbf{Spectral Convolution} utilizes 1D convolutional filters to capture spectral patterns. \textbf{Spatial Convolution} employs 2D convolutional filters for spatial patterns and local dependencies. Finally, \textbf{Fusion and Classification} combines features from both dimensions, enhancing accuracy by leveraging complementary spectral and spatial information. Alternatively, joint spatial-spectral information can also be extracted using 3D CNN models.

%%%%%%%%%%%%%%%%%%%%%%%%%%
\begin{figure}[!hbt]
    \centering
    \includegraphics[width=0.48\textwidth]{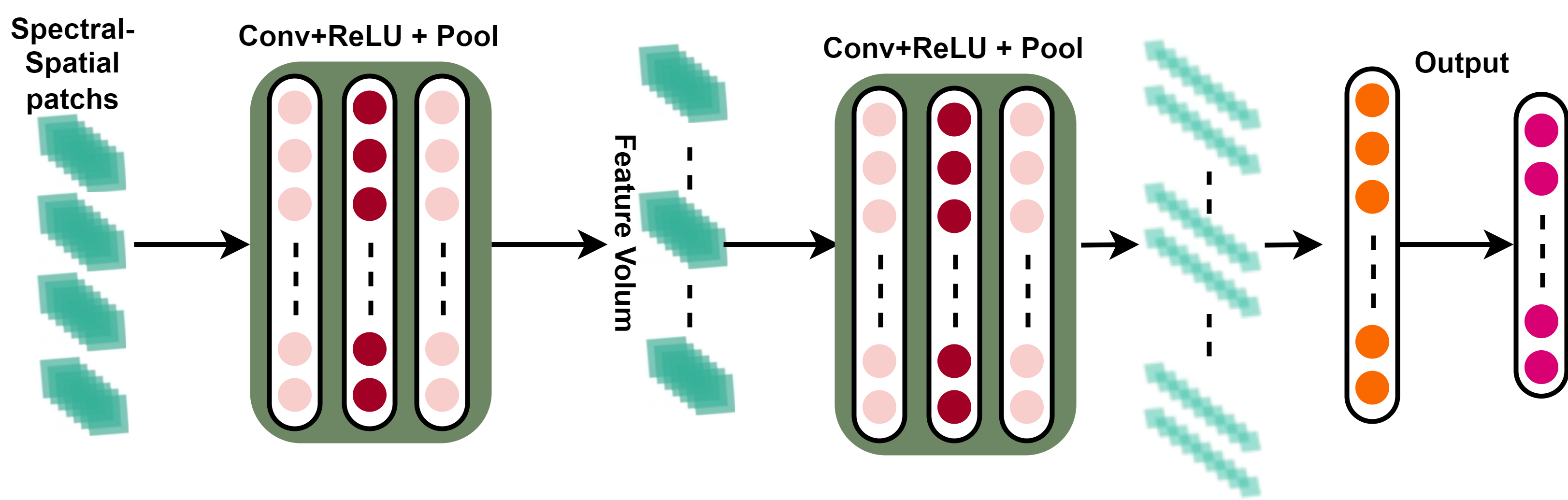}
    \caption{Basic example of Spatial-Spectral Convolutional Neural Network.}
    \label{Fig10}
\end{figure}
%%%%%%%%%%%%%%%%%%%%%%%%%%

Integrating spatial features with spectral information enhances spectral-spatial pixel-wise HSC. For instance, Ran et al. \citep{ran2017hyperspectral} introduced spatial pixel pair features (PPF), emphasizing immediate neighborhood pairs with identical labels for enhanced discriminative power. Zhong et al. \citep{zhong2018spectral} proposed a Spectral-Spatial Residual Network (SSRN) utilizing 3D-convolutions in spectral and spatial residual blocks to extract joint representations. Paoletti et al. \citep{paoletti2018new} presented a 3D CNN framework effectively capturing spectral-spatial relationships for improved classification accuracy. Li et al. \citep{li2019adaptive} employed adaptive weight learning for dynamic adjustment of spatial information importance, capturing variations in spatial context. Roy et al. \citep{roy2020attention} proposed adaptively determined adjustable receptive fields, enhancing joint feature extraction in a spectral-spatial residual network. Additionally, Roy et al. \citep{roy2021maxmin} fused maximum and minimum convolutional features, reported to enhance classification performance in HSC tasks.

Paoletti et al. \citep{paoletti2020rotation} addressed 3D CNN limitations in exploiting rotation equivariance by introducing translation-equivariant representations, enhancing spatial feature robustness in HSC tasks. Zhang et al. \citep{zhang2019hyperspectral} proposed an end-to-end 3D lightweight CNN to counter overfitting and gradient vanishing challenges with limited labeled samples. Jia et al. \citep{jia2020lightweight} introduced spatial-spectral Schroedinger eigenmaps (SSSE) to balance numerous trainable parameters and scarce labeled samples. Roy et al. \citep{roy2020lightweight} incorporated a lightweight bag-of-feature paradigm into a spectral-spatial squeeze-and-excitation residual network. Roy et al. \citep{roy2021morphological} introduced Morphological CNN (MorphCNN), combining spectral and spatial features for effective information utilization.

Li et al. \citep{li2019adaptive} proposed a two-stage framework with adaptive learning of input patch weights in the first stage, extracting joint shallow features and obtaining deep hierarchical features through a Stacked AE (SAE) network. Classification is performed using a Multinomial Logistic Regression (MLR) layer. Li et al. \citep{li2017spectral} introduced a 3D-CNN model, comparing its performance with spectral-based Deep Belief Networks (DBN), SAE, and 2D-spatial CNNs. Roy et al. \citep{roy2020fusenet} presented a bilinear fusion mechanism combining global pooling and max-pooling branches. Jiao et al. \citep{jiao2017deep} introduced a deep multiscale spectral-spatial feature extraction approach, learning discriminative features from spatially diverse images using a Fully Convolutional Network (FCN). These spatial features are fused with spectral information using a weighted fusion strategy for accurate pixel-wise classification.

He et al. \citep{he2018feature} used PCA-transformed images to generate multi-scale cubes, extracting handcrafted features with multi-scale covariance maps to capture spectral-spatial details. Zhang et al. \citep{zhang2017spectral} implemented a dual-channel CNN framework, employing a 1D-CNN for spectral and a 2D-CNN for spatial hierarchical feature extraction, combining both for classification. He et al. \citep{he2017multi} introduced a multiscale 3D deep CNN for end-to-end HSC, jointly learning 1D spectral and 2D multiscale spatial features without pre-processing like PCA. Dong et al. \citep{dong2019band} embedded a band attention module within a CNN to selectively focus on informative bands, reducing redundancy and noise in HSC. Cheng et al. \citep{cheng2018exploring} combined CNN with a metric learning-based framework, using CNN for deep spatial information extraction and metric learning for spectral and spatial feature fusion. Gong et al. \citep{gong2019cnn} merged a multi-scale convolution-based CNN with diversified deep metrics based on determinantal point process (DPP) priors \citep{zhong2015learning}, employing multi-scale filters and DPP-based diversified metrics for enhanced HSI representational ability.

Liu et al. \citep{liu2020multiscale} introduced an HSC framework for multi-scale spatial feature extraction, constructing a three-channel RGB image from HSIs to leverage existing networks. Sellami et al. \citep{sellami2019hyperspectral} presented an adaptive band selection-based semi-supervised 3D CNN, jointly exploiting spectral-spatial features. Ma et al. \citep{ma2018hyperspectral} proposed a two-branch Deep CNN, with one branch extracting spatial information and the other utilizing a contextual DNN for spectral features. Roy et al. \citep{roy2020darecnet} explored a dual-attention-based AE-decoder network for unsupervised HS band selection and joint feature extraction. Roy et al. \citep{roy2021lightweight} introduced a lightweight HetConv3D for HSC with noisy labels, combining spectral and spatial kernel features. Paoletti et al. \citep{paoletti2021separable} presented a separable attention network dividing input feature maps for global contextual information encoding. Roy et al. \citep{roy2021revisiting} introduced G2C-Conv3D to combine intensity-level semantic information and gradient-level detailed information during convolution operations. 

\textbf{Advantages:} Spectral-spatial CNNs offer enhanced discriminative power by capturing detailed relationships between spectral bands and spatial patterns, enabling effective differentiation of land cover classes. These models incorporate spatial convolutional filters to grasp spatial contextual information, crucial for handling spatial structures and textures in HSC. The fusion of spectral and spatial features in spectral-spatial CNNs takes advantage of the complementary nature of these dimensions, resulting in improved classification performance through the exploitation of unique information from both spectral and spatial components. 

\textbf{Limitations:} The integration of both spatial and spectral information increases the model's computational demands, requiring more processing power and memory, which can be a constraint for large-scale or real-time applications. These models are prone to overfitting, especially when trained on small or imbalanced datasets. This is due to the high number of parameters that need to be learned, which can lead to poor generalization of unseen data. Spatial-spectral CNNs require substantial amounts of labeled training data to perform effectively. Obtaining such data for HSIs can be challenging and costly. The design and tuning of Spatial-Spectral CNNs can be complex, requiring careful selection of hyperparameters, network architecture, and training strategies to achieve optimal performance. Like other DL models, Spatial-spectral CNNs often lack interpretability. Understanding the specific features or patterns the model uses to make classification decisions can be difficult, hindering trust and adoption in certain applications. 

%%%%%%%%%%%%%%%%%%%%%%%%%%
\subsection{\textbf{Graph Convolutional Networks for HSC}}

\begin{figure}[!hbt]
    \centering
    \includegraphics[width=0.48\textwidth]{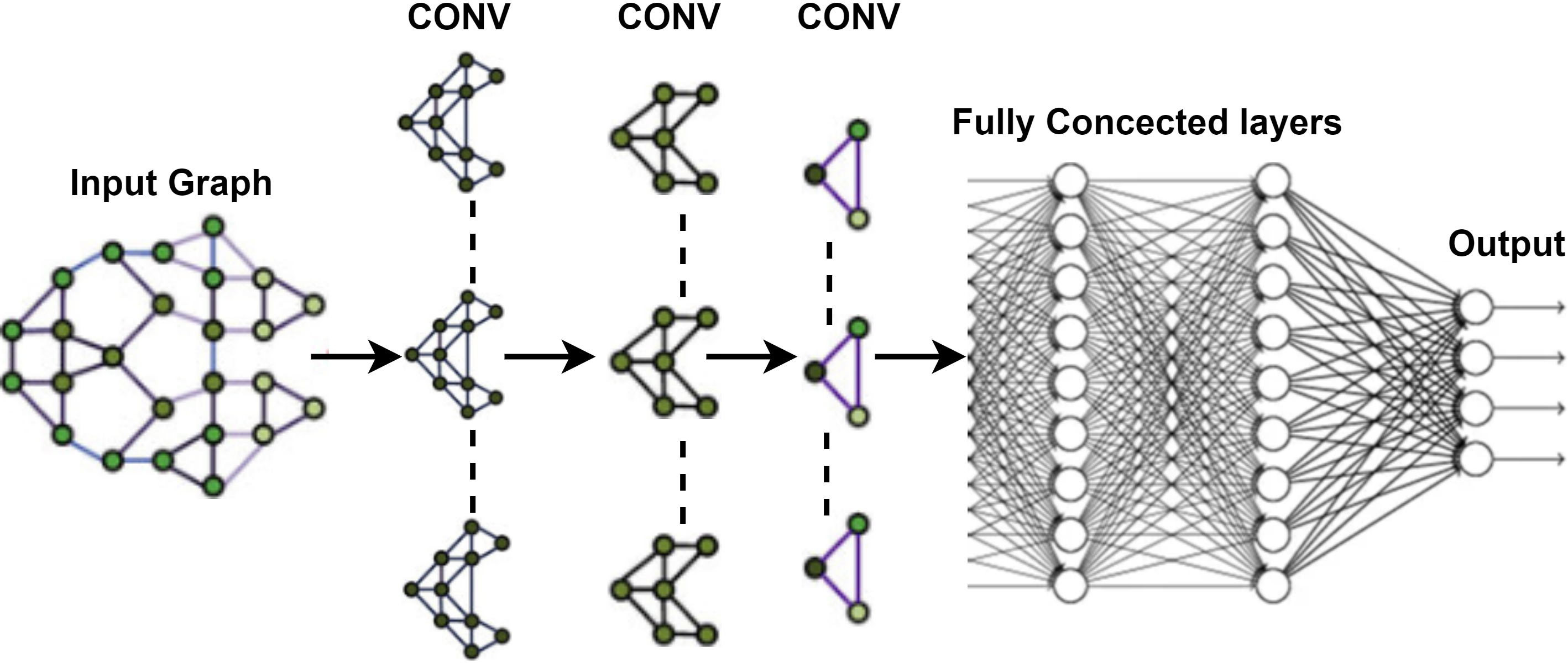}
    \caption{Basic example of Graph Convolutional Neural Network.}
    \label{Fig11}
\end{figure}

Graph Convolutional Networks (GCNs) have garnered attention for their adeptness in processing non-grid high-dimensional data and their adaptable network architecture \citep{kipf2016semi}. These characteristics open up possibilities for effective handling of HS data \citep{yu2023gpf, 10387500}. GCNs are particularly proficient in capturing relationships between data or samples, making them well-suited for modeling spatial relationships among spectral signatures in HSIs, as depicted in Figure \ref{Fig11}. However, the construction of large graphs poses challenges due to computational demands, limiting the popularity of GCNs in HSC compared to CNNs. Despite this, exploratory studies have begun to leverage GCNs in certain HSC tasks \citep{hong2021graph}.

%%%%%%%%%%%%%%%%%%%%%%%%%%
%%%%%%%%%%%%%%%%%%%%%%%%%%

Here are some details on how GCNs can be used for HSC: In HSIs, the graph representation assigns each pixel as a node, utilizing the spectral signatures as node features. The construction of the graph involves connecting neighboring pixels based on spatial proximity, with the edges denoting spatial relationships between pixels. The fundamental component of GCNs is the graph convolutional layer, facilitating information propagation and aggregation across the graph. In each layer, node features are updated by aggregating information from neighboring nodes, employing a convolutional operation that considers both node features and edge connections. Graph pooling operations can be applied to handle the high-dimensional nature of HS data, reducing spatial resolution while retaining essential information. These pooling operations aggregate nodes based on importance or connectivity, resulting in a coarser graph representation. To further enhance GCNs, graph attention mechanisms can be integrated to weigh the importance of neighboring nodes during information aggregation, enabling adaptive attention to nodes based on their significance in the classification task.

By incorporating graph structures and spatial relationships, GCNs provide a robust framework for HSC, capturing intricate interactions between spectral signatures and leveraging spatial context for improved accuracy. Despite their potential, it is crucial to address computational complexities and scalability issues with large HS datasets. Researchers have proposed strategies to mitigate these challenges. For instance, a second-order GCN in \citep{qin2018spectral} models spatial-spectral relations on manifolds \citep{hong2019learnable} to reduce computational costs, while \citep{wan2019hyperspectral} utilizes superpixel segmentation to enhance GCN efficiency in handling larger pixel sets for land cover classification. However, fundamental issues persist. Hong et al. proposed miniGCN in \citep{hong2021graph}, training GCNs in a mini-batch fashion akin to CNNs, effectively reducing computational costs and enabling quantitative comparisons and fusion with CNNs, leading to the development of FuNet for HSC.

Transfer learning is a valuable technique for GCNs in HSC. By pre-training a GCN on a large or related dataset, it captures generic spatial patterns, and fine-tuning the target HSC task with a smaller labeled dataset enhances performance, especially when labeled HS data is limited. Moreover, integrating GCNs with other DL models, such as CNNs, proves effective. A two-stream architecture can be employed, where one stream utilizes GCNs for spectral information, and the other employs CNNs for spatial information. The fusion of outputs from both streams contributes to the final classification decision, leveraging the complementary strengths of the two models.

%%%%%%%%%%%%%%%%%%%%%%%%%%
\subsection{\textbf{Future Research Directions}}

While CNNs have demonstrated impressive performance in HSC, integrating spatial and spectral information poses a challenge. Existing frameworks often sacrifice spectral details in pursuit of enhanced spectral-spatial representation through DR techniques. Future research should prioritize the development of robust HSC models that effectively integrate spatial and spectral information without compromising spectral details. However, such approaches increase computational complexity, demanding efficient solutions for real-time deployment on resource-constrained platforms. Utilizing parallel processing techniques with Field-Programmable Gate Arrays (FPGAs) and GPUs can address these challenges, ensuring both computational efficiency and performance accuracy.

In addition to computational concerns, the depth of CNNs requires abundant labeled training data, which is limited in HSI. To overcome this, integrating CNNs with unsupervised or semi-supervised learning approaches is crucial. Leveraging unlabeled or partially labeled data enhances the training process, improving classification performance. Furthermore, the generalization ability of CNNs needs exploration beyond traditional grid structures. Investigating their applicability to diverse HSI data can be enhanced by combining CNNs with GCNs, thereby creating a versatile and generalized framework. This integration allows the capture of spatial relationships in non-grid data, contributing to a more adaptable HSC approach. Overcoming challenges related to data availability, generalization, and incorporating GCNs will advance CNNs for HSC, unlocking their full potential for efficient and accurate classification tasks. Ongoing research in these areas is essential for further advancements in the field.

%%%%%%%%%%%%%%%%%%%%%%%%%%
\section{\textbf{Deep Belief Network for HSC}}
\label{DBN}

\begin{figure}[!hbt]
    \centering
    \includegraphics[width=0.48\textwidth]{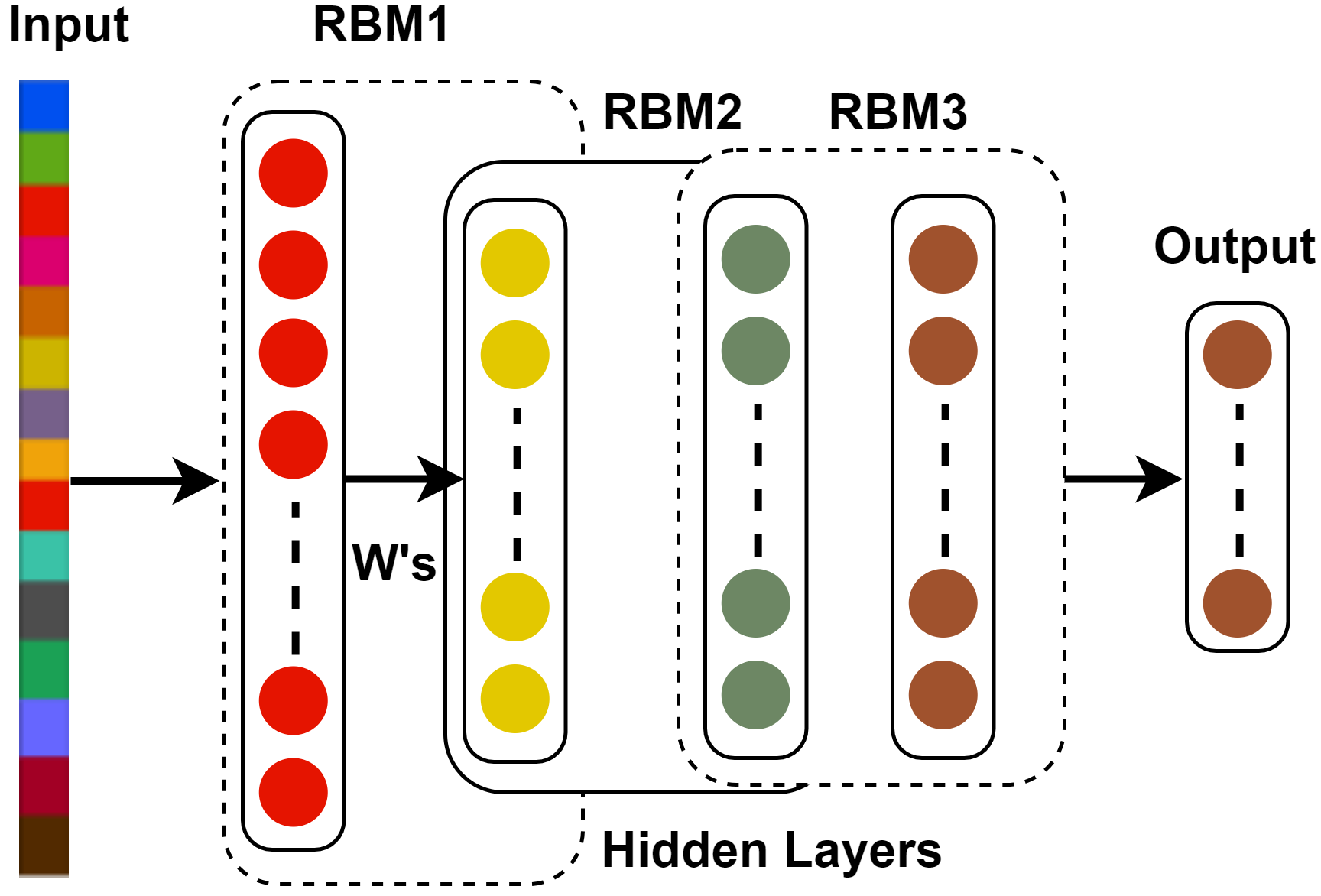}
    \caption{Basic example of three layered Deep Belief Network for HSC.}
    \label{Fig12}
\end{figure}

Introduced by Hinton et al. in 2006 \citep{hinton2006fast}, the Deep Belief Network (DBN) is a hierarchical DNN that utilizes unsupervised learning to sequentially learn features from input data as shown in Figure \ref{Fig12}. Constructed with multiple layers using Restricted Boltzmann Machines (RBMs), RBMs are two-layer neural networks where visible units connect to hidden units \citep{zhang2018overview}. RBMs play a crucial role in extracting informative features from input data, forming a layer-by-layer architecture in DBNs. This hierarchical structure enables greedy training, with each RBM capturing progressively abstract features from HS data. 

%%%%%%%%%%%%%%%%%%%%%%%%%%
%%%%%%%%%%%%%%%%%%%%%%%%%%

Widely applied in HSC, DBNs, such as in Ayhan et al.'s work \citep{ayhan2017application}, have been utilized for land cover classification, emphasizing the integration of spectral and spatial information. The typical training process involves unsupervised pre-training with unlabeled samples, followed by supervised fine-tuning using labeled samples. However, this approach may lead to co-adaptation issues among hidden units and activation neuron sparsity, limiting the diversity of learned features. Addressing these challenges \citep{shaham2016deep, xiong2015diversity}, Zhong et al. proposed a diversified DBN model \citep{zhong2017learning}. This model introduces a regularization technique during both pre-training and fine-tuning stages, promoting diversity among hidden units and activation neurons. By mitigating co-adaptation and enhancing activation selectivity, the diversified DBN model provides a potential solution to improve DBN classification performance in HSC, enhancing representation power and discriminative capabilities.

In HSC, researchers have introduced diverse strategies for efficient texture feature extraction. Li et al. proposed a DBN-based texture feature enhancement framework \citep{li2018hyperspectral}, incorporating band grouping, sample band selection, and guided filter techniques to improve texture features. Both methods utilize DBN-based frameworks and parallel layers architectures to enhance texture features, emphasizing the importance of effective feature extraction for accurate HSC.

To improve classification accuracy, studies have focused on integrating spectral and spatial information \citep{li2018hyperspectral}. Li et al. presented a DBN framework with a logistic regression layer, showcasing that joint exploitation of spectral-spatial features enhances accuracy \citep{li2019deep}. Similarly, Sellami et al. proposed a spectral-spatial graph-based RBM method \citep{sellami2019spectra}, constructing a spectral-spatial graph to measure similarity based on both spectral and spatial details. The RBM extracted joint spectral-spatial features, contributing to improved classification performance when integrated into a DBN and logistic regression layer. These approaches highlight the significance of considering both spectral and spatial information for effective HSC.

The above discussion highlighted the limited use of DBNs in HSC compared to other DNNs, emphasizing the need for robust techniques integrating spatial and spectral features. A promising avenue for research involves refining the pretraining and fine-tuning processes in DBNs to address issues like inactive or over-tolerant neurons. Improving the training process can enhance performance and generalization in HSC applications. Further advancements in HSC should explore DBN-based approaches that efficiently combine spatial and spectral features while focusing on regularization techniques to overcome challenges related to neuron activity. These efforts can contribute to more effective and resilient DBN-based methods for accurate HSC.

%%%%%%%%%%%%%%%%%%%%%%%%%%
\section{\textbf{Recurrent Neural Networks for HSC}}
\label{RNN}

\begin{figure}[!hbt]
    \centering
    \includegraphics[width=0.48\textwidth]{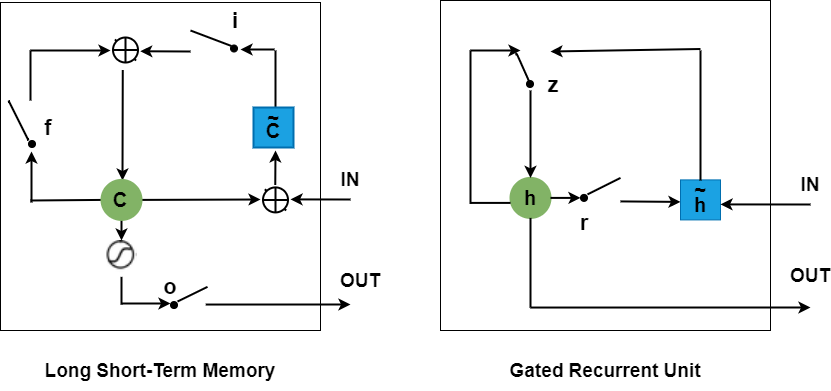}
    \caption{Internal architecture of LSTM and GRU \cite{ahmad2021hyperspectral}}
    \label{Fig13}
\end{figure}

RNNs, featuring loop connections that link current and previous steps, excel in learning from temporal sequences \citep{williams1989learning}. In the context of HS data, RNNs treat spectral bands as sequential time steps, providing a temporal processing advantage \citep{paoletti2020scalable}. Three fundamental RNN models include Vanilla, LSTM, and GRU (Figure \ref{Fig13}), each offering distinct strategies to address challenges related to learning and capturing long-term dependencies in sequential data. While Vanilla RNNs are straightforward, they may struggle with extended sequences, prompting the use of LSTM and GRU models designed to mitigate the vanishing gradient problem and handle long-term dependencies more effectively. Leveraging the temporal aspects of RNNs for sequential processing in HSI data allows researchers to harness their power in capturing and modeling temporal dynamics.

%%%%%%%%%%%%%%%%%%%%%%%%%%
%%%%%%%%%%%%%%%%%%%%%%%%%%

Vanilla RNN, the simplest RNN form, can face information degradation issues, impacting its performance with high-dimensional data like HSI due to the vanishing/exploding gradient problem. LSTM models are developed to tackle this, introducing cell and hidden states along with input, forget, and output gates, enabling selective information control. LSTM excels at capturing and retaining relevant information over long sequences, addressing the temporal dependencies in data. GRU, a variant simplifying LSTM by combining gates, offers comparable performance while being computationally efficient.

Hang et al. \citep{hang2019cascaded} proposed an RNN framework using the novel parametric rectified tanh activation function and GRU for sequential HS data analysis. Zhang et al. \citep{zhang2018spatial} introduced an RNN-based local spatial sequential (LSS) method, fusing Gabor filters and differential morphological profiles for low-level feature extraction, followed by RNN modeling for high-level feature extraction. Sharma et al. \citep{sharma2018land} utilized LSTM cells for multi-temporal and multi-spectral information, capturing complex patterns and dependencies in HSI for effective land cover classification.

Wu et al. \citep{wu2017convolutional} introduced a convolutional RNN model for HSC, using initial convolutional layers to extract position-invariant middle-level features and recurrent layers to capture spectral-contextual details. In a semi-supervised setting, Wu et al. \citep{wu2017semi} applied a similar CRNN model, incorporating pseudo labels for leveraging unlabeled data. Luo et al. \citep{luo2018shorten} introduced a CNN-parallel GRU-based RNN architecture, simplifying GRU training and enhancing performance by combining it with the CNN component. Liu et al. \citep{liu2017bidirectional} presented the Bidirectional CLSTM model, jointly exploiting spectral-spatial features of HSI. Shi et al. \citep{shi2018multi} combined multiscale local spectral-spatial features from a 3D CNN with a hierarchical RNN. Yang et al. \citep{yang2018hyperspectral} proposed recurrent 2D CNN and 3D CNN models for HSC. Hang et al. \citep{hang2019cascaded} introduced a cascade RNN architecture with two layers of GRU-based RNNs, reducing redundant spectral bands in the first layer and learning features in the second layer.

The aforementioned discussion explored recent advances in RNN-based techniques for HSC, acknowledging their notable success in classification performance. However, challenges exist, particularly in handling lengthy input sequences that may lead to overfitting. Addressing this, future research should explore strategies to reduce input sequence length, preserving important spectral and spatial information. Additionally, investigating parallel processing tools can enhance computational efficiency, and alternative approaches like spectral band grouping merit exploration for improved discrimination. Another promising direction involves extending RNN-based HSC frameworks to multi-temporal HS imagery, enabling models to capture dynamic changes over time for more accurate and robust results. These efforts will not only overcome challenges but also enhance the practical utility of RNN-based models in real-world scenarios and contribute to advancements in remote sensing applications.

%%%%%%%%%%%%%%%%%%%%%%%%%%
\section{\textbf{Autoencoders for HSC}}
\label{AEs}

\begin{figure}[!hbt]
    \centering
    \includegraphics[width=0.45\textwidth]{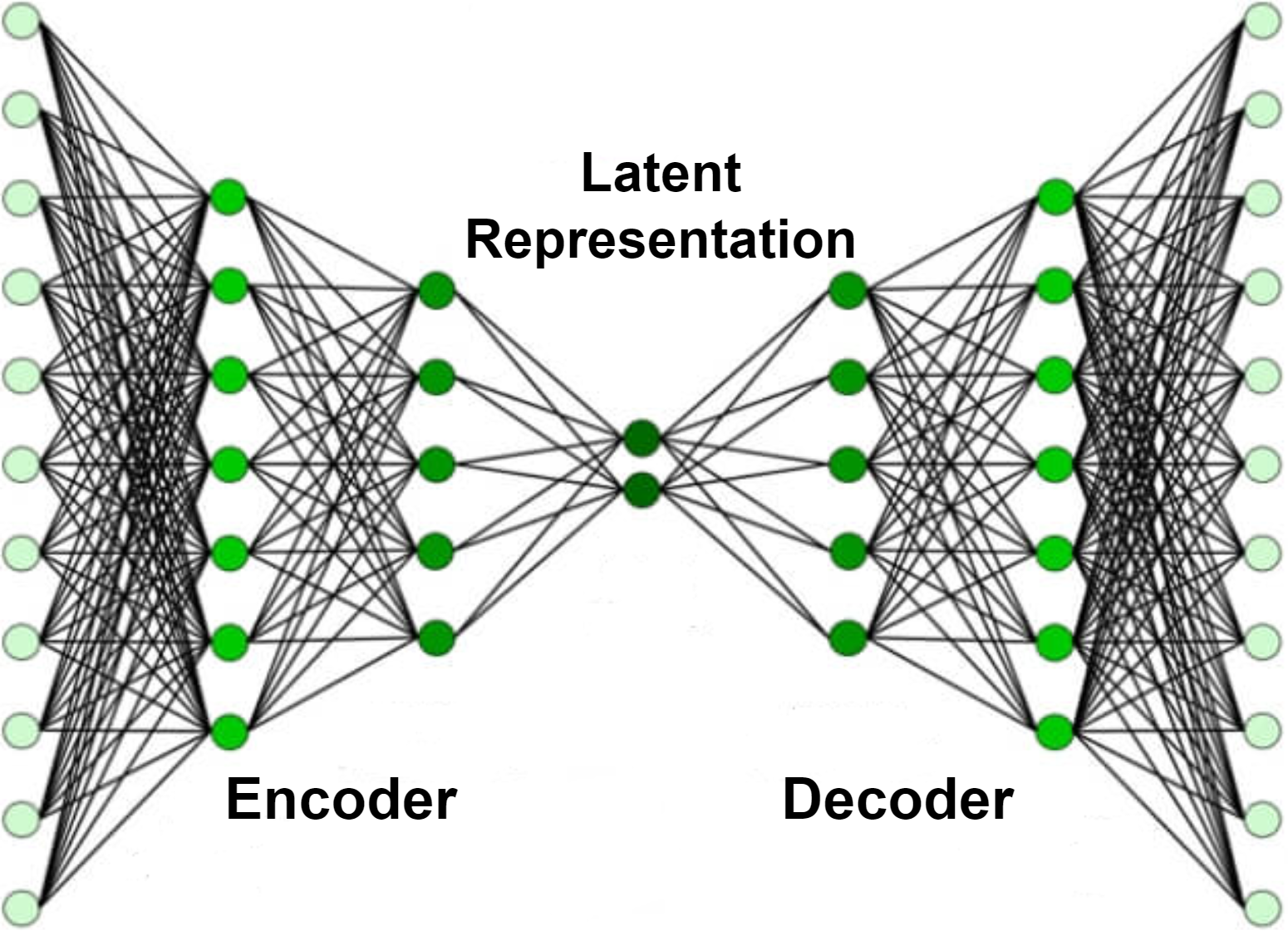}
    \caption{Autoencoder architecture.}
    \label{Fig14}
\end{figure}

The AE is a widely adopted symmetric neural network in HSC for unsupervised feature learning \citep{10509579, li2024s2mae,10528333}. It prioritizes generating a compressed feature representation of high-dimensional HS data instead of direct classification. The AE architecture typically includes an input layer, a hidden or encoding layer, a reconstruction or decoding layer, and an output layer as shown in Figure \ref{Fig14}. During training, the AE learns to encode input data into a latent representation that accurately reconstructs the original input. This is achieved by minimizing the reconstruction error, measuring the difference between the input and output, and enabling effective learning of a compressed feature representation.

%%%%%%%%%%%%%%%%%%%%%%%%%%
%%%%%%%%%%%%%%%%%%%%%%%%%%

The AE is designed to extract essential features from input data, discarding irrelevant information through unsupervised learning. This process yields compact representations, beneficial for tasks like classification in HSC. The Stacked AE (SAE) \citep{10444024} extends this architecture, stacking multiple AE layers to learn increasingly abstract features. The Denoising AE (DAE) \citep{10472513}, another variant, intentionally corrupts input data with noise, forcing the model to reconstruct the original signal. This enhances the DAE's ability to handle noisy input, making it useful in scenarios with data corruption. Overall, the AE, SAE, and DAE are powerful tools in HSC for learning compressed, abstract features and handling noisy input.

In Zhu et al.'s \citep{zhu2017auto} approach, multi-layer AEs with maximum noise fraction were combined to learn high-level representations from HS data. Hassanzadeh et al. \citep{hassanzadeh2017unsupervised} employed a multi-manifold learning framework combined with the Counteractive AE for unsupervised HSC. Zhang et al. \citep{zhang2017recursive} proposed a framework utilizing a Recursive AE (RAE) network to jointly exploit spectral-spatial features, focusing on recursive feature extraction from neighboring pixels based on spectral similarity. Hao et al. \citep{hao2017two} introduced a two-stream DNN with a class-specific fusion scheme, incorporating an SAE for spectral features and a CNN for spatial information. Sun et al.'s architecture \citep{sun2017encoding} integrated PCA, guided filters, and sparse AEs with batch-based training to fuse spectral and spatial information. In Zhao et al.'s framework \citep{zhao2017spectral}, an HSC approach involved selecting spatial resolution from HSIs and utilizing stacked sparse AEs for high-level feature extraction, with classification performed using Random Forest. The hybrid architectures proposed by Sun et al. \citep{sun2017encoding} and Zhao et al. \citep{zhao2017spectral} effectively combined PCA, guided filters, and sparse AEs for feature extraction in both spectral and spatial domains.

Wan et al. \citep{wan2017stacked} employed SAEs to extract diverse representations, including spectral-spatial features and multi-fractal features, alongside higher-order statistical representations. Lv et al. \citep{lv2017remote} proposed a combination of SAEs and ELMs, where features were segmented, transformed using SAEs, and rearranged for input to an ELM-based classifier. Ahmad et al. \citep{ahmad2019multi} introduced a computationally efficient multi-layer ELM-based AE for HSC, aiming to enhance feature learning efficiency and effectiveness. Zhou et al. \citep{zhou2019learning} addressed intra-class variability and inter-class similarity challenges in HSC by incorporating a local Fisher discriminant regularization technique for learning compact and discriminative features. Lan et al. \citep{lan2019hyperspectral} combined a k-sparse denoising AE with spectral-restricted spatial features to handle high intra-class variability in spatial features. Paul et al. \citep{paul2018spectral} segmented spectral components in HSI based on mutual information measure, reducing computation time during feature extraction using SAE and incorporating spatial information with Empirical Mode Decomposition (EMD). Liu et al. \citep{liu2019spatial} utilized an SAE-based approach for classifying oil slicks on the sea surface, jointly exploiting spectral-spatial features to improve accuracy.

The above discussion explored recent advancements in AE-based techniques for HSC. While these approaches demonstrate strong predictive performance and generalization, there remains a need for more sophisticated methods. Many discussed techniques underutilized spatial information in HSIs, prompting further research to effectively harness joint spatial and spectral information for improved HSC. Moreover, challenges such as high intra-class variability and inter-class similarity persist in HSC, impacting classification accuracy. While some reviewed methods address these issues, additional research is crucial. Exploring strategies like pre-training, co-training, and adaptive neural networks within AE-based frameworks could offer promising solutions to enhance accuracy and robustness.

%%%%%%%%%%%%%%%%%%%%%%%%%%
\section{\textbf{Limitations of Traditional Deep Learning}}
\label{LTDL}

Applying DL to HSC faces challenges arising from the unique characteristics of HS imaging. These challenges include a large number of narrow spectral bands, high spectral resolution, low spatial resolution, and limited labeled training data. The curse of dimensionality, exacerbated in high-dimensional data scenarios, impacts classification performance, particularly when labeled data is insufficient. The Hughes phenomena, associated with small labeled datasets, can lead to overfitting and poor generalization. The scarcity of labeled HS data, a time-consuming and expensive task, hampers model training. Addressing these challenges requires strategies like feature selection, dimensionality reduction, and augmenting training data through unsupervised or semi-supervised learning. Efficient labeling methods can enhance dataset availability. High intra-class variability, instrumental noise, redundant spectral bands, and spectral mixing further complicate HSC \citep{hong2018augmented}. Researchers are exploring techniques involving domain knowledge, denoising algorithms, feature extraction, and spatial-spectral fusion to tackle these challenges, improving the robustness and accuracy of HSC models. Following are some main challenges that come across when DL is applied to HSC:

\textbf{Complex Training:} Training deep models for HS analysis, belonging to the NP-complete problem class, poses challenges in convergence, especially with a large number of parameters \citep{ahmad2021artifacts, nguyen2017optimization}. Optimization techniques like SGD, SGDM, RMSProp, Adam, AdamW, diffGrad, RAdam, gradient centralization, and AngularGrad have addressed this, enhancing training efficiency and convergence for HS analysis \citep{dubey2019diffgrad, liu2019variance, yong2020gradient}. \textbf{Model Interpretability:} The interpretability of DNNs remains challenging due to their black-box nature. Despite efforts to enhance interpretability, DNNs are inherently complex, impacting understanding and decision-making during optimization. \textbf{Limited Training Data:} Supervised deep models in HS imaging face challenges due to limited labeled data, risking overfitting, known as the Hughes phenomena \citep{erhan2010does,dutta2013deep}. The high dimensionality of HSI exacerbates this challenge, requiring careful parameter tuning during training \citep{paoletti2019deep}. \textbf{High Computational Burden:} DNNs handling large datasets encounter challenges in memory, computation, and storage. Advances in parallel and distributed architectures, along with high-performance computing, enable efficient processing of large datasets \citep{alom2019state}. \textbf{Training Accuracy Degradation:} The assumption that deeper networks yield higher accuracy is not always valid. Deeper networks may face challenges like the exploding or vanishing gradient problem, adversely affecting convergence \citep{he2016deep, bengio1994learning}. By mitigating these artifacts and advancing robust techniques, researchers can enhance the reliability and accuracy of HSC, facilitating applications in remote sensing, agriculture, and environmental monitoring.

%%%%%%%%%%%%%%%%%%%%%%%%%%
\section{\textbf{Transformers for HSC}}
\label{Trans}

The Transformer architecture, introduced by Vaswani et al. \citep{NIPS2017_3f5ee243} in 2017, has emerged as a powerful DL model for various tasks, such as machine translation, language understanding, text generation, etc. Unlike traditional RNNs or CNNs, the Transformer architecture relies on a self-attention mechanism to capture dependencies between different elements in a sequence, enabling it to effectively model long-range dependencies. This self-attention mechanism allows the model to attend to different parts of the input sequence to build contextualized representations. The Transformer architecture has also found applications in the field of HSC as shown in Figure \ref{Fig15}. HSC involves the classification of each pixel in an HSI into different land cover or material classes \citep{10399798}. The Transformer architecture offers several advantages in this context:

%%%%%%%%%%%%%%%%%%%%%%%%%%
\begin{figure*}[!hbt]
    \centering
    \includegraphics[width=0.95\textwidth]{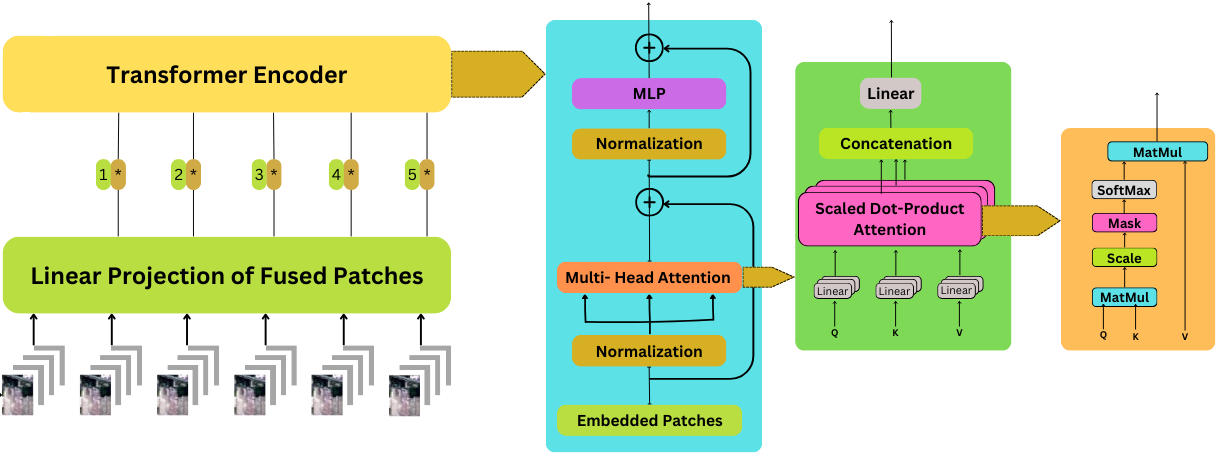}
    \caption{Spatial-Spectral Transformer architecture.}
    \label{Fig15}
\end{figure*}
%%%%%%%%%%%%%%%%%%%%%%%%%%

\begin{enumerate}
    \item \textbf{Capturing Spectral-Spatial Dependencies:} The Transformer architecture can effectively capture dependencies between spectral bands and spatial locations in HSIs. By attending to both the spectral and spatial dimensions, the model can learn to extract discriminative features that leverage the rich information present in HS data \citep{10419080}.

    \item \textbf{Handling Long-Range Dependencies:} HSIs typically exhibit long-range dependencies, where the spectral characteristics of a pixel may depend on distant pixels in both spectral and spatial dimensions \cite{10419133}. The self-attention mechanism in the Transformer architecture allows the model to capture these long-range dependencies, enabling it to model complex relationships between pixels and make more accurate predictions \citep{10387414}.

    \item \textbf{Contextualized Representation Learning:} The Transformer architecture can generate contextualized representations for each pixel in an HSI. By considering the entire image context, the model can learn to encode information about neighboring pixels and their classes, leading to improved classification accuracy.

    \item \textbf{Interpretability:} The attention mechanism in the Transformer architecture provides interpretability by highlighting the relevant spectral and spatial features that contribute to the classification decision \cite{10401936}. This can help in understanding the reasoning process of the model and identifying important features for classification.
\end{enumerate}

In recent studies, researchers have explored different variations of the Transformer architecture for HSC \citep{ahmad2024pyramid, BUTT2024103773, 9848995, li2024casformer, hong2020x}, including adaptations like the HS Transformer Network and the Spectral Transformer Network. Transformer models have shown promising results in HSC tasks, achieving competitive performance compared to traditional approaches like CNNs and RNNs \citep{yu2024hypersinet, 8710262, 9851412}. Overall, the Transformer architecture offers a powerful and flexible framework for HSC \citep{ahmad2024importance}. Its ability to capture spectral-spatial dependencies, handle long-range dependencies, and generate contextualized representations makes it a promising approach for extracting discriminative features from HS data and improving classification accuracy.

Qing et al. \citep{qing2021improved} introduced SAT Net, an enhanced Transformer network by incorporating self-attention and spectral attention mechanisms. SAT Net adeptly captures spectral-spatial characteristics. It addresses issues like gradient dispersion and overfitting through the utilization of multi-head self-attention modules and a residual network structure. He et al. \citep{he2021spatial} proposed a spatial-spectral Transformer (SST) framework. SST integrates a spatial feature extraction CNN with a modified Transformer model known as DenseTransformer to capture sequential correlations within spectral data. 

Scheibenreif et al. \citep{scheibenreif2023masked} explored self-supervised masked image reconstruction to enhance Transformer models. Through modifications to the vision Transformer architecture, such as block-wise patch embeddings, spatial-spectral self-attention, spectral positional embeddings, and masked self-supervised pre-training, the study achieves notable accuracy improvements. Additionally, it demonstrates the effectiveness of self-supervised pre-training in enhancing Transformer accuracy, especially in scenarios with limited labeled training data. Qin et al. \citep{qin2023spatial} presented a Transformer-based framework that leverages spatial-spectral-associative contrastive learning to extract spatial and spectral features. This framework employs spatial and spectral Transformer modules to learn high-level semantic features and generates label information through spatial-spectral augmentation and image entropy. Associative optimization integrates cross-domain features, while a Transformer-based classifier addresses class imbalance by incorporating class intersection over union in the loss function.

Ma et al. \citep{ma2022hyperspectral} presented the NEHT network, aiming to overcome the limitations of CNNs in capturing long-range spectral dependencies. The NEHT network integrates a 2D convolution module for dimensionality reduction, stacked CNNs to preserve spectral distribution, and a Transformer self-attention mechanism to capture long-term relationships. It extracts group-wise spatial-spectral characteristics, employs a feature fusion mechanism, and demonstrates superior performance compared to existing methods. Dang et al. \citep{dang2023double} proposed the Double-Branch Feature Fusion Transformer (DBFFT) model. This model combines CNNs and Transformers to extract spectral and spatial features, utilizing two attention modules to dynamically adjust the relevance of spectral bands and pixels for classification. 

Tan et al. \citep{tan2021deep} presented a novel method by integrating DL, endmember extraction and purification (EMP), and Transformer models. Unlike CNNs, this approach harnesses the Transformer models' capacity to capture long-range dependencies and global context information. Initially, EMP features are extracted from the HSI to maximize spatial and spectral information while reducing the band count. Subsequently, a deep network model incorporating transformer-iN-transformer (TNT) modules is constructed for end-to-end classification. These modules extract patch- and pixel-level data from the input EMP cubes, effectively utilizing both global and local information.

Hong et al. \citep{9627165} proposed a Transformer-based backbone network designed for HSC. It includes two main modules: GroupWise spectral embedding (GSE) to capture local spectral differences and cross-layer adaptive fusion (CAF) for enhancing information flow between layers. He et al. \citep{he2022two} proposed Spa-Spe-TR, a two-branch pure Transformer model comprising a spatial Transformer (Spa-TR) for spatial feature extraction and a spectral Transformer (Spe-TR) for capturing spectral dependencies. Their branch fusion technique efficiently combines joint spatial-spectral features, enhancing their relevance and discriminability.

The HSI Transformer (HiT) \citep{yang2022hyperspectral} achieved higher accuracy using two modules: the Spectral-Adaptive 3D Convolution Projection (SACP) module and the convolutional Permutator module. The SACP module employs spectral-adaptive 3D convolutional layers to extract both local spatial and long-term spectral information, while the convolutional Permutator module encodes representations in height, width, and spectral dimensions through depthwise and pointwise convolutions. Roy et al. \citep{roy2023spectral} introduced the Spectral-Spatial Morphological Attention Transformer (MorphFormer) model, leveraging learnable spectral and spatial morphological convolutions. MorphFormer captures intrinsic shape information. It incorporates a novel morphological attention mechanism to fuse the CLS token with HSI patch tokens, enriching the abstract representation. Tu et al. \citep{tu2022local} proposed LSFAT, a Local Semantic Feature Aggregation-based Transformer architecture. LSFAT enhances the ability of Transformers to represent long-range dependencies of multi-scale features by introducing homogeneous region concepts and neighborhood-aggregation-based embedding (NAE) and attention (NAA) modules. These modules adeptly form multi-scale features and hierarchically capture local spatial semantics. Additionally, LSFAT incorporates a reusable classification token in attention calculation, and a fully connected layer is employed in the final stage for classification after Transformer encoding.

Tang et al. \citep{tang2023double} proposed DATE, a Double-Attention Transformer Encoder architecture that efficiently integrates both local spatial and global spectral data by employing two self-attention modules: spectral attention (SPE) and spatial attention (SPA). SPE captures global spectral information from all spectral tokens, while SPA combines local spatial and global spectral properties for each band, facilitating the effective fusion of spectral and spatial semantic data. Zhong et al. \citep{zhong2021spectral} introduced SSTN, a Spectral-Spatial Transformer Network, alongside the Factorized Architecture Search (FAS) framework. SSTN incorporates novel spatial attention and spectrum association modules to capture long-range interactions and spectral-spatial correlations in the data efficiently. The FAS framework aids in discovering optimal network designs by decomposing the search process into two distinct subprocedures. This study highlights the advantages of Transformer-based architectures and the effectiveness of factorized architecture search. Zhao et al. \citep{zhao2023hyperspectral} proposed a novel framework by combining a multi-attention Transformer (MAT) with an adaptive superpixel segmentation-based AL (ASSAL) technique. The MAT model leverages self-attention and outlook-attention methods to extract both long-range and local contextual variables from input data. To address the challenge of limited labeled samples, the ASSAL technique utilizes an adaptive superpixel segmentation algorithm to identify informative samples for AL. This approach significantly reduces the number of labeled samples required for training while maintaining high classification accuracy.

Moreover, in general, the Swin Transformer (ST) showcases significant strengths, primarily deriving from its hierarchical attention mechanism, which enables the model to efficiently capture information across different scales, analyzing both local and global features \citep{9868046, ZU2024108041, rs15153721}. Furthermore, the utilization of windowing-based processing enhances the scalability of STs in contrast to traditional Transformers, facilitating the effective handling of large images with reduced computational complexity \citep{FAROOQUE2023107070}. The adaptability of the architecture to diverse tasks makes STs suitable for various applications \citep{rs15143491}. In the domain of HSC, STs have demonstrated state-of-the-art (SOTA) performance, outperforming traditional Transformers and CNNs in specific scenarios, underscoring their efficacy for HSC tasks \citep{Selen2105668, rs15102696}. However, despite these achievements, STs have limitations. While excelling in capturing spatial relationships, they may encounter challenges in dealing with sequential data, rendering tasks reliant on spectral dependencies less optimal for them \citep{10123084}. Additionally, the hierarchical attention mechanism introduces added complexity. Training large ST models necessitates substantial computational resources, posing challenges for researchers with limited access to high-performance computing. Moreover, akin to other deep models, the interpretability of STs raises concerns, particularly in complex tasks like HSIC, necessitating ongoing research to understand their decision-making processes \citep{rs15153721}.

Likewise, the vision and spatial-spectral Transformers (SSTs) stand out in their ability to capture global contextual information \citep{10423821, 10409287, hou2024linearly}. The self-attention mechanism enables the model to consider relationships between all HSI regions simultaneously, providing a comprehensive understanding of the visual context \citep{9895238}. Unlike CNNs, SSTs demonstrate robust scalability to high-resolution HISs, effectively processing large HSI datasets without the need for complex pooling operations. This versatility has led to successful applications of SSTs in HSC, and their adaptable architecture contributes to their broad applicability. Moreover, SSTs reduce reliance on handcrafted features by directly learning hierarchical representations from raw pixel values \citep{9627165}, simplifying the model-building process and often yielding improved performance. Additionally, the attention maps generated by SSTs offer insights into the model's decision-making process, enhancing interpretability by highlighting the image regions most influential in specific predictions \citep{10387571}.

Despite the advancements, SSTs do come with certain limitations. Notably, training large SSTs can pose significant computational challenges, particularly as the model size increases \citep{10419133, 10379170}. The self-attention mechanism introduces quadratic complexity with respect to sequence length, potentially impeding scalability \citep{HUANG2024109897}. Unlike CNNs, which inherently possess translation invariance through shared weight convolutional filters, SSTs may struggle to capture spatial relationships that remain invariant to small translations in the input \citep{SUN2024102163}. Moreover, SSTs rely on dividing input images into fixed-size patches during the tokenization process, which may not efficiently capture fine-grained details \citep{10418237, 10400415}. The quadratic scaling of self-attention poses challenges, particularly when dealing with long sequences. Additionally, achieving optimal performance with SSTs often requires substantial amounts of training data. Attempting to train these models on smaller datasets may lead to overfitting, thus limiting their effectiveness in scenarios with limited labeled data \citep{10399798}.

Considering these limitations, various potential solutions have been proposed in the literature, including the integration of SSTs with other architectures. For instance, a hybrid model could leverage the strengths of both approaches by combining the global context understanding of SSTs with the local feature extraction capabilities of CNNs \citep{10415455, 10400402}. However, a potential drawback of hybrid Transformers is the increased complexity in model architecture. Integrating different Transformer variants or combining Transformers with other architectures might introduce additional intricacy, rendering the model more challenging to interpret and potentially demanding more computational resources for training and inference.

Moreover, there exists a realm for exploration in fine-tuning the attention mechanism, where endeavors concentrate on sparse attention patterns or alternative attention mechanisms aimed at tackling the computational complexity challenges \citep{SHU2024107351, MA2024102148, rs16020404}. Capturing long-range dependencies and integrating global and local features in Transformers entails the challenge of finding a delicate equilibrium between model complexity and computational efficiency. Although optimizing attention patterns can elevate performance, it may concurrently heighten computational requirements, rendering the model more demanding on resources and possibly constraining its utility in scenarios with limited resources.

Expanding the capabilities of SSTs to accommodate multi-modal inputs emerges as a promising avenue for further research \citep{10379015}. This extension has the potential to widen their applicability and augment their ability to comprehend intricate relationships within diverse data types \citep{10382626, Miao22970, liu2021modality}. However, integrating information from various modalities necessitates thoughtful consideration of feature representations and alignment. The model might encounter difficulties in effectively learning meaningful cross-modal relationships, and crafting architectures capable of efficiently fusing and processing diverse data sources remains an ongoing research challenge. Additionally, gathering and annotating large-scale multi-modal datasets for training can be resource-intensive. Lastly, developing efficient tokenization strategies is crucial, aiming to capture fine-grained details while alleviating the quadratic scaling issue. Such strategies could significantly boost the performance of SSTs in tasks requiring high spatial resolution. In summary, while both ST and SST are Transformer-based architectures tailored for image classification tasks, they diverge in their approach to handling image data. Here are some of the major differences between ST and SST:

\textbf{Image Patch Processing:} ST adopts a hierarchical structure by initially dividing the image into non-overlapping patches. It then processes these patches hierarchically using a window-based self-attention mechanism, efficiently capturing both local and global context. In contrast, SST divides the input image into fixed-size non-overlapping patches, linearly embeds each patch, and subsequently flattens the 2D spatial information into a 1D sequence for input into the Transformer.

\textbf{Hierarchical Self-Attention:} ST introduces a shifted window-based self-attention mechanism, where instead of attending to all positions equally, it employs local self-attention windows that slide across the sequence hierarchically. This approach enables the model to capture both local and long-range dependencies effectively. On the other hand, SST utilizes a single self-attention mechanism across the entire sequence of patches.

\textbf{Positional Encoding:} In ST, shifted windows in self-attention are utilized, eliminating the need for explicit positional embeddings. The local windows implicitly encode positional relationships. Conversely, SST relies on positional embeddings to furnish the model with information regarding the spatial arrangement of patches.

The Transformer-based models as compared to the traditional DL models for HSC are a nuanced exploration that delves into the strengths and weaknesses of each approach. HSC is a critical task in remote sensing, where the goal is to assign each pixel in an image to a specific land cover class based on the spectral information acquired across numerous bands. Traditional DL models, especially CNNs, have been the workhorse for HSC. CNNs are adept at learning hierarchical spatial features from the data, enabling them to capture intricate patterns and relationships. However, HS data introduces unique challenges due to the high dimensionality of the spectral information. Traditional models may struggle to effectively handle the complex interactions between different spectral bands and to capture long-range dependencies. Whereas, Transformer-based models were originally designed for natural language processing tasks. Transformers, with their attention mechanism, can capture global dependencies and model interactions between distant spectral bands effectively. This mechanism allows the model to assign varying levels of importance to different parts of the HS spectrum, enabling more nuanced feature learning.

One of the primary advantages of Transformer-based models in HSC lies in their capacity to handle long-range dependencies. The self-attention mechanism allows the model to weigh the importance of different spectral bands dynamically, facilitating the learning of complex spectral-spatial relationships. This can be particularly beneficial in scenarios where understanding the interactions between distant bands is crucial for accurate classification. However, the success of Transformer-based models in HSC is contingent on several factors. One significant consideration is the amount of available training data. Transformers often require large datasets to generalize well, and HS data can be inherently limited due to acquisition costs and data availability. Conventional DL models may still perform admirably when faced with smaller datasets.

Computational resources also play a role in this comparison. Transformer-based models, with their self-attention mechanisms, are computationally more demanding compared to traditional CNNs. The feasibility of deploying Transformer-based models may be restricted by the available hardware and processing capabilities. Moreover, the complexity of the classification task influences the performance of these models. In cases where the spectral patterns are relatively simple, traditional models may achieve satisfactory results without the need for the more complex architecture of Transformers. In short, the comparison of Transformer-based models with traditional DL models for HSC is a multifaceted analysis. While Transformer-based models hold promise in capturing global dependencies and intricate spectral-spatial relationships, their effectiveness depends on factors such as data size, computational resources, and the complexity of the classification task. Conventional DL models continue to be valuable and efficient, especially in scenarios with limited data and less intricate spectral patterns. The choice between these approaches should be made judiciously, considering the specific characteristics of the HS data and the goals of the classification task.

%%%%%%%%%%%%%%%%%%%%%%%%%%
\subsection{\textbf{Benefits and Challenges of Transformers}}

The application of Transformers in HSI analysis introduces a range of benefits and challenges, reflecting the unique characteristics of HS data and the capabilities of Transformer-based architectures. Here we first discuss the Benefits of Transformers for HSI analysis, later the drawbacks will be portrayed. 

\textbf{Global Contextual Understanding:} Transformers excel at capturing global dependencies and long-range interactions within data. In HSI analysis, this ability is crucial for understanding the complex relationships between different spectral bands. Transformers can consider the entire spectrum simultaneously, allowing for a more comprehensive contextual understanding of the HS scene. \textbf{Adaptive Feature Learning:} The attention mechanism in Transformers enables adaptive feature learning, allowing the model to assign varying levels of importance to different spectral bands. This adaptability is particularly valuable in HS data, where certain bands may contain more discriminative information for specific classes or features. Transformers can dynamically focus on relevant bands, enhancing their ability to extract meaningful features. \textbf{Efficient Handling of Spectral-Spatial Information:} Transformers are well-suited for capturing spectral-spatial interactions in HSIs. The self-attention mechanism facilitates the modeling of relationships between distant pixels, enabling the identification of intricate patterns and structures. This is especially beneficial in scenarios where spatial context is essential for accurate classification. \textbf{Transferability Across Tasks:} Pre-trained Transformer models on large datasets for general computer vision tasks can be fine-tuned for HSI analysis. The transferability of pre-trained models reduces the need for extensive HS-specific training data and can lead to improved performance in scenarios with limited labeled examples.

While Transformers offer several benefits for HS analysis, there are also some challenges associated with their use. For instance: \textbf{Computational Demands:} Transformers are computationally more demanding than conventional DL models, such as CNNs, RNN, AEs, etc. The self-attention mechanism involves processing interactions between all pairs of pixels, leading to a quadratic increase in computational complexity with the size of the input. This can pose challenges in terms of training time and hardware requirements, particularly for large HSIs. \textbf{Limited Data Availability:} HS datasets are often limited in size due to the high cost and complexity of data acquisition. Transformers, known for their data-hungry nature, may struggle to generalize effectively when faced with insufficient training samples. This limitation could lead to overfitting or suboptimal performance, especially in the absence of extensive HS-specific pre-training data. \textbf{Model Interpretability:} Transformers, being complex models with numerous parameters, might lack interpretability compared to simpler models. Understanding how the model arrives at specific classification decisions in HSI analysis is essential, especially in applications where interpretability is critical, such as in scientific research or decision-making processes. \textbf{Hyperparameter Sensitivity:} Transformers often involve a large number of hyperparameters, and their performance can be sensitive to these choices. Fine-tuning the hyperparameters for optimal performance on HS data may require extensive experimentation, and the optimal configuration might vary depending on the specific characteristics of the dataset.

In short, the use of Transformers in HSI analysis brings notable advantages in terms of global contextual understanding, adaptive feature learning, and efficient handling of spectral-spatial information. However, challenges related to computational demands, limited data availability, model interpretability, and hyperparameter sensitivity must be carefully addressed to harness the full potential of Transformer-based architectures in HS applications. Researchers and practitioners need to strike a balance between the benefits and challenges, considering the specific requirements and constraints of their HSI analysis tasks.

%%%%%%%%%%%%%%%%%%%%%%%%%%
\subsection{\textbf{Emerging Trends and Advancements}}

As HS technology continues to evolve, several emerging trends and advancements in HSC have garnered attention within the remote sensing and image-processing communities. These trends reflect ongoing efforts to enhance the accuracy, efficiency, and applicability of HS data analysis. The integration of DL and Transformer-based models has revolutionized HSC. DL architectures excel at automatically learning hierarchical and complex features from HS data, improving classification accuracy. Transformers, with their ability to capture global dependencies, are being explored to address challenges related to long-range spectral interactions. In the recent past, several trends have emerged, aiming to enhance the overall generalization performance of DL models, and a few of them are listed in the following.

%%%%%%%%%%%%%%%%%%%%%%%%%%
\subsubsection{\textbf{Domain Adaptation and Transfer Learning}} With the limited availability of labeled HS datasets, domain adaptation, and transfer learning have emerged as key strategies to leverage pre-trained models from other domains or modalities. This approach aids in addressing the data scarcity issue by transferring knowledge learned from one dataset to another, enhancing classification performance even with limited labeled HS samples. Depending on the presence of labeled training instances, transfer learning frameworks can be classified into supervised or unsupervised transfer learning. Typically, it is assumed that both the source and target domains are related but not identical. In scenarios like HSC, where the categories of interest are consistent but data in the two domains may differ due to distinct acquisition circumstances, the distributions may not be the same.

In DNN-based HSC, the model undergoes hierarchical feature learning, with lower layers typically extracting generic features when trained on diverse images. Consequently, the features acquired by these lower layers can be transferred to train a new classifier for the target dataset. For example, Yang et al. \citep{yang2017learning} employed a two-branch spectral-spatial CNN model, initially trained on a substantial amount of data from other HSIs. The lower layers of this pre-trained model were then applied to the target network, ensuring robust classification of the target HSI. To capture target-specific features, the higher layers of the target network were randomly initialized, and the entire network underwent fine-tuning using limited labeled instances from the target HSI. Similarly, Windrim et al. \citep{windrim2018pretraining} introduced an effective method for pre-training and fine-tuning a CNN network, making it adaptable for the classification of new HSIs. The study conducted by Liu et al. \citep{liu2018hyperspectral} incorporated both data augmentation and transfer learning approaches to address the shortage of training data, aiming to enhance the performance of HSC.

As discussed above, data in the source and target domains may differ in various aspects. For instance, in the context of HSIs, the dimensions of two HSIs may vary due to acquisition from different sensors. Addressing such cross-domain variations and facilitating knowledge transfer between them is referred to as heterogeneous transfer learning, and a comprehensive survey of such methods can be found in \citep{day2017survey}. Within the literature on HSC, numerous studies have been put forth to bridge the gap and transfer knowledge between two HSIs with varying dimensions and/or distributions. For instance, an effective framework for HSC based on heterogeneous transfer learning was introduced by Lin et al. \citep{li2017iterative}. This framework demonstrates efficacy with both homogeneous and heterogeneous HSIs. Another approach, proposed by Li et al. \citep{liu2020transfer}, utilizes an iterative re-weighting mechanism within a heterogeneous transfer learning framework for HSC. A work proposed by Liu et al. \citep{lin2017structure} advocates a band selection-based transfer learning approach to pre-train a CNN, maintaining a consistent number of dimensions across various HSIs. Additionally, Lin et al. put forward an unsupervised transfer learning technique designed to classify entirely unknown target HSIs. Furthermore, Pires et al. \citep{pires2020convolutional} demonstrate that networks trained on natural images can enhance the performance of transfer learning for remote sensing data classification compared to networks trained from scratch using smaller HS datasets.

%%%%%%%%%%%%%%%%%%%%%%%%%%
\subsubsection{\textbf{Data Augmentation}} Augmentation is a proven and effective technique in HSC to address the challenge of limited training samples. It involves generating new samples from the original training data without the need for additional labeling costs. Data augmentation approaches can be broadly categorized into two main strategies:  \textbf{Data Wrapping:} This strategy focuses on encoding invariances such as translational, size, viewpoint, and illumination variations by applying geometric and color-based transformations while preserving the original labels. These transformations help create new samples that exhibit similar characteristics to the original data but with variations that can improve the model's robustness and generalization. Examples of data wrapping techniques include geometric transformations like rotation, scaling, and flipping, as well as color-based transformations like brightness and contrast adjustments. \textbf{Oversampling:} Oversampling-based augmentation methods aim to inflate the training data by generating synthetic samples based on the original data distribution \citep{shorten2019survey}. These techniques help address class imbalance issues and provide the model with a more balanced representation of different classes. Oversampling techniques include mixture-based instance generation, where new samples are created by blending existing samples from the same class, and feature space augmentations, where new samples are generated by perturbing the feature vectors of existing samples.

In the realm of HSC literature, various frameworks utilizing data augmentation techniques have been implemented to enhance classification performance and mitigate potential overfitting issues, often stemming from limited training data availability. For example, Yu et al. \citep{yu2017deep} improved training data by applying three augmentation operations (flip, rotate, and translation), leveraging this augmented dataset to train CNN for HSC. Similarly, in another comparative study \citep{rochac2019data}, a data augmentation-based CNN exhibited a 10\% increase in HSC accuracy compared to a PCA-based CNN model.

The aforementioned approaches employ offline data augmentation methods, which involve increasing the training data by generating new instances during or before the model training process. A recent innovative data augmentation framework for HSI is introduced in \citep{nalepa2019training}. In contrast to expanding the training data, this framework generates samples at test time. A DNN trained on the original training data, coupled with a voting scheme, is utilized to determine the final class label. To enhance the generalization capability of DNN models, \citep{nalepa2019training} also puts forth two rapid data augmentation techniques for high-quality data synthesis. Additionally, a similar online data augmentation strategy based on PCA is proposed in \citep{nalepa2019hyperspectral}. This strategy synthesizes new instances during the inference phase rather than during the training phase.

%%%%%%%%%%%%%%%%%%%%%%%%%%
\subsubsection{\textbf{Generative Adversarial Networks}} The Generative Adversarial Network (GAN), introduced by Goodfellow et al. \citep{goodfellow2014generative}, consists of two neural networks: a generator and a discriminator. GANs can learn and replicate samples by leveraging intricate details within the data distribution. In the context of Semi-Supervised Learning-based HSC, Zhan et al. \citep{zhan2017semisupervised} presented a GAN approach that relies on spectral features. In a similar vein, He et al. \cite{he2017generative} introduced a GAN-based framework for spectral-spatial HSC. Likewise, Zhu et al. \citep{zhu2018generative} devised 1D-GAN and 3D-GAN architectures, based on CNN, aimed at enhancing classification performance. Zhan et al. \cite{zhan2018semi} employed a customized 1D-GAN to generate spectral features, utilized by a CNN for subsequent feature extraction, and concluded the process with majority voting in HSC. Recently, Feng et al. \citep{feng2019classification} presented a Spatial-Spectral Multi-Class GAN (MSGAN) employing two generators to produce spatial and spectral information through multiple adversarial objectives. In addressing data imbalance issues for HSC, Zhong et al. \citep{zhong2019generative} proposed a semi-supervised model combining GAN with Conditional Random Fields (CRFs).

Similarly, Wang et al. \citep{wang2019caps} explored the Caps-TripleGAN model, which efficiently generates new samples through a 1D structure Triple GAN (TripleGAN) and classifies the generated HSI samples using the capsule network (CapsNet). Additionally, Xue et al. \citep{xue2019semi} suggested the adoption of a 3D CNN-based generator network and a 3D deep residual network-based discriminator network for HSC. To learn high-level contextual features, the combination of a capsule network and a Convolutional LSTM (ConvLSTM) based discriminator model was introduced by Wang et al. \citep{wang2020generative} in the context of HSC. The study conducted by Alipour et al. \citep{alipour2020structure} aimed to overcome the limitation of scarce training examples by employing a GAN model. In this approach, the discriminator's performance is enhanced through an auxiliary classifier, generating more structurally coherent virtual training samples. Additionally, Roy et al. \citep{roy2021generative} introduced a generative adversarial minority oversampling-based technique to augment model performance. This approach specifically addresses the persistent challenge of class-wise data imbalance in HSC.

%%%%%%%%%%%%%%%%%%%%%%%%%%
\subsubsection{\textbf{Active Learning (AL)}} Given the cost and effort associated with acquiring labeled HS data, AL strategies are becoming more prevalent. These techniques involve selecting the most informative samples for annotation, iteratively improving the model's performance with minimal labeled data. AL helps in optimizing the learning process, making HSC more efficient and cost-effective. The process of selecting the most informative samples is performed to ensure that the chosen samples are both informative and representative of the overall input distribution, ultimately enhancing accuracy. Depending on the criteria for adding new instances to the training set, AL frameworks can be categorized as either stream-based or pool-based. In the stream-based selection, one instance is drawn at a time from an existing set of unlabeled samples, and the model determines whether to label it or not based on its perceived usefulness. Conversely, in the pool-based strategy, samples are queried from a pool or subset of unlabeled data. This selection is based on ranking scores computed from various measures to assess the sample's usefulness. The study by Ganti et al. \citep{ganti2012upal} observed that stream-based selection yields lower learning rates compared to pool-based selection, primarily due to the former's tendency to query additional instances. In pool-based selection, emphasizing diversity within the pool of samples is crucial to prevent redundancy. Typically, the selection or querying of the most valuable samples centers around three main aspects: heterogeneity behavior, model performance, and representativeness of samples. 

\textbf{Heterogeneity-based selection:} These strategies focus on selecting samples to add back to the original training set that exhibits greater heterogeneity compared to the existing training samples. This heterogeneity is assessed in terms of model diversity, classification uncertainty, and discord among a committee of diverse classifiers. Models based on heterogeneity include uncertainty sampling, expected model change, and query-by-committee. 

\textbf{Performance-based Selection:} These methods take into account the impact of adding queried samples on the model's performance. They aim to optimize the model's performance by minimizing variance and error. Performance-based sampling can be broadly categorized into two types i.e., Expected Error Reduction \citep{aggarwal2014active} and Expected Variance Reduction \citep{settles2009active}. 

\textbf{Representativeness-based selection:} Representative sampling tends to query instances that are representative of the overall input distribution, thus steering clear of outliers and unrepresentative samples. These methods assign a higher importance to the denser input region during the querying process. Representativeness sampling approaches, such as density-weighted techniques like information density, consider both the representativeness of samples and heterogeneity behavior. These models are often referred to as hybrid models \citep{aggarwal2014active}.

AL has gained substantial prominence in recent times within HSC. Liu et al. \citep{liu2017feature} introduced a feature-driven AL framework, aiming to establish a well-constructed feature space for HSC. Zhang et al. \citep{zhang2019active} proposed a semi-supervised AL method based on Random Forest, leveraging spectral-spatial features to formulate a query function. This function is employed to select the most informative samples as target candidates for the training set. Guo et al. \citep{guo2016superpixel} introduced a framework that integrates the spectral and spatial features of superpixels. Likewise, Xue et al. \citep{xue2018active} incorporated neighborhood and superpixel information to augment the uncertainty of queried samples. In a recent study, Bhardwaj et al. \citep{bhardwaj2020spectral} leveraged attribute profiles to integrate spatial information within an AL-based HSC framework.

The diversity of samples becomes crucial in batch-mode techniques to prevent redundancy. A multi-criteria batch-mode method introduced by Patra et al. \citep{patra2017spectral} defines a novel query function based on diversity, uncertainty, and cluster assumption measures. These criteria leverage the properties of KNN, SVM, and K-means clustering, respectively. Genetic algorithms are subsequently employed to choose the most effective batch of samples. Similarly, Zhang et al. \citep{zhang2017batch} proposed a regularized multi-metric batch-mode framework for HSC, exploiting various features of HSI. A Multiview AL (MVAL) framework was introduced by Xu et al. \citep{xu2017multiview}. This framework analyzes objects from various perspectives and gauges sample informativeness through multiview Intensity-based query criteria. Similarly, Pradhan et al. \citep{pradhan2018fisher} embraced the concept of multiview learning, employing the Fisher Discriminant Ratio to generate multiple views. In another study, Zhang et al. \citep{zhang2019adaptive} put forth an innovative adaptive MVAL framework for HSC, simultaneously leveraging spatial and spectral features in each view. More recently, Li et al. \citep{li2020subpixel} proposed an MVAL technique utilizing pixel-level, subpixel-level, and superpixel-level details to generate multiple views for HSC. Furthermore, the proposed method exploits joint posterior probability estimation and dissimilarities among multiple views to query representative samples.

Sun et al. \citep{sun2016active} combined an AE with an AL technique, while Liu et al. \citep{liu2016active} introduced a DBN-based AL framework for HSC. Similarly, Haut et al. \citep{haut2018active} married a Bayesian CNN with the AL paradigm for spectral-spatial HSC. More recently, Cao et al. \citep{cao2020hyperspectral} proposed a CNN-based AL framework to leverage unlabeled samples for HSC effectively. Several studies have incorporated AL with transfer learning for HSC \citep{ahmad2022disjoint}. For instance, Lin et al. \citep{lin2018active} introduced a framework that identifies salient samples and utilizes high-level features to establish correlations between source and target domain data. Another approach, suggested by Deng et al. \citep{deng2018AT}, presented an Active Transfer Learning technique based on Stacked Sparse AE. This technique jointly utilizes both spectral and spatial features for HSC. Additionally, Deng et al. \citep{deng2018active} combined domain adaptation and AL methods, incorporating multiple kernels for HSC.

AL in HSC introduces sophisticated frameworks aimed at improving the generalization capabilities of models. For example, Ahmad et al. \citep{Ahmad2018} presented a fuzziness-based to enhance the generalization performance of both discriminative and generative classifiers. This method calculates the fuzziness-based distance for each instance and the estimated class boundary. Instances with higher fuzziness values and smaller distances from class boundaries are selected as candidates for the training set. Recently, a non-randomized spectral-spatial framework for multiclass HSC was introduced in \citep{articleAhmad}. This framework combines the spatial prior fuzziness approach with Multinomial Logistic Regression via a Splitting and Augmented Lagrangian classifier. The authors conducted a comprehensive comparison of the proposed framework with state-of-the-art sample selection methods and diverse classifiers.

%%%%%%%%%%%%%%%%%%%%%%%%%%
\subsubsection{\textbf{Explainable AI for HS Analysis}} As HS data finds applications in critical decision-making processes, the demand for explainable AI has grown. Researchers are working on developing models that provide interpretable results, enabling end-users to understand the reasoning behind classification decisions. This is especially crucial in fields such as environmental monitoring, agriculture, and defense.

%%%%%%%%%%%%%%%%%%%%%%%%%%
\subsubsection{\textbf{Integration of Multimodal Data}} Combining HS data with other imaging modalities, such as Light Detection and Ranging (LiDAR) or Synthetic Aperture Radar (SAR), is an emerging trend. The fusion of multimodal data sources enhances the overall understanding of the scene \citep{hong2023decoupled}, providing complementary information for improved classification accuracy and discrimination of complex land cover classes.

In short, multimodal data refers to the combination of various types of data (e.g., spectral, spatial, textual, and contextual) that can provide complementary information for improving the accuracy and robustness of HSC. In the context of HSC, multimodal data can encompass several modalities, including: \textbf{Spectral Data:} Captured through HS sensors, this modality provides continuous spectral profiles for each pixel, enabling the identification of unique spectral signatures associated with different materials. \textbf{Spatial Data:} Information about the spatial arrangement of pixels, including texture and contextual relationships, can significantly enhance classification accuracy. Spatial information can be derived from multi-resolution imagery, such as RGB or multispectral images. \textbf{Textual Data:} Descriptive textual information about the scene, such as land cover type, can be integrated into the classification framework. This data may come from ancillary sources like metadata, environmental reports, or expert annotations. \textbf{Contextual Data:} Additional contextual features, such as elevation, climate conditions, or geographical information, can provide valuable insights into the classification process, especially in heterogeneous landscapes.

Integrating multimodal data offers several advantages for HSC: \textbf{Enhanced Discriminative Power:} By combining spectral and spatial information, classifiers can leverage the complementary strengths of each modality, leading to improved separability of classes. \textbf{Robustness to Noise:} Multimodal integration can mitigate the effects of noise present in HS data, as different modalities can corroborate each other's information and reduce uncertainty in the classification output. \textbf{Improved Generalization:} Utilizing diverse data sources helps models generalize better across varying conditions, making them less prone to overfitting on training data. \textbf{Better Representation of Complex Scenes:} Multimodal approaches can capture the complexities of real-world environments more effectively, enabling better classification of mixed or ambiguous land cover types.

Several techniques exist for fusing multimodal data in HSC for instance, \textbf{Early Fusion:} In this approach, multimodal data is concatenated at the feature level before classification. For example, spectral features from HS data can be combined with texture features extracted from spatial data. \textbf{Late Fusion:} This method involves separately classifying each modality and then combining the resulting classification maps. Techniques such as voting schemes or probability averaging can be applied to merge the outputs. \textbf{Hybrid Fusion:} A combination of early and late fusion strategies, hybrid approaches aim to leverage the strengths of both methods. For instance, features can be extracted from each modality and then combined with late-stage classification outputs. \textbf{Deep Learning Approaches:} Recent advancements in DL have enabled the development of multimodal neural networks capable of learning joint representations from different modalities. For instance, CNNs can process spatial data, while RNNs can handle sequential textual data. \textbf{Feature-Level Fusion:} Techniques like early fusion concatenate features from multiple modalities at the input stage, allowing for a more holistic representation of the data, while late fusion combines predictions from separate models trained on individual modalities. \textbf{Attention Mechanisms:} Recent approaches incorporate attention mechanisms that dynamically weigh the importance of various modalities based on the classification task, enhancing the model’s ability to focus on relevant features. \textbf{Graph-Based Methods:} Innovations in graph-based methods allow for the modeling of relationships between different data modalities, which can improve the robustness of classification results in complex environments.

Despite its advantages, multimodal data integration presents several challenges: \textbf{Data Alignment:} Ensuring spatial and spectral data are aligned correctly is crucial. Misalignment can lead to inaccurate classifications and degraded performance. \textbf{Dimensionality Management:} The integration of multiple data sources can result in significantly high dimensionality, leading to computational challenges and the curse of dimensionality. Effective dimensionality reduction techniques, such as PCA or t-SNE, may be necessary. \textbf{Model Complexity:} Incorporating multiple modalities often requires more sophisticated models, which can increase the complexity of training and inference processes. \textbf{Data Availability:} In some cases, obtaining multimodal data can be challenging, especially if certain modalities are expensive or difficult to capture.

%%%%%%%%%%%%%%%%%%%%%%%%%%
\subsubsection{\textbf{Edge Computing for Real-time Processing}} The need for real-time HSC in applications like precision agriculture and disaster response has led to the exploration of edge computing solutions. Deploying lightweight models on edge devices allows for on-site processing, reducing latency and enabling timely decision-making.

%%%%%%%%%%%%%%%%%%%%%%%%%%
\subsubsection{\textbf{Development of Benchmark Datasets}} Efforts are being made to create standardized benchmark datasets for HSC, facilitating fair comparisons among different algorithms and approaches. These datasets help in evaluating the generalization capabilities of models and advancing the reproducibility of research findings.

In short, the field of HSC is witnessing rapid advancements driven by innovations in DL, domain adaptation, graph-based methods, AL, explainable AI, multimodal data integration, edge computing, and the establishment of benchmark datasets \citep{hong2023cross}. These trends collectively contribute to the refinement of HS data analysis techniques and expand the applicability of HSI in various domains. As technology continues to progress, further breakthroughs will likely shape the future landscape of HSC.

%%%%%%%%%%%%%%%%%%%%%%%%%%
\section{\textbf{Explainable AI and Interoperability}}
\label{EXI}

Explainable AI (EAI) and interpretability play crucial roles in HSC, enabling users to understand and trust the decision-making process of the models \citep{10282341,  ZHANG2023100491, hong2024multimodal}. Here is a detailed discussion on EAI and interpretability in this context:

\begin{enumerate}
    \item \textbf{Importance of EAI:} In HSC, where accurate and reliable decisions are vital, it is essential to understand why and how a model arrives at a particular classification result. EAI techniques aim to provide transparency and interpretability to the decision-making process of complex models. By gaining insights into the model's reasoning, users can assess the model's reliability, detect biases, identify important features, and diagnose potential errors or limitations.

    \item \textbf{Interpretability Techniques:} Various techniques are employed to achieve interpretability in HSC \citep{10416691, 10499826}: 
    \begin{itemize}
        \item \textbf{Feature Importance:} Feature importance methods, such as permutation importance \citep{PUTKIRANTA2024114175} or feature saliency maps \citep{zhou2024probing}, help identify the most influential spectral bands or spatial locations contributing to the classification decision. These methods provide insights into which features are critical for classification and can guide further analysis or data collection.
        
        \item \textbf{Attention Mechanisms:} Attention mechanisms \citep{10543045, 10506764}, commonly used in Transformer-based models, allow users to visualize the attention weights assigned to different spectral bands or spatial locations. By highlighting the relevant regions, attention maps provide interpretability by revealing which parts of the HSI are crucial for classification.
        
        \item \textbf{Rule Extraction:} Rule extraction techniques aim to extract human-understandable rules or decision trees from complex models \citep{CASTANO2024105903}. These rules provide explicit conditions that determine the classification outcome for a given input, making the decision process more transparent and interpretable.
        
        \item \textbf{Model Visualization:} Visualizing the internal representations of the model, such as intermediate feature maps or activation patterns, can aid in understanding how the model processes and represents the HS input data. Visualization techniques help identify patterns, correlations, or anomalies that contribute to the classification outcome.
    \end{itemize}
    
    \item \textbf{Domain Expert Collaboration:} In HSC, the collaboration between AI experts and domain experts is crucial to achieving interpretability \citep{10415093, 10416249}. Domain experts possess in-depth knowledge of the spectral signatures and spatial characteristics of the objects or materials of interest. By involving domain experts in the interpretability process, their expertise can be leveraged to validate and refine the explanations provided by the AI models, ensuring that the interpretations align with domain-specific knowledge.

    \item \textbf{Transparency and Trust:} EAI and interpretability techniques help build transparency and trust in HSC models \citep{10486908, 10499268, 10463296}. By providing understandable and justifiable explanations, users can assess the model's reliability, understand its limitations, and make informed decisions based on the model's outputs. This transparency is particularly important in critical applications, such as environmental monitoring, precision agriculture, or target detection, where the consequences of misclassifications can be significant.

    \item \textbf{Regulatory and Ethical Considerations:} In some domains, regulatory frameworks or ethical considerations may require explanations for AI-based decisions \citep{add6778}. For example, in healthcare or remote sensing applications, it may be necessary to explain the reasons behind a diagnosis or land cover classification. Incorporating interpretability techniques in HSC models can help meet requirements and ensure compliance with regulations and ethical standards by providing insights into how the model arrives at its decisions. Interpretability techniques offer transparency by explaining which features or spectral signatures are influential in the classification process. This transparency enables users to understand the model's reasoning, identify potential biases or errors, and ensure that decisions align with ethical and regulatory guidelines. Additionally, interpretability techniques facilitate model validation and verification, aiding in the assessment of model robustness and reliability.
\end{enumerate}

There are certain challenges as well, for instance:

\begin{enumerate}
    \item \textbf{Trade-off with Performance:} Achieving high interpretability may come with a trade-off in terms of model performance \citep{HANCHATE2023184}. Simpler models or interpretable methods might not capture the complexity of HS data as effectively as more complex models.

    \item \textbf{Interpretable DL:} DL models, particularly CNNs and Transformers, can be challenging to interpret due to their hierarchical and non-linear nature \citep{9165231, 10138912, li2023lrr}. Balancing interpretability with the power of DL is an ongoing research challenge. 

    \item  \textbf{Human Understanding:} The interpretability provided by a model should align with the human understanding of the data \citep{10430776}. It is essential to present information in a way that is meaningful and relevant to domain experts and end-users.
\end{enumerate}

In short, EAI and interpretability play a pivotal role in HSC, addressing issues of trust, ethics, and regulatory compliance. Choosing appropriate interpretability methods and balancing them with model performance and ethical considerations, identifying important features, detecting biases, collaborating with domain experts, and complying with regulations are essential for building trustworthy and effective HSC systems. Ongoing research in this area aims to develop more advanced techniques that strike an optimal balance between interpretability and predictive accuracy.

%%%%%%%%%%%%%%%%%%%%%%%%%%
\section{\textbf{Diffusion Models for Hyperspectral Imaging}}
\label{DFs}

Diffusion models for HS imaging are a class of computational techniques used to enhance and analyze HSIs \citep{Miao_2023_ICCV, Wu_2023_ICCV, 10234379, 10327767, 8481477, CAO2024102158, ZHANG2024123796}. Diffusion models aim to exploit the underlying spatial and spectral information present in HS data to improve image quality, denoise noisy images, and enhance valuable spectral features for various applications such as remote sensing, environmental monitoring, and object detection. Diffusion models are typically based on the principles of diffusion processes, which simulate the spreading and smoothing of information within an image. These models leverage the notion that neighboring pixels in an image are likely to have similar spectral characteristics. By propagating information across neighboring pixels, diffusion models can effectively enhance the spatial and spectral coherence of HSIs.

One popular diffusion model used in HS imaging is the Perona-Malik (PM) diffusion model \citep{9323535, 9516067, 5416595}. The PM model is based on the heat equation and aims to remove noise while preserving meaningful image structures. It achieves this by diffusing image gradients according to a diffusion coefficient that is adaptively modulated by the local image characteristics. The PM model has been successfully applied to HS denoising tasks, effectively reducing noise while preserving important spectral features.

Another widely used diffusion model is the Total Variation (TV) model \citep{8725896, 9893196, 8481477, 8463507, 8233403}. The TV model seeks to denoise HSIs by minimizing the TV of the image, which measures the sum of the absolute differences between neighboring pixel values. By promoting piecewise smoothness in the image, the TV model can effectively suppress noise while preserving sharp edges and fine details.

In addition to denoising, diffusion models can also be employed for other tasks in HS imaging, such as image fusion and feature extraction \citep{10234379, 10179942, 10491382, zhou2024general}. For image fusion, diffusion-based methods can be used to integrate information from multiple HSIs or different modalities, thereby enhancing the overall image quality and reducing noise \citep{10268922, 5416595}. In terms of feature extraction, diffusion models can be employed to enhance discriminative spectral features, making them more distinguishable for subsequent classification or target detection tasks \citep{wu2019orsim,10496287, 9553196, 10268922,wu2022uiu}.

It is worth noting that while diffusion models have shown promising results in HS imaging, they also come with certain limitations. These models may struggle with preserving fine details and subtle spectral variations, as excessive diffusion can lead to over-smoothing \citep{NEURIPS2023_d51e2a46}. Additionally, diffusion models can be computationally intensive, especially when applied to large-scale HS datasets, requiring efficient algorithms and parallel computing techniques to ensure real-time or near-real-time processing. In summary, diffusion models provide valuable tools for enhancing HSIs by leveraging spatial and spectral information. These models, such as the PM and TV models, offer effective denoising, image fusion, and feature extraction capabilities. However, careful parameter tuning and algorithm optimization are necessary to strike a balance between noise suppression and preservation of important spectral details. Ongoing research continues to explore and refine diffusion models for HS imaging to address their limitations and expand their applicability in various domains.

Diffusion models are relevant to discuss here for several reasons, for instance, \textbf{Preprocessing for Classification:} Diffusion models can play a crucial role in preprocessing HSIs before classification. Enhancing image quality, denoising, and reducing noise can improve the performance of subsequent classification algorithms. \textbf{Feature Extraction:} Diffusion models can also be employed for feature extraction as discussed above. By enhancing discriminative spectral features and reducing noise, diffusion models can improve the separability of different classes and enable more effective classification. \textbf{Noise Robustness:} HSIs often suffer from noise due to various sources such as atmospheric interference or sensor limitations. Diffusion models, such as the PM and TV models, are known for their noise-reduction capabilities.

%%%%%%%%%%%%%%%%%%%%%%%%%%
\section{State Space Model (Mamba)}
\label{SSM}

The Mamba model represents an innovative approach to HSC, leveraging the state space model (SSM) framework to process spectral and spatial information efficiently. Traditional DL models, such as CNNs and Transformers, have demonstrated strong performance in HSC by capturing spatial and spectral dependencies \citep{ahmad2024multihead}. However, these models often face challenges with computational complexity and scalability, especially when handling high-dimensional data typical of HSIs. The Mamba model addresses these limitations by utilizing the efficiency of state space operations, offering a powerful alternative to conventional methods.

The Mamba model is fundamentally an SSM adapted for HS data analysis. SSMs are known for their capacity to represent sequential data efficiently by capturing both spatial and temporal dependencies through a set of hidden states. In the context of HSIs, each pixel is represented as a sequence of spectral bands, and Mamba uses state space equations to model this sequence dynamically. The Mamba model is structured with key components: \textbf{Spectral-Spatial Token Generation:} Generates tokens representing both spatial and spectral features, offering a more compact representation of HS data. \textbf{State Transition Mechanism:} Efficiently encodes dependencies among spectral bands, capturing relationships with fewer computations than standard attention mechanisms used in Transformers. \textbf{Output Layer:} Provides a classification output by processing the state representations, which contain sufficient spectral-spatial information for accurate classification. The core of Mamba relies on state space equations, which are expressed as:

\begin{equation}
    x_{t+1} = Ax_t + Bu_t
\end{equation}
\begin{equation}
    y_t = Cx_t + Du_t
\end{equation}
where $x_t$ represents the state vector at time $t$, $u_t$ is the input vector (e.g., spectral feature), $A$ and $B$ are the transition matrices that determine state evolution, and $C$ and $D$ project the state to the output space. This representation enables Mamba to maintain a compact set of states for each spectral band, reducing the need for extensive parameters and computations, as opposed to the dense self-attention layers in Transformer-based models. 

The Mamba model offers several significant advantages over Transformers and other conventional models: \textbf{Reduced Computational Complexity:} The Mamba model’s use of state space representations inherently reduces computational requirements. Unlike Transformers, which rely on quadratic complexity in terms of sequence length due to self-attention, Mamba processes sequential data linearly. This is particularly beneficial for HSIs, which have long spectral sequences, making Mamba a faster and more resource-efficient choice. \textbf{Memory Efficiency:} Transformers often require extensive memory for handling large spectral sequences due to the self-attention mechanism’s need to store intermediate representations. The Mamba model, by contrast, uses fewer parameters and requires less memory, enabling it to handle larger HS datasets without sacrificing performance. \textbf{Ability to Capture Long-Range Dependencies:} While CNNs typically struggle with capturing long-range dependencies, the state space formulation in Mamba naturally encodes these dependencies through state transitions, allowing for effective learning of spectral-spatial relationships in HSIs. \textbf{Improved Scalability:} The model’s reduced complexity allows it to scale well with high-resolution HS datasets, which are often challenging for other models. Mamba’s efficient structure enables it to perform well even as the dimensionality of the input data grows, a crucial advantage when dealing with real-world HS data. \textbf{Noise Robustness:} HS data often contain noise, which can impair classification accuracy. The state transitions in Mamba smooth out such noise by maintaining continuity between states. This robustness is further enhanced by the token generation process, which aggregates spectral and spatial information, providing resilience to small variations in the input.

The MHSSMamba \citep{ahmad2024multihead} model enhances Spatial-Spectral Mamba by incorporating multi-head self-attention and token enhancement, effectively capturing complex spectral-spatial relationships and managing long-range dependencies in hyperspectral data. This approach improves computational efficiency while preserving critical contextual information across spectral bands. WaveMamba \citep{ahmad2024wavemamba} introduces an innovative integration of wavelet transformation with Spatial-Spectral Mamba architecture, enhancing hyperspectral image classification by capturing both local textures and global contexts. The Spatial-Spectral Morphological Mamba (MorpMamba) \citep{ahmad2024morphologicalmamba} model optimizes HSC by integrating morphological operations and an SSM for computational efficiency, addressing the limitations of Transformers. Through spatial-spectral token generation, feature enhancement, and multi-head self-attention, MorpMamba effectively fuses detailed structural and spectral information, improving classification and producing high-quality ground truth maps.

The MambaHSI \citep{10604894} model leverages Mamba's efficient long-range modeling capabilities, introducing spatial and spectral Mamba blocks to adaptively capture HSI data's spatial-spectral interactions. This approach, validated across four HSI datasets, demonstrates MambaHSI’s superiority in classification accuracy and highlights Mamba's potential as a next-generation backbone for HSI models. The Mamba-in-Mamba (MiM) \citep{ZHOU2025128751} architecture pioneers a new approach for HSC, utilizing an efficient Mamba-Cross-Scan (MCS) transformation and tokenized feature learning to overcome computational challenges. By integrating multi-scale loss and enhanced semantic tokenization, MiM offers robust performance, especially suited for remote sensing tasks requiring high precision with limited training data.

The HyperMamba model \citep{10720896} introduces a spectral-spatial adaptive Mamba for HSC, balancing spatial and spectral feature extraction efficiently. By employing Spatial Neighborhood Adaptive Scanning (SNAS) and Spectral Adaptive Enhancement Scanning (SAES), HyperMamba effectively captures intricate spatial-spectral dependencies, achieving state-of-the-art performance across multiple HSI datasets. The Local Enhanced Mamba (LE-Mamba) network \citep{WANG2024104092} offers an efficient solution for HSC by combining a Local Enhanced Spatial SSM (LES-S6) and Central Region Spectral SSM (CRS-S6) for precise spatial-spectral integration, while the Multi-Scale Convolutional Gated Unit (MSCGU) ensures robust feature aggregation, achieving strong non-causal feature extraction and enhanced classification performance.

The HLMamba \citep{10679212} model leverages a gradient joint algorithm (GJA) for edge contour extraction and a multimodal Mamba fusion module (MMFM) for efficient feature integration across HSI and LiDAR data, enhancing classification performance by capturing complex interrelationships and long-distance dependencies in multimodal data. The spatial-spectral interaction super-resolution CNN-Mamba fusion network \citep{10695805} enhances the resolution of low spatial resolution HSIs and high spatial resolution multispectral images through mutual guidance, ensuring effective computational efficiency in extracting both local and global features for improved image fusion.

In HSC, computational efficiency is crucial due to the high dimensionality of the data. The Mamba model demonstrates not only comparable or superior classification accuracy to Transformers and CNNs but also achieves this with lower computational cost. This makes Mamba a practical solution for resource-constrained applications, such as onboard processing in drones or real-time monitoring systems. Additionally, due to its compact state space representation, the model can be implemented in edge devices for field applications in agriculture, environmental monitoring, and disaster response, where power efficiency and computational efficiency are essential.

\subsection{Current Limitations of the Mamba Model:} 

\textbf{Scalability to High Dimensions:} While the Mamba model exhibits linear complexity compared to traditional Transformers, it may still face challenges when scaling to extremely high-dimensional HS data. As the number of spectral bands increases, the model's performance can degrade due to the increased complexity of capturing the intricate spectral relationships.

\textbf{Integration of Temporal Dynamics:} The original Mamba model was primarily designed for spatial and spectral interactions, which limits its capability to effectively incorporate temporal dynamics in HSI sequences. This is particularly relevant in applications involving dynamic changes in the environment, such as agriculture or urban development.

\textbf{Limited Robustness to Noisy Data:} Although the Mamba architecture demonstrates strong performance under controlled conditions, it may not be robust enough to handle noisy or corrupted HS data, which is common in real-world scenarios. This limitation can lead to suboptimal classification outcomes, especially in the presence of atmospheric disturbances or sensor artifacts.

\textbf{Feature Representation and Interpretation:} The model's representation of features, particularly in terms of interpretability, can be limited. While Mamba models capture complex relationships, understanding how these relationships contribute to final decisions remains a challenge, especially in critical applications where interpretability is essential.

\textbf{Heterogeneous Data Fusion:} Current Mamba implementations may struggle with the effective fusion of multimodal data (e.g., combining HSIs with LiDAR or RGB data). The model's architecture needs to be adapted to seamlessly integrate heterogeneous features while preserving important spatial and spectral information.

\subsection{Possible Future Research Directions}

\textbf{Enhanced Feature Representation Techniques:} Future research could focus on developing advanced feature extraction methods that explicitly address the high-dimensional nature of HS data. This could include the integration of hybrid architectures to enhance the model's ability to extract meaningful features.

\textbf{Temporal Mamba Extensions:} To address the limitations regarding temporal dynamics, researchers could explore the incorporation of temporal models into the Mamba framework. This could involve developing a spatiotemporal Mamba variant that effectively models changes over time, enabling applications in dynamic monitoring and analysis of environments.

\textbf{Robustness to Noisy Data:} Investigating noise-resistant training strategies, such as adversarial training or data augmentation techniques, can enhance the robustness of the Mamba model. Research could focus on developing specific regularization techniques tailored for HS data that improve classification performance in noisy conditions.

\textbf{Interpretable Mamba Architectures:} Efforts should be made to design interpretable Mamba architectures that can elucidate the reasoning behind classification decisions. Techniques such as attention visualization, feature importance mapping, and explainable AI frameworks can be integrated to improve transparency in model predictions.

\textbf{Multimodal and Heterogeneous Data Fusion:} Future work could investigate the extension of Mamba models to handle the integration of various types of data more effectively. This could involve developing new fusion strategies that can dynamically adapt to different data sources while maintaining the spatial-spectral integrity of HSIs.

%%%%%%%%%%%%%%%%%%%%%%%%%%
\section{\textbf{Open Challenges and Research Questions}}
\label{RQs}

HSC poses several challenges and prompts various research questions, reflecting the complexity and unique characteristics of HS data. Addressing these challenges and answering pertinent research questions is essential for advancing the field and improving the accuracy and applicability of HSC. Below is a detailed discussion of some key challenges and associated research questions:

%%%%%%%%%%%%%%%%%%%%%%%%%%
\begin{figure*}[!t]
    \centering
    \includegraphics[width=0.99\textwidth]{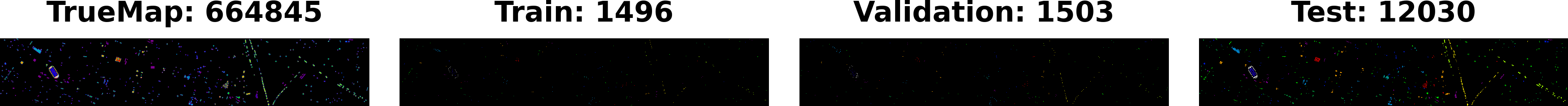}
    \caption{University of Houston ground truth maps, disjoint training, validation, and test ground truth maps.}
    \label{Fig16}
\end{figure*}
%%%%%%%%%%%%%%%%%%%%%%%%%%

\begin{enumerate}
    \item \textbf{Limited Labeled Training Data:} \textbf{Challenge:} Acquiring labeled training data for HSIs can be expensive and time-consuming. Limited labeled data may hinder the development of accurate and robust classification models \citep{10493163, 10379170}. \textbf{Research Questions:} How can transfer learning \citep{10445916} be effectively applied to leverage knowledge from related domains or existing labeled datasets to improve classification performance with limited labeled HS data? What AL strategies can be employed to strategically select and query the most informative samples, optimizing the use of limited labeled data?

    \item \textbf{Curse of Dimensionality:} \textbf{Challenge:} HS data typically exhibits high dimensionality due to the large number of spectral bands. This high-dimensional space can lead to the curse of dimensionality, affecting the performance of traditional classification algorithms \citep{hong2019learning}. \textbf{Research Questions:} How can dimensionality reduction techniques be tailored to HS data to preserve relevant information while mitigating the challenges associated with the curse of dimensionality? What DL architectures or AE can effectively capture and represent the intrinsic structure of HS data in a lower-dimensional space?

    \item \textbf{Spectral-Spatial Complexity:} \textbf{Challenge:} HSIs often contain intricate spectral-spatial patterns that are challenging to capture using traditional classifiers \citep{10387414, 10495370}. \textbf{Research Questions:} How can advanced machine learning models, such as DNNs, be designed to integrate spectral and spatial information for improved classification accuracy effectively? What role do graph-based models or attention mechanisms play in capturing spectral-spatial dependencies, and how can they be optimized for HSC?

    \item \textbf{Class Imbalance and Rare Events:} \textbf{Challenge:} Class imbalance is prevalent in HS datasets, where certain classes may have fewer instances. Additionally, rare events or anomalies may be crucial but challenging to detect \citep{10439239, 10453272}. \textbf{Research Questions:} How can HSI classifiers handle class imbalance, and what techniques can be employed to address the challenge of rare events or anomalies in the data?   What role can generative models, such as GANs, play in augmenting rare class samples and enhancing classification performance?

    \item \textbf{Robustness to Environmental Variability:} \textbf{Challenge:} HS data may be sensitive to variations in environmental conditions, such as lighting, atmospheric effects, and seasonal changes, leading to reduced model generalization \citep{YU2024120452}. \textbf{Research Questions:} How can HS classifiers be made robust to environmental variability, and what preprocessing or normalization techniques are effective in reducing the impact of these variations? What transfer learning strategies can be employed to adapt models trained in one environment to perform well in different environmental conditions?

    \item \textbf{Explainability and Interpretability:} \textbf{Challenge:} DL models and complex classifiers may lack interpretability, hindering user understanding and trust in the classification results \citep{10416691, 10499826}. \textbf{Research Questions:} What methods and tools can be developed to provide explainability and interpretability in HSC, especially for DL models? How can the interpretability of models be aligned with the specific needs and expectations of domain experts and end-users in different application domains?

    \item \textbf{Real-Time Processing and Resource Constraints:} \textbf{Challenge:} In certain applications, HSC needs to operate in real-time or under resource constraints, necessitating efficient algorithms \citep{10474407, 10381732, ZHANG2024122273, MARTINS2024104998}. \textbf{Research Questions:} How can lightweight and efficient models be designed for real-time HSC, considering the constraints of processing power, memory, and energy consumption? What trade-offs can be made between model complexity and accuracy to achieve optimal performance in resource-constrained environments?
\end{enumerate}

In short, addressing these challenges and answering the associated research questions is crucial for advancing the field of HSC. Innovative solutions to these issues will contribute to the development of more accurate, robust, and interpretable HSC models, enabling their successful application in various domains such as agriculture, environmental monitoring, and remote sensing.

%%%%%%%%%%%%%%%%%%%%%%%%%%
\section{\textbf{Experimental Results and Discussion}}
\label{Res}

Research-oriented publications in the literature commonly provide a thorough experimental evaluation to elucidate the strengths and weaknesses of the proposed methodologies. Nonetheless, these works might adopt varying experimental configurations, such as the percentage/number of training, validation, and test samples. While the number or percentage of samples in these sets might be the same, the actual samples could differ as they are often selected randomly. Therefore, ensuring a fair comparison among different works from the literature necessitates the adoption of consistent and identical experimental settings. The standardized experimental settings should entail identical samples, ensuring that geographical locations remain consistent for all selected models rather than varying between them. Furthermore, the number of samples chosen for each round of training within the cross-validation process must be uniform. Typically, these samples are selected randomly, creating a potential discrepancy if models are executed at different times, as they might use different sets of samples.

%%%%%%%%%%%%%%%%%%%%%%%%%%
\begin{table}[!t]
    \centering
    \caption{Summary of the HSI datasets used for experimental evaluation.}
    \resizebox{\columnwidth}{!}{\begin{tabular}{c|c|c|c} \hline 
        --- & \textbf{IP \cite{green1998imaging}} & \textbf{PU \cite{hong2020multimodal}} & \textbf{UH \cite{hong2021multimodal}} \\  \hline 
        \textbf{Download Link} & \href{https://www.ehu.eus/ccwintco/index.php/Hyperspectral_Remote_Sensing_Scenes}{Indian} & \href{https://www.ehu.eus/ccwintco/index.php/Hyperspectral_Remote_Sensing_Scenes}{Pavia} & \href{https://hyperspectral.ee.uh.edu/?page_id=459}{Houston} \\  
        \textbf{Year} & 1992 & 2001 & 2013 \\
        \textbf{Sensor} & AVIRIS & ROSIS-03 & CASI \\
        \textbf{Spatial} & $145\times 145$ & $610 \times 610$ & $340\times 1905$ \\
        \textbf{Spectral} & 220 & 115 & 144 \\
        \textbf{Wavelength} & $400-2500~nm$ & $430-860~nm$ & $350-1050~nm$ \\
        \textbf{Samples} & 21025 & 207400 & 1329690 \\
        \textbf{Classes} & 16 & 9 & 15\\
        % \textbf{Sensor} & Aerial & Aerial & Aerial \\
        \textbf{Resolution} & $20~m$ & $1.3~m$ & $2.5~mpp$ \\ \hline 
    \end{tabular}}
    \label{Tab.1}
\end{table}
%%%%%%%%%%%%%%%%%%%%%%%%%%

Another concern prevalent in recent literature is the overlap between training and test samples \citep{7762146}. In many cases, training and validation samples are randomly chosen (considering or disregarding the aforementioned point) for the training and validation phases. However, during the testing phase, the entire dataset is often passed through, leading to a highly biased model, as the model has already encountered the training samples, thereby inflating accuracy results. In this study, while the training and test samples are selected randomly (as all models are executed simultaneously), special attention has been given to the above issue. Specifically, measures have been taken to ensure that there is no overlap among these samples, preserving the integrity of the evaluation process. Table \ref{Tab.1} furnishes a concise overview of each dataset utilized in the subsequent experiments and further details regarding the experimental datasets are provided in the following.

The \textbf{University of Houston} (UH) dataset, released by the IEEE Geoscience and Remote Sensing Society as part of the Data Fusion Contest in 2013 \citep{hong2021multimodal}, was collected by the Compact Airborne Spectrographic Imager (CASI). With dimensions of $340 \times 1905$ pixels and 144 spectral bands, the dataset features a spatial resolution of $2.5$ meters per pixel (mpp) and a wavelength range from $0.38$ to $1.05$ $\mu$m. The ground truth for this dataset encompasses 15 distinct land-cover classes. Detailed class descriptions and ground truth maps are depicted in Figure \ref{Fig16}. 

\begin{figure}[!t]
    \centering
    \includegraphics[width=0.48\textwidth]{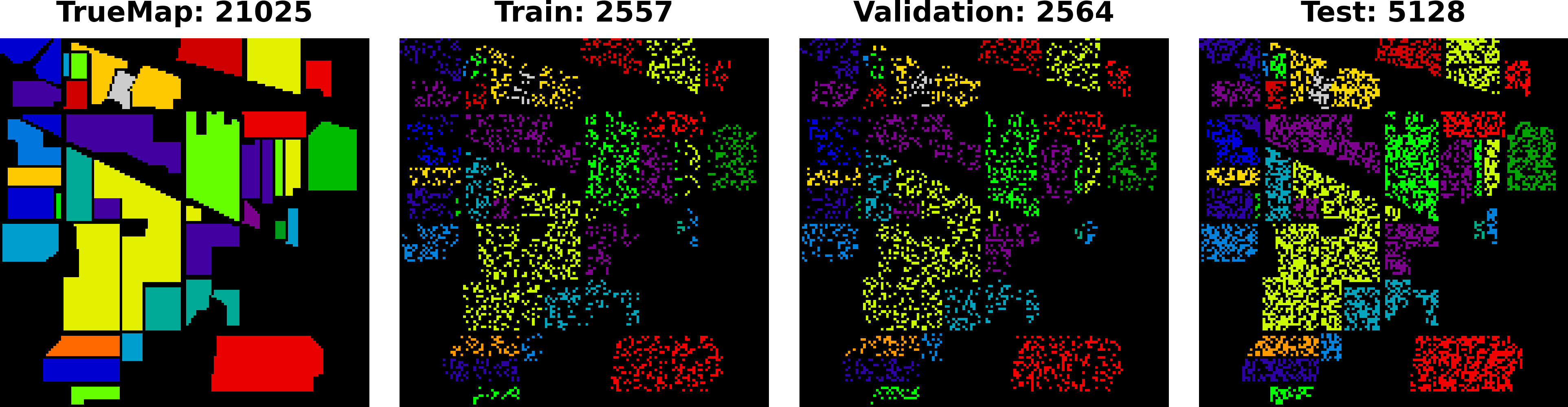}
    \caption{Indian Pines ground truth maps, disjoint training, validation, and test ground truth maps.}
    \label{Fig17}
\end{figure}

The \textbf{Indian Pines} (IP) dataset was acquired by the Airborne Visible/Infrared Imaging Spectrometer (AVIRIS) \citep{green1998imaging} over the Indian Pines test site in North-western Indiana. Comprising $224$ spectral bands within a wavelength range of $400$ to $2500$ $nm$, the dataset excludes $24$ null and corrupted bands. The image has a spatial size of $145\times{145}$ pixels, with a spatial resolution of 20 mpp. It encompasses $16$ mutually exclusive vegetation classes. Disjoint Training/Test sample maps are illustrated in Figure \ref{Fig17}. 

%%%%%%%%%%%%%%%%%%%%%%%%%%
%%%%%%%%%%%%%%%%%%%%%%%%%%

\begin{figure}[!hbt]
    \centering
    \includegraphics[width=0.48\textwidth]{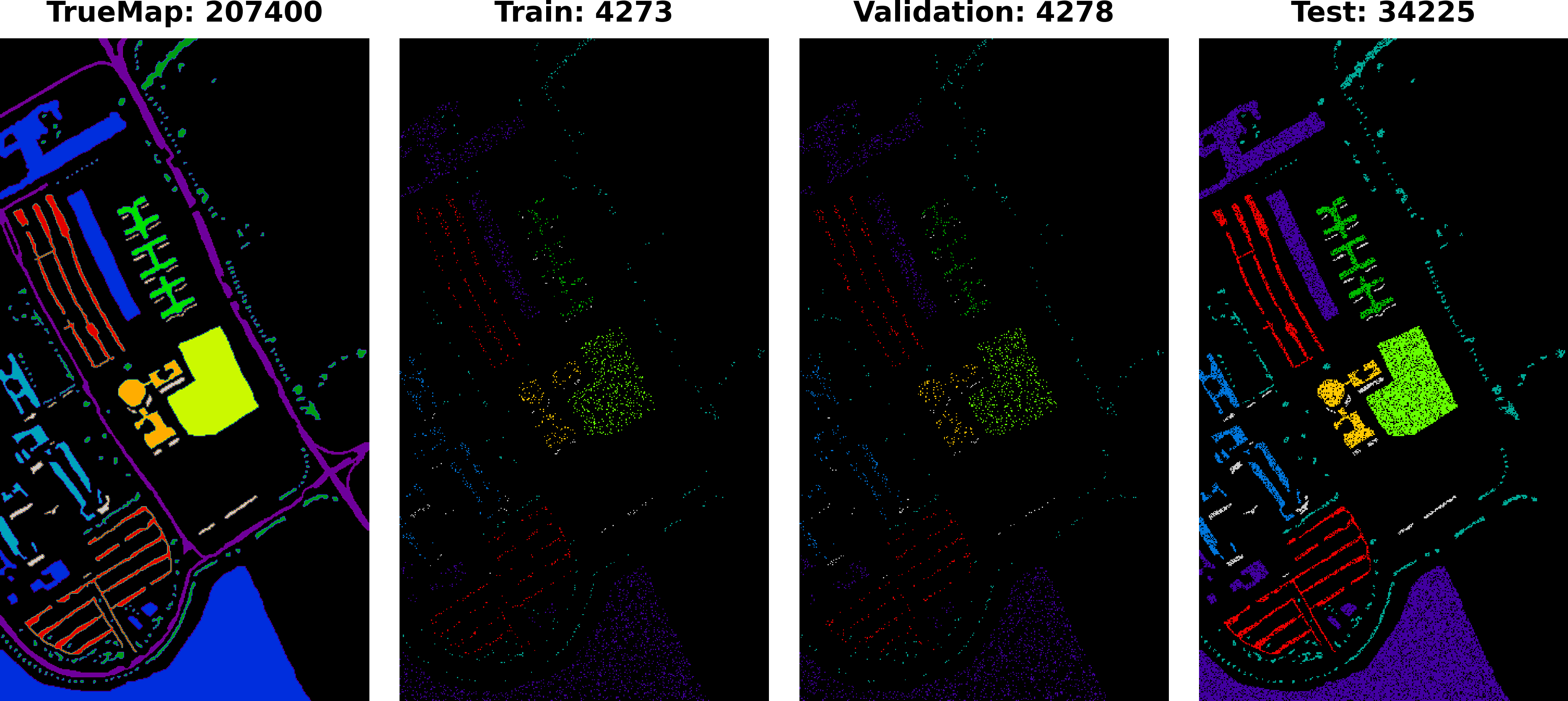}
    \caption{Pavia University ground truth maps, disjoint training, validation, and test ground truth maps.}
    \label{Fig18}
\end{figure}

The \textbf{University of Pavia} (UP) dataset, captured by the Reflective Optics System Imaging Spectrometer (ROSIS) sensor during a flight campaign over the university campus in Pavia, Northern Italy \cite{hong2020multimodal}, features dimensions of $610\times{340}$ pixels. With $103$ spectral bands spanning the wavelength range from $430$ to $860~{nm}$ and a spatial resolution of 2.5 mpp, the dataset includes 9 urban land-cover classes. Detailed class descriptions and ground truth maps can be found in Figure \ref{Fig18}. 

%%%%%%%%%%%%%%%%%%%%%%%%%%
%%%%%%%%%%%%%%%%%%%%%%%%%%

%%%%%%%%%%%%%%%%%%%%%%%%%%
\begin{table*}[htb]
  \centering
  % \rotatebox{90}{
  %   \begin{minipage}{0.99\textwidth}
      \centering
      \caption{Per-class classification results for \textbf{Pavia University} using various models with disjoint test set and the complete HSI as the test set with a patch size of $8\times 8$, 10\% allocated for training and validation samples, and the remaining 80\% for disjoint test samples.}
      \resizebox{\textwidth}{!}{\begin{tabular}{c|cc||cc||cc||cc||cc||cc||cc||cc||cc||cc||cc} \hline 
      \multirow{2}{*}{\textbf{Class}} & \multicolumn{2}{c|}{\textbf{2D CNN}} & \multicolumn{2}{c|}{\textbf{3D CNN}} & \multicolumn{2}{c|}{\textbf{Hybrid CNN}} & \multicolumn{2}{c|}{\textbf{2D Inception}} & \multicolumn{2}{c|}{\textbf{3D Inception}} & \multicolumn{2}{c|}{\textbf{Hybrid Inception}} & \multicolumn{2}{c|}{\textbf{2D Xception}} & \multicolumn{2}{c|}{\textbf{(2+1)D Xception}} & \multicolumn{2}{c|}{\textbf{SCSNet}} & \multicolumn{2}{c|}{\textbf{Attention Graph}} & \multicolumn{2}{c}{\textbf{Transformer}} \\ \cline{2-23}
        & \textbf{Test} & \textbf{HSI} & \textbf{Test} & \textbf{HSI} & \textbf{Test} & \textbf{HSI} & \textbf{Test} & \textbf{HSI} & \textbf{Test} & \textbf{HSI} & \textbf{Test} & \textbf{HSI} & \textbf{Test} & \textbf{HSI} & \textbf{Test} & \textbf{HSI} & \textbf{Test} & \textbf{HSI} & \textbf{Test} & \textbf{HSI} & \textbf{Test} & \textbf{HSI} \\ \hline

        Asphalt & 99.24 & 99.97 & 99.75 & \textbf{99.99} & 99.92 & \textbf{99.99} & 99.18 & 99.97 & 99.83 & \textbf{99.99} & 99.96 & \textbf{99.99} & 91.40 & 99.70 & 99.83 & \textbf{99.99} & 98.96 & 99.96 & 97.66 & 99.91 & 97.15 & 99.90 \\
        
        Meadows & 99.67 & 99.71 & 99.97 & 99.96 & 99.97 & 99.96 & 99.71 & 99.72 & \textbf{100} & \textbf{100} & 99.97 & 99.96 & 98.65 & 98.78 & 99.93 & 99.94 & 99.75 & 99.79 & 99.10 & 99.13 & 99.73 & 99.75 \\
        
        Gravel & 90.41 & 91.61 & 96.30 & 96.71 & 95.29 & 95.75 & 88.45 & 89.75 & 92.55 & 93.33 & 97.67 & \textbf{97.99} & 73.39 & 76.08 & 95.59 & 95.90 & 90.23 & 91.32 & 94.34 & 94.90 & 88.80 & 90.32 \\
        
        Trees  & 97.87 & 98.20 & 98.93 & 99.05 & 99.51 & 99.57 & 96.45 & 96.76 & 98.00 & 98.17 & 99.06 & \textbf{99.18} & 84.46 & 85.99 & 98.57 & 98.82 & 97.38 & 97.45 & 96.53 & 96.76 & 97.14 & 97.32 \\
        
        Painted & \textbf{100} & \textbf{100} & \textbf{100} & \textbf{100} & \textbf{100} & \textbf{100} & \textbf{100} & \textbf{100} & \textbf{100} & \textbf{100} & \textbf{100} & \textbf{100} & 89.96 & 91.00 & \textbf{100} & \textbf{100} & \textbf{100} & \textbf{100} & 99.81 & 99.85 & \textbf{100} & \textbf{100} \\
        
        Soil & 98.21 & 98.46 & 99.40 & 99.52 & \textbf{100} & \textbf{100} & 99.75 & 99.80 & 99.90 & 99.92 & \textbf{100} & \textbf{100} & 90.53 & 91.42 & 99.92 & 99.94 & 95.05 & 95.64 & 99.72 & 99.78 & 98.06 & 98.30 \\
        
        Bitumen & 99.34 & 99.39 & 99.15 & 99.24 & 99.53 & \textbf{99.62} & 97.27 & 97.74 & 99.90 & 99.92 & 99.34 & 99.39 & 90.60 & 91.72 & 98.96 & 99.09 & 94.64 & 95.11 & 95.48 & 95.93 & 95.95 & 96.46 \\
        
        Bricks & 96.91 & 97.28 & 98.40 & 98.61 & 99.18 & 99.26 & 94.53 & 95.11 & \textbf{99.76} & 99.72 & 99.01 & 99.15 & 68.22 & 71.37 & 99.01 & 99.18 & 92.90 & 93.48 & 93.48 & 93.91 & 93.04 & 93.69 \\
        
        Shadows & 96.56 & 97.04 & 99.60 & 99.68 & 99.73 & \textbf{99.78} & 98.94 & 99.15 & 99.47 & 99.57 & 99.20 & 99.26 & 57.91 & 61.77 & 98.15 & 98.31 & 90.50 & 91.12 & 92.87 & 93.98 & 98.28 & 98.20 \\ \hline 
        
        \textbf{Time(s)} & \textbf{2.72} & 82.92 & 5.51 & 97.59 & 2.96 & 88.29 & 5.52 & 85.66 & 20.89 & 140.14 & 10.64 & 101.53 & 10.64 & 98.41 & 41.35 & 272.71 & 17.63 & 167.59 & 4.76 & 93.31 & 10.64 & 138.70 \\ 

        \textbf{Kappa} & 98.07 & 99.15 & 99.27 & 99.68 & 99.50 & \textbf{99.77} & 97.77 & 99.00 & 99.23 & 99.65 & 99.57 & 99.81 & 87.11 & 94.16 & 99.29 & 99.68 & 96.68 & 98.48 & 97.12 & 98.65 & 96.95 & 98.62 \\
        
        \textbf{OA} & 98.54 & 99.73 & 99.45 & 99.90 & 99.62 & 99.92 & 98.32 & 99.68 & 99.42 & 99.89 & 99.68 & \textbf{99.94} & 90.27 & 98.19 & 99.46 & 99.90 & 97.50 & 99.53 & 97.83 & 99.58 & 97.7 & 99.57 \\
        
        \textbf{AA} & 97.58 & 97.97 & 99.06 & 99.20 & 99.24 & 99.33 & 97.15 & 97.56 & 98.83 & 98.96 & 99.36 & \textbf{99.44} & 82.78 & 85.32 & 98.89 & 99.02 & 95.49 & 95.99 & 96.56 & 97.13 & 96.47 & 97.11 \\ \hline 

      \end{tabular}}
    % \end{minipage}}
    \label{Tab3}
\end{table*}
%%%%%%%%%%%%%%%%%%%%%%%%%%
\begin{figure*}[!t]
    \centering
	\begin{subfigure}{0.32\textwidth}
		\includegraphics[width=0.99\textwidth]{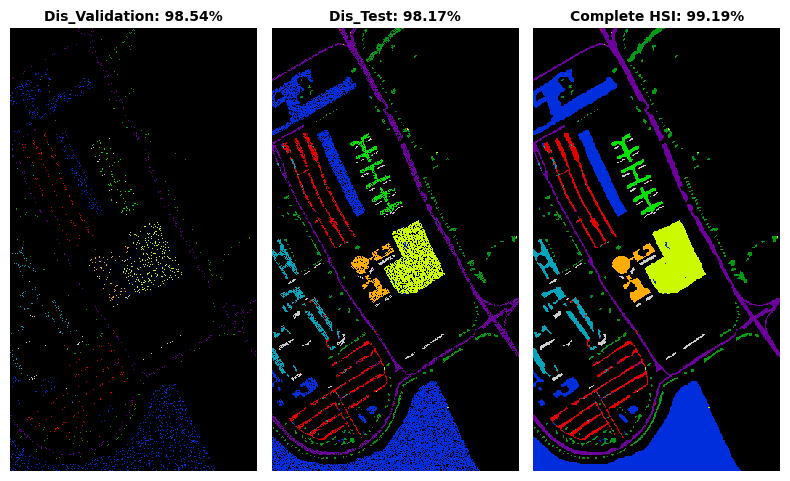}
		\caption{2D CNN} 
		\label{Fig19A}
	\end{subfigure}
	\begin{subfigure}{0.32\textwidth}
		\includegraphics[width=0.99\textwidth]{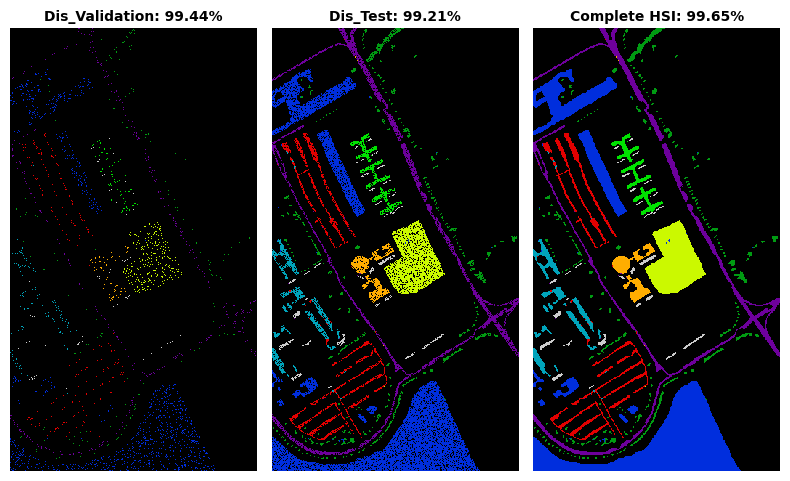}
		\caption{3D CNN}
		\label{Fig19B}
	\end{subfigure} 
	\begin{subfigure}{0.32\textwidth}
		\includegraphics[width=0.99\textwidth]{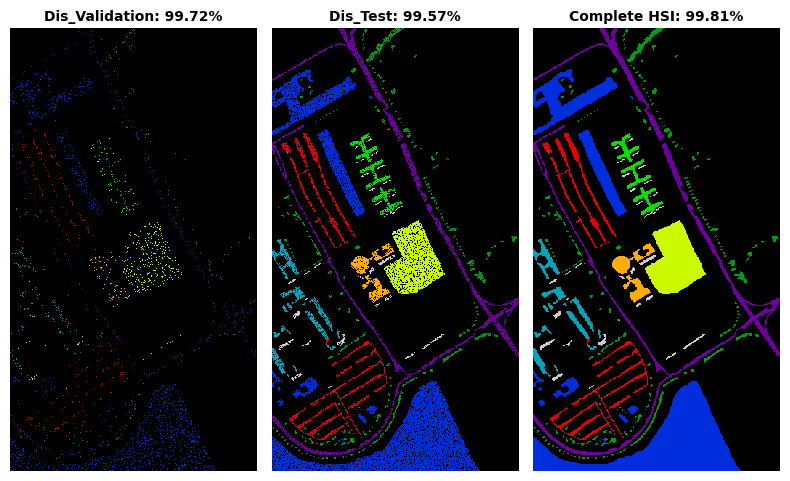}
		\caption{Hybrid CNN}
		\label{Fig19C}
	\end{subfigure} 
	\begin{subfigure}{0.32\textwidth}
		\includegraphics[width=0.99\textwidth]{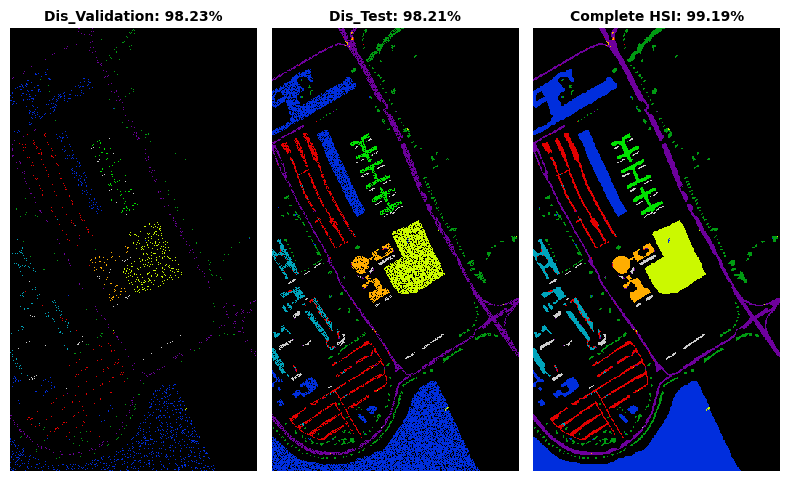}
		\caption{2D IN}
		\label{Fig19D}
	\end{subfigure} 
	\begin{subfigure}{0.32\textwidth}
		\includegraphics[width=0.99\textwidth]{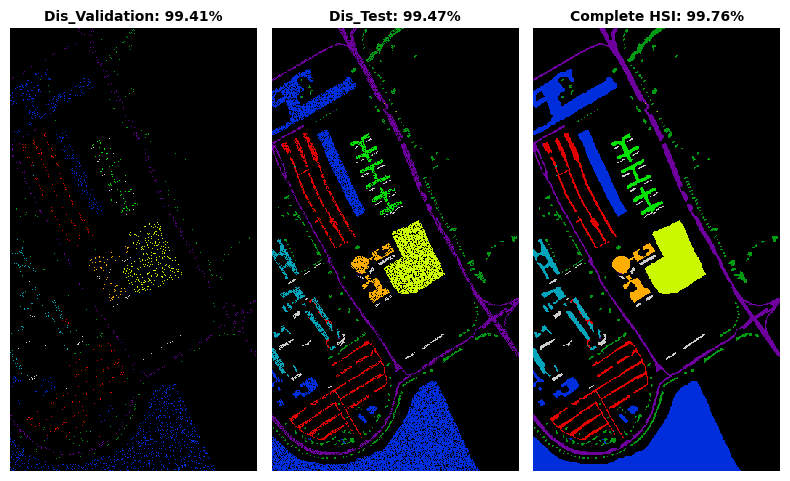}
		\caption{3D IN}
		\label{Fig19E}
	\end{subfigure} 
	\begin{subfigure}{0.32\textwidth}
		\includegraphics[width=0.99\textwidth]{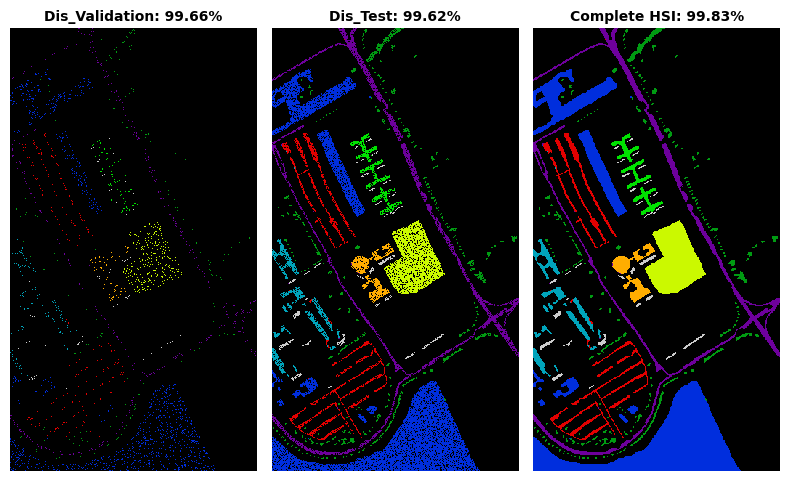}
		\caption{Hybrid IN}
		\label{Fig19F}
	\end{subfigure} 
	\begin{subfigure}{0.32\textwidth}
		\includegraphics[width=0.99\textwidth]{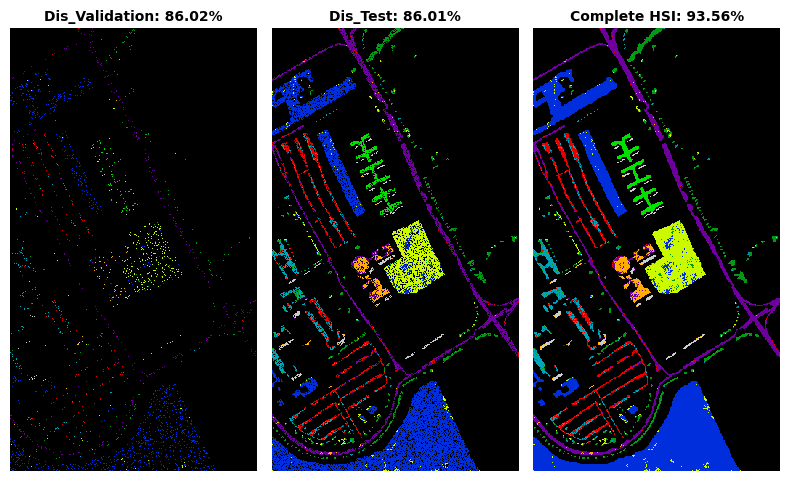}
		\caption{2D Exception Net}
		\label{Fig19G}
	\end{subfigure} 
	\begin{subfigure}{0.32\textwidth}
		\includegraphics[width=0.99\textwidth]{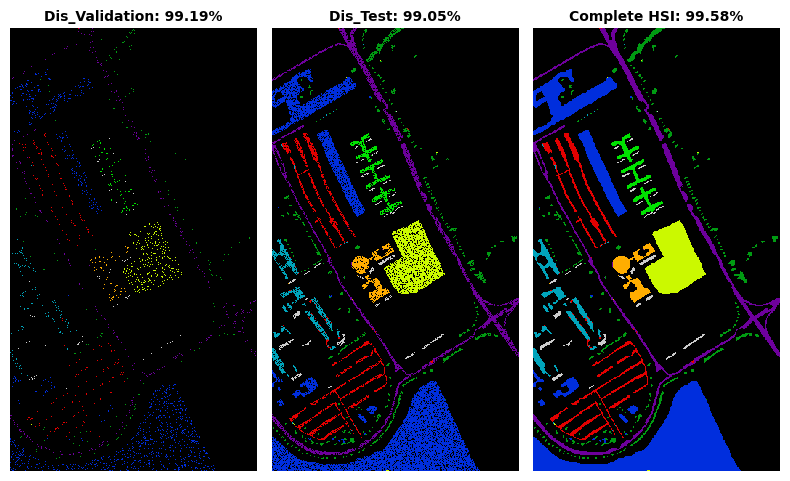}
		\caption{(2+1)D Exception Net}
		\label{Fig19H}
	\end{subfigure} 
    \begin{subfigure}{0.32\textwidth}
		\includegraphics[width=0.99\textwidth]{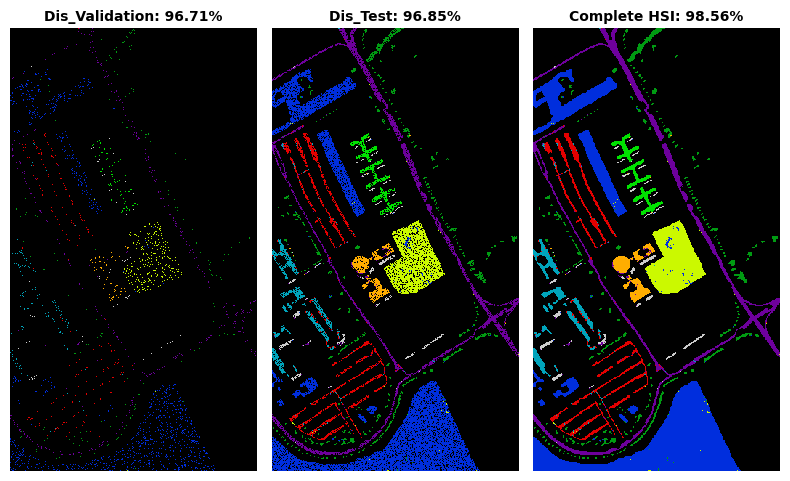}
		\caption{SCSNet}
		\label{Fig19I}
	\end{subfigure} 
	\begin{subfigure}{0.32\textwidth}
		\includegraphics[width=0.99\textwidth]{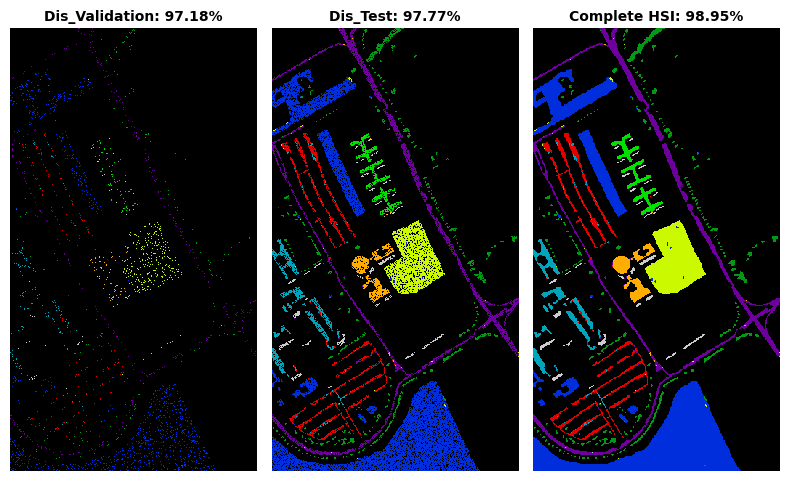}
		\caption{Attention Graph CNN}
		\label{Fig19J}
	\end{subfigure} 
	\begin{subfigure}{0.32\textwidth}
		\includegraphics[width=0.99\textwidth]{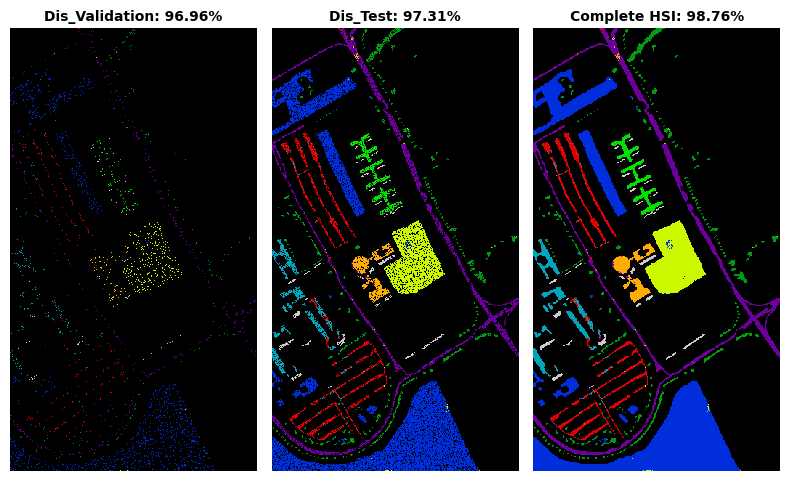}
		\caption{Spatial-Spectral Transformer}
		\label{Fig19K}
	\end{subfigure} 
\caption{\textbf{Pavia University Dataset:} Predicted land cover maps for disjoint validation, test, and the entire HSI used as a test set are provided. Comprehensive class-wise results can be found in Table \ref{Tab3}.}
\label{Fig19}
\end{figure*}
%%%%%%%%%%%%%%%%%%%%%%%%%%
%%%%%%%%%%%%%%%%%%%%%%%%%%
\begin{figure*}[!t]
    \centering
	\begin{subfigure}{0.49\textwidth}
		\includegraphics[width=0.99\textwidth]{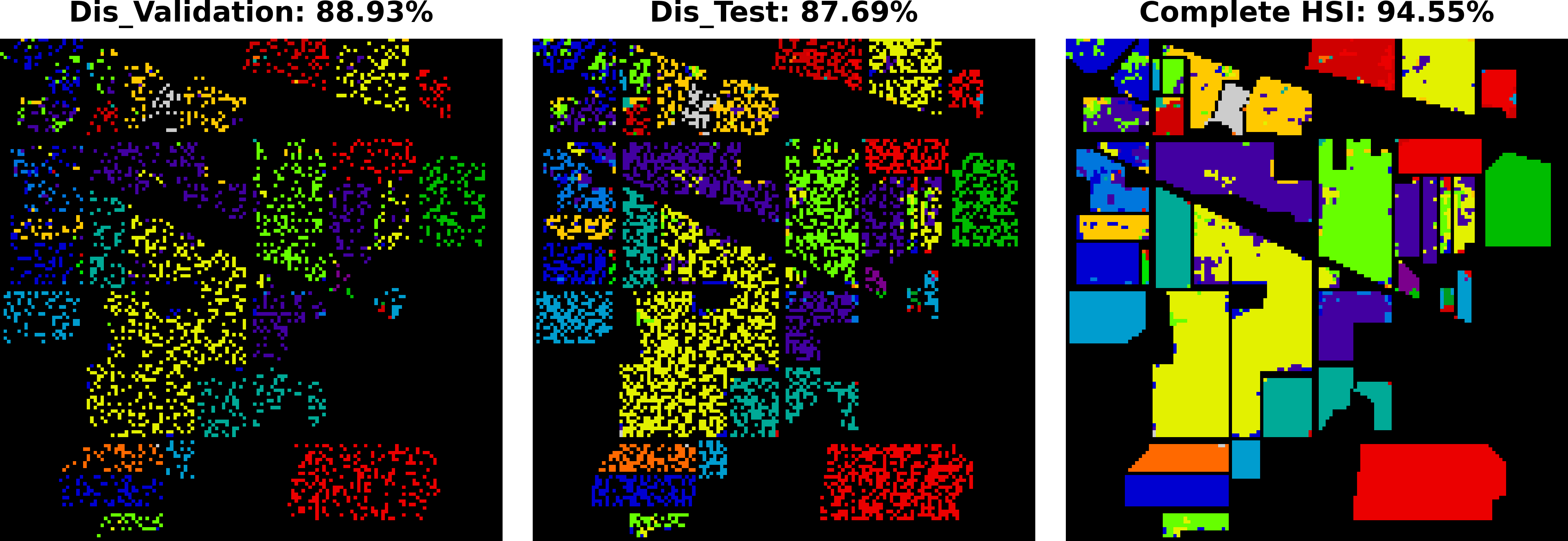}
		\caption{2D CNN} 
		\label{Fig20A}
	\end{subfigure}
	\begin{subfigure}{0.49\textwidth}
		\includegraphics[width=0.99\textwidth]{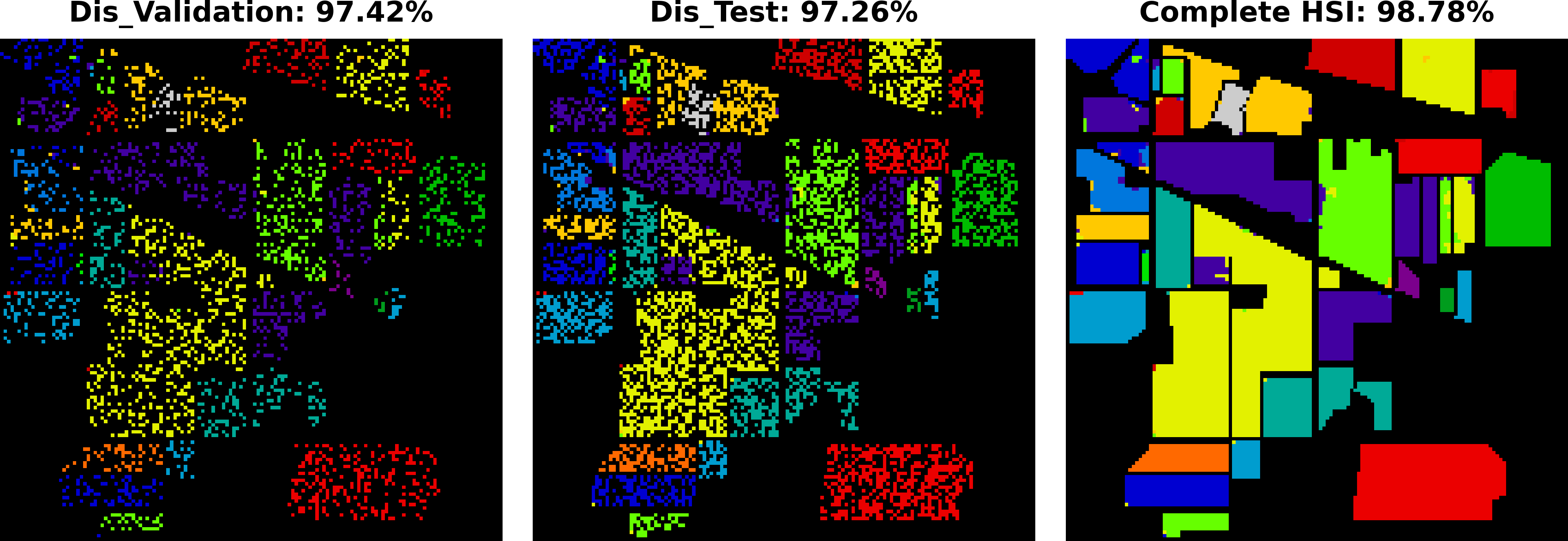}
		\caption{3D CNN}
		\label{Fig20B}
	\end{subfigure} 
	\begin{subfigure}{0.49\textwidth}
		\includegraphics[width=0.99\textwidth]{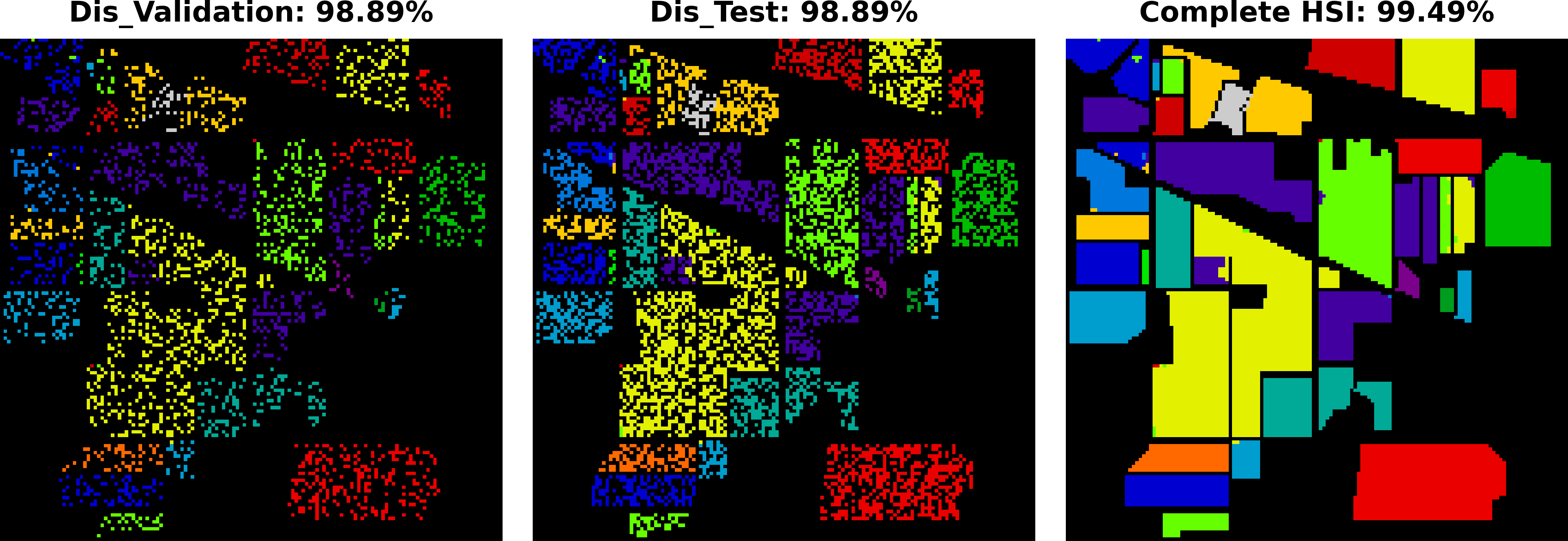}
		\caption{Hybrid CNN}
		\label{Fig20C}
	\end{subfigure} 
	\begin{subfigure}{0.49\textwidth}
		\includegraphics[width=0.99\textwidth]{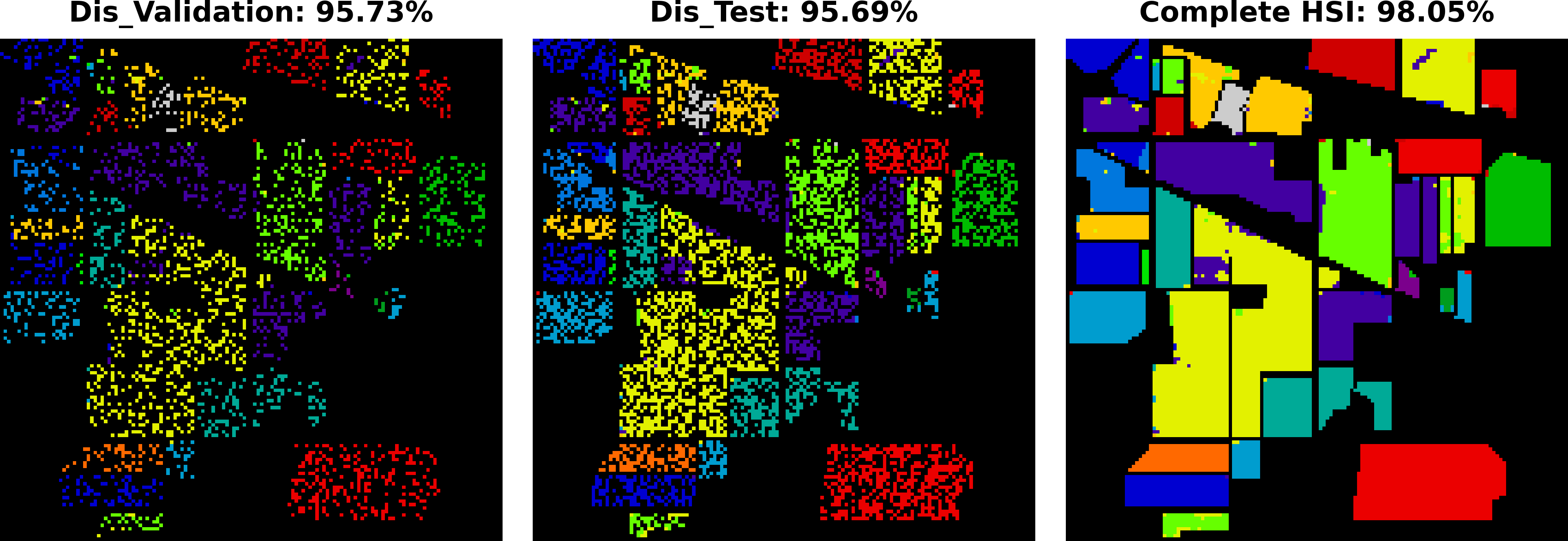}
		\caption{2D IN}
		\label{Fig20D}
	\end{subfigure} 
	\begin{subfigure}{0.49\textwidth}
		\includegraphics[width=0.99\textwidth]{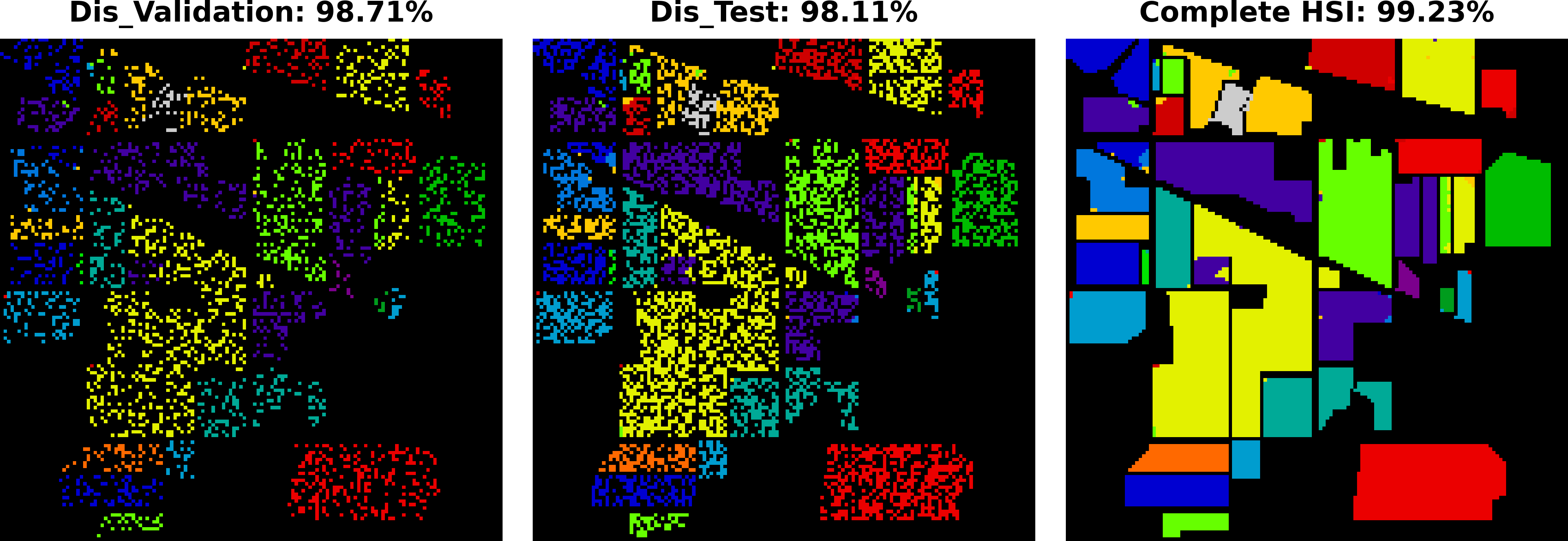}
		\caption{3D IN}
		\label{Fig20E}
	\end{subfigure} 
	\begin{subfigure}{0.49\textwidth}
		\includegraphics[width=0.99\textwidth]{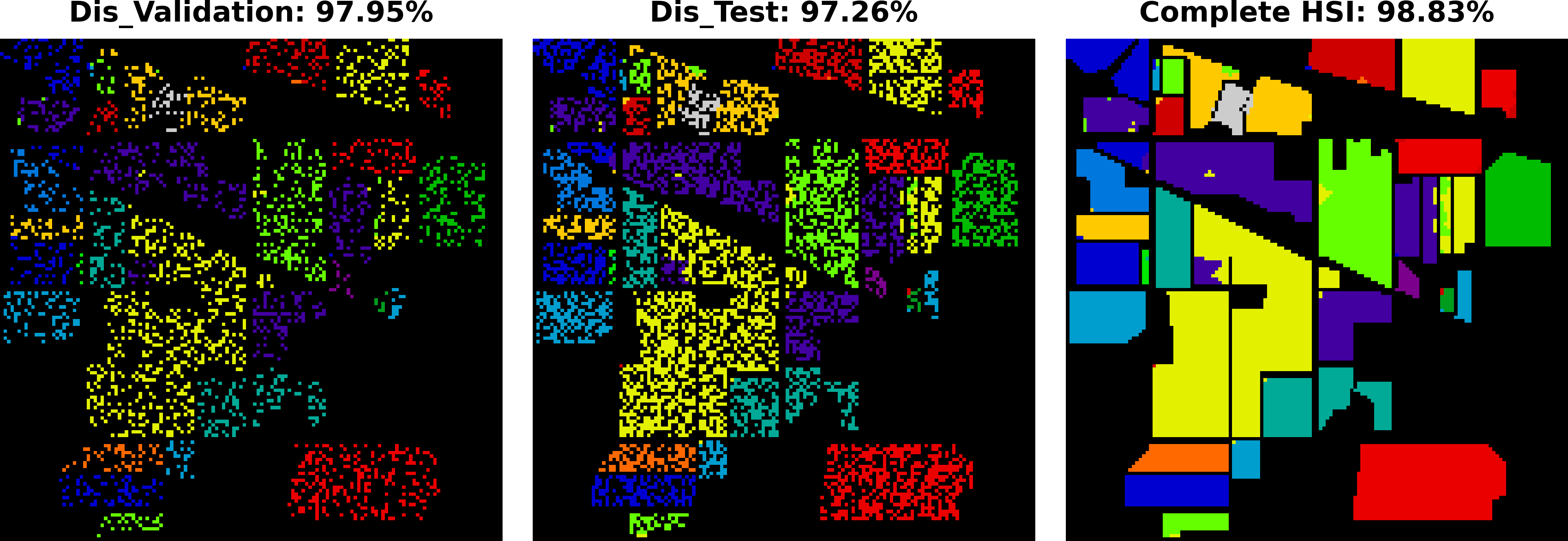}
		\caption{Hybrid IN}
		\label{Fig20F}
	\end{subfigure} 
	\begin{subfigure}{0.49\textwidth}
		\includegraphics[width=0.99\textwidth]{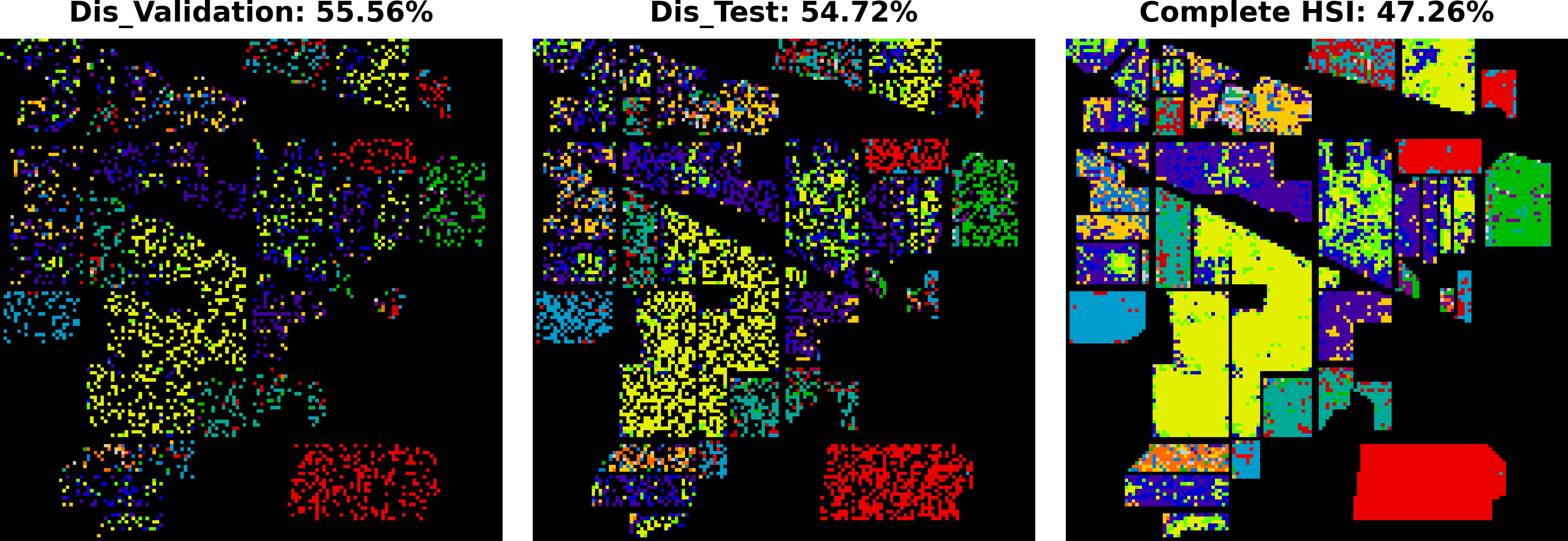}
		\caption{2D Exception Net}
		\label{Fig20G}
	\end{subfigure} 
	\begin{subfigure}{0.49\textwidth}
		\includegraphics[width=0.99\textwidth]{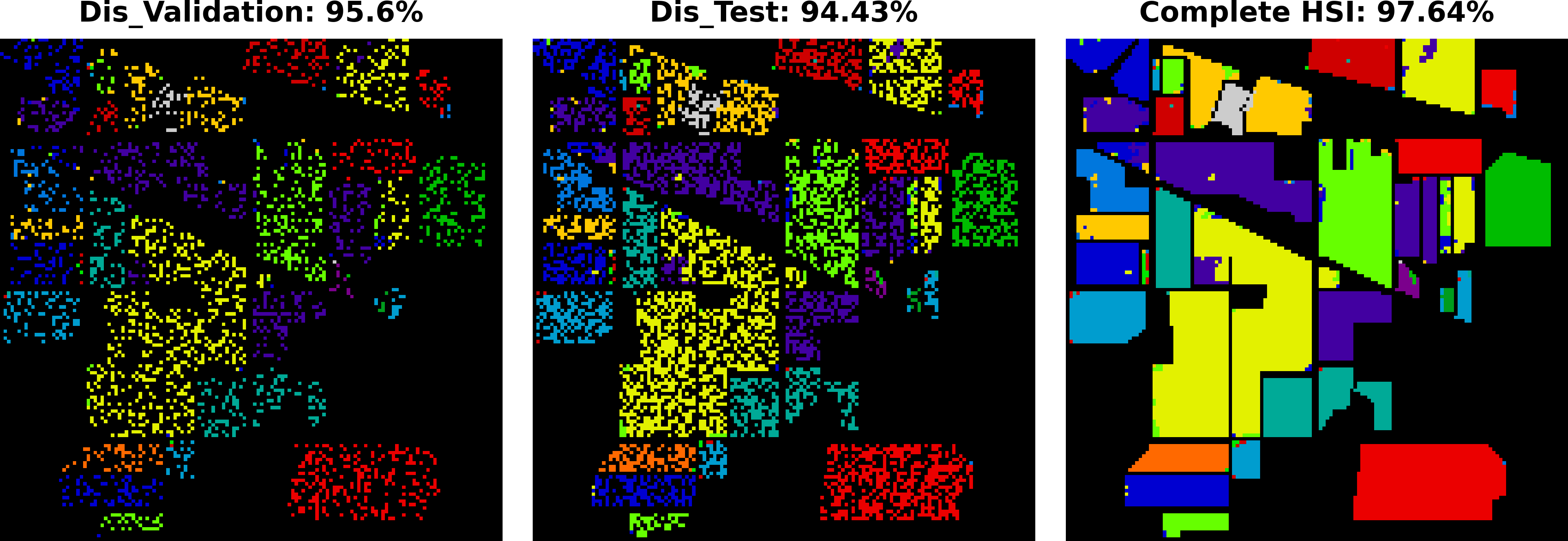}
		\caption{(2+1)D Exception Net}
		\label{Fig20H}
	\end{subfigure} 
    \begin{subfigure}{0.49\textwidth}
		\includegraphics[width=0.99\textwidth]{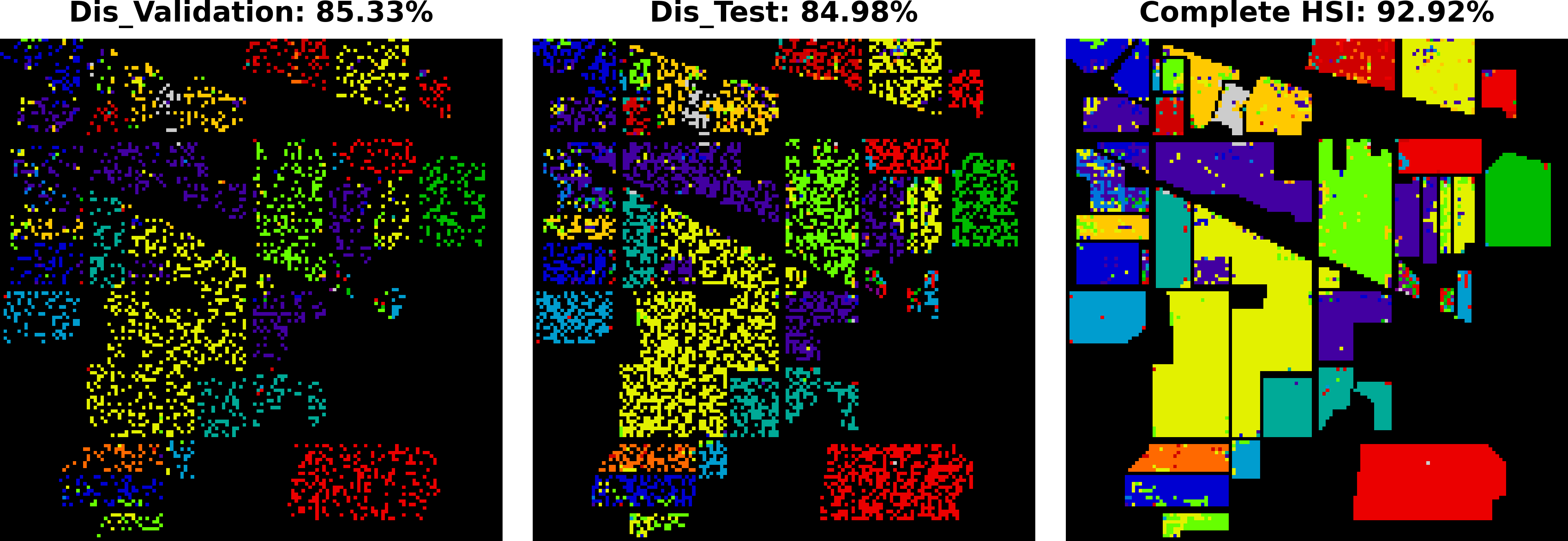}
		\caption{SCSNet}
		\label{Fig20I}
	\end{subfigure} 
	\begin{subfigure}{0.49\textwidth}
		\includegraphics[width=0.99\textwidth]{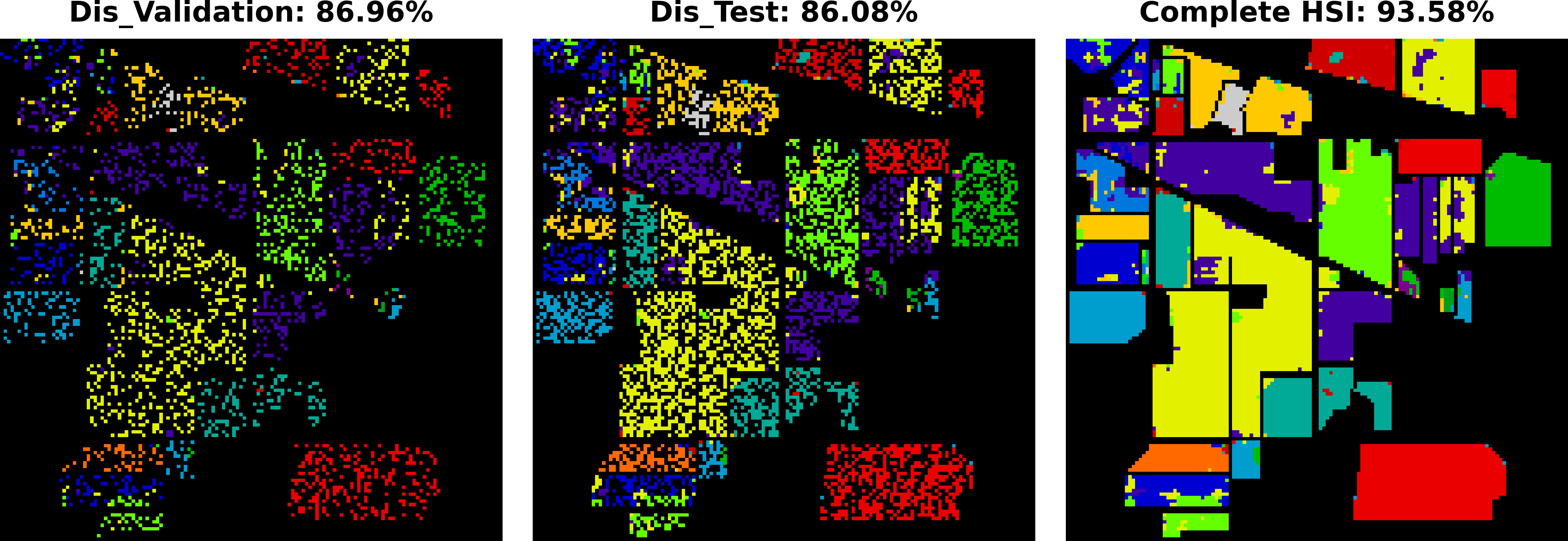}
		\caption{Attention Graph CNN}
		\label{Fig20J}
	\end{subfigure} 
	\begin{subfigure}{0.49\textwidth}
		\includegraphics[width=0.99\textwidth]{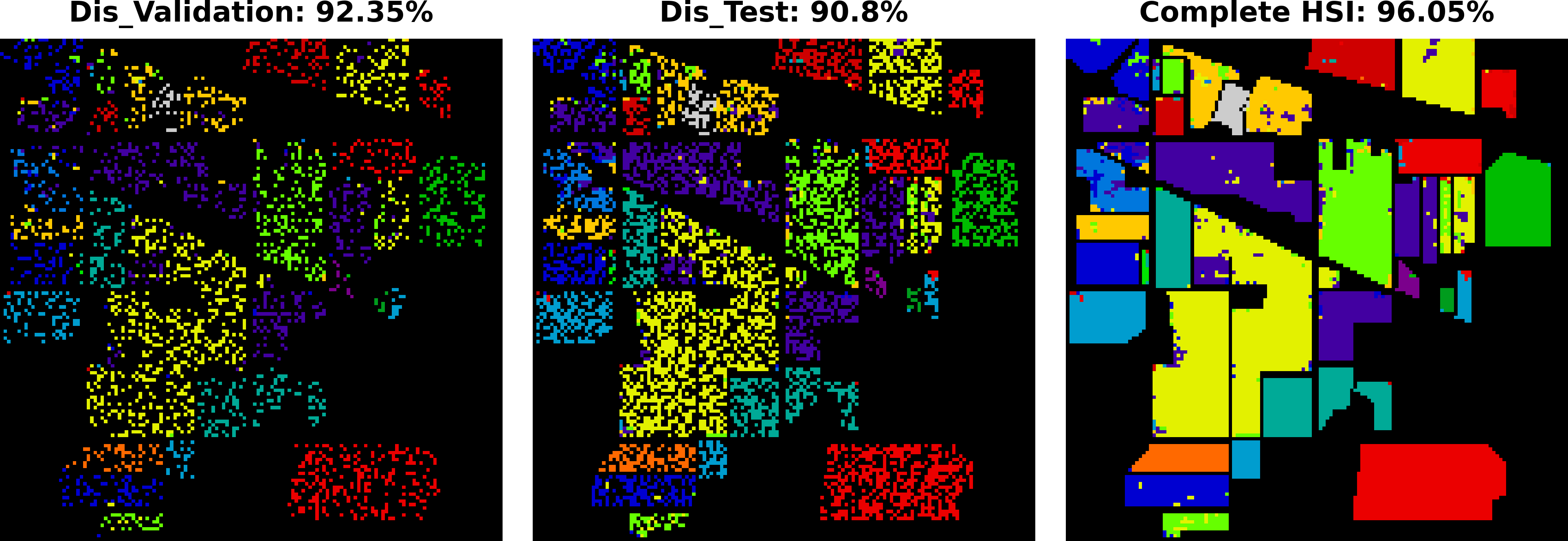}
		\caption{Spatial-Spectral Transformer}
		\label{Fig20K}
	\end{subfigure} 
\caption{\textbf{Indian Pines Dataset:} Predicted land cover maps for disjoint validation, test, and the entire HSI used as a test set are provided. Comprehensive class-wise results can be found in Table \ref{Tab4}.}
\label{Fig20}
\end{figure*}
%%%%%%%%%%%%%%%%%%%%%%%%%%
\begin{table*}[htb]
  \centering
  % \rotatebox{90}{
  %   \begin{minipage}{0.99\textwidth}
      \centering
      \caption{Per-class classification results for \textbf{Indian Pines} using various models with disjoint test set and the complete HSI as the test set with a patch size of $8\times 8$, 25\% allocated for training and validation samples, and the remaining 50\% for disjoint test samples.}
      \resizebox{\textwidth}{!}{\begin{tabular}{c|cc||cc||cc||cc||cc||cc||cc||cc||cc||cc||cc} \hline 
      \multirow{2}{*}{\textbf{Class}} & \multicolumn{2}{c|}{\textbf{2D CNN}} & \multicolumn{2}{c|}{\textbf{3D CNN}} & \multicolumn{2}{c|}{\textbf{Hybrid CNN}} & \multicolumn{2}{c|}{\textbf{2D Inception}} & \multicolumn{2}{c|}{\textbf{3D Inception}} & \multicolumn{2}{c|}{\textbf{Hybrid Inception}} & \multicolumn{2}{c|}{\textbf{2D Xception}} & \multicolumn{2}{c|}{\textbf{(2+1)D Xception}} & \multicolumn{2}{c|}{\textbf{SCSNet}} & \multicolumn{2}{c|}{\textbf{Attention Graph}} & \multicolumn{2}{c}{\textbf{Transformer}} \\ \cline{2-23}
        & \textbf{Test} & \textbf{HSI} & \textbf{Test} & \textbf{HSI} & \textbf{Test} & \textbf{HSI} & \textbf{Test} & \textbf{HSI} & \textbf{Test} & \textbf{HSI} & \textbf{Test} & \textbf{HSI} & \textbf{Test} & \textbf{HSI} & \textbf{Test} & \textbf{HSI} & \textbf{Test} & \textbf{HSI} & \textbf{Test} & \textbf{HSI} & \textbf{Test} & \textbf{HSI} \\ \hline

        Alfalfa & 47.82 & 99.81 & \textbf{100} & 99.99 & \textbf{100} & \textbf{100} & \textbf{100} & 99.99 & \textbf{100} & \textbf{100} & \textbf{100} & \textbf{100} & - & 99.57 & 82.60 & 99.92 & 39.13 & 99.81 & 86.95 & 99.95 & 91.30 & 99.96 \\
        
        Corn-notill & 95.65 & 96.63 & 98.03 & 98.17 & 99.01 & \textbf{99.22} & 93.97 & 95.02 & 98.73 & 99.01 & 98.73 & 99.01 & 68.34 & 7.98 & 91.31 & 92.85 & 90.47 & 92.15 & 93.83 & 95.30 & 91.73 & 94.04 \\
        
        Corn-mintill & 90.12 & 93.01 & 95.42 & 96.86 & 97.34 & \textbf{98.55} & 93.49 & 95.78 & 96.86 & 98.07 & 96.38 & 97.71 & 16.86 & 0.60 & 77.34 & 81.20 & 80.72 & 85.18 & 88.19 & 92.28 & 84.33 & 89.75 \\
        
        Corn & 82.35 & 83.96 & 96.63 & 96.62 & 98.31 & 97.46 & 91.59 & 91.98 & \textbf{100} & \textbf{100} & 99.15 & 99.15 & 31.93 & - & 84.87 & 89.02 & 48.73 & 54.43 & 64.70 & 74.68 & 74.78 & 79.74 \\
        
        Grass-pasture & 94.21 & 96.06 & 98.76 & 98.96 & 99.58 & \textbf{99.79} & 97.93 & 98.7 & 98.76 & 99.37 & 99.58 & \textbf{99.79} & 78.09 & 2.07 & 92.56 & 94.40 & 93.80 & 94.82 & 93.80 & 96.27 & 95.86 & 97.10 \\

        Grass-trees & 98.90 & 99.45 & 99.17 & 99.58 & 99.72 & 99.86 & 98.63 & 99.31 & 99.45 & 98.55 & 99.72 & \textbf{99.86} & 79.72 & 7.67 & 98.35 & 99.04 & 98.35 & 98.35 & 98.08 & 99.04 & 99.17 & 99.58 \\

        Grass-mowed & 50.00 & 64.28 & 92.85 & 96.42 & \textbf{100} & \textbf{100} & 50 & 67.85 & \textbf{100} & 99.72 & 78.57 & 85.71 & - & - & 92.85 & 96.42 & 14.28 & 25 & 85.71 & 89.28 & 92.85 & 96.42 \\

        Hay-windrowed & 99.16 & 99.58 & \textbf{100} & \textbf{100} & \textbf{100} & \textbf{100} & \textbf{100} & \textbf{100} & \textbf{100} & \textbf{100} & \textbf{100} & \textbf{100} & 73.64 & - & 99.58 & 99.79 & 96.23 & 97.48 & \textbf{100} & \textbf{100} & 99.16 & 99.58 \\

        Oats & 80 & 85 & 90 & 95 & \textbf{100} & \textbf{100} & 80 & 85 & 80 & \textbf{100} & 30 & 45 & - & 5 & 70 & 85 & 20 & 30 & 70 & 80 & 60 & 70 \\

        Soybean-notill & 90.94 & 93.41 & 92.79 & 95.16 & 97.53 & \textbf{98.45} & 88.47 & 92.28 & 97.94 & 90 & 94.65 & 95.88 & 5.96 & 0.10 & 79.42 & 81.79 & 80.86 & 86.00 & 83.74 & 88.06 & 83.53 & 87.75 \\

        Soybean-mintill & 88.43 & 91.20 & 99.02 & \textbf{99.30} & 98.20 & 98.77 & 94.13 & 95.80 & 98.94 & 99.14 & 98.85 & 99.22 & 88.92 & - & 90.55 & 93.07 & 90.55 & 93.36 & 95.27 & 96.65 & 92.91 & 94.66 \\

        Soybean-clean & 93.26 & 94.26 & 98.31 & 98.65 & \textbf{99.32} & \textbf{99.32} & 93.60 & 95.61 & 97.97 & 98.65 & 95.28 & 97.47 & 39.05 & - & 97.30 & 98.31 & 70.37 & 77.57 & 93.93 & 94.94 & 83.83 & 88.02 \\

        Wheat & \textbf{100} & \textbf{100} & \textbf{100} & \textbf{100} & \textbf{100} & \textbf{100} & \textbf{100} & \textbf{100} & \textbf{100} & 99.51 & \textbf{100} & \textbf{100} & 20.38 & - & 97.08 & 97.56 & 97.08 & 97.56 & 99.02 & 99.02 & \textbf{100} & \textbf{100} \\

        Woods & 98.26 & 98.57 & 98.57 & 98.89 & 99.05 & 99.28 & 98.10 & 98.26 & 99.52 & \textbf{99.60} & 98.73 & 99.20 & 87.51 & - & 98.42 & 99.13 & 95.89 & 97.54 & 97.78 & 98.41 & 97.78 & 98.26 \\
        
        Buildings & 94.30 & 95.59 & 97.92 & 98.18 & 99.48 & 99.74 & 92.74 & 95.07 & \textbf{100} & \textbf{100} & \textbf{100} & \textbf{100} & 27.46 & - & 90.15 & 94.04 & 78.75 & 84.45 & 89.11 & 93.78 & 94.30 & 96.37 \\
        
        Stone-Steel & \textbf{100} & \textbf{100} & 95.74 & 97.84 & 95.74 & 97.84 & \textbf{100} & \textbf{100} & 97.87 & 98.92 & \textbf{100} & \textbf{100} & - & - & 95.74 & 95.69 & 34.04 & 43.01 & 93.61 & 95.69 & 97.87 & 98.92 \\ \hline

        \textbf{Time(s)} & \textbf{0.63} & 8.25 & 0.66 & 8.95 & 0.50 & 8.81 & 0.78 & 9.66 & 2.19 & 14.90 & 1.36 & 10.60 & 1.61 & 11.90 & 10.32 & 28.65 &  5.20 & 21.46 & 1.39 & 9.02 & 1.78 & 15.16 \\ 

        \textbf{Kappa} & 91.97 & 96.29 & 97.51 & 99.21 & 98.51 & 99.38 & 93.96 & 97.35 & 98.62 & \textbf{99.41} & 97.80 & 99.08 & 54.47 & 29.26 & 89.71 & 95.08 & 85.02 & 93.07 & 92.01 & 96.59 & 90.72 & 95.92 \\
        
        \textbf{OA} & 92.94 & 97.37 & 97.82 & 99.21 & 98.69 & 99.56 & 94.69 & 98.12 & 98.79 & \textbf{99.58} & 98.07 & 99.35 & 60.78 & 52.14 & 90.93 & 96.52 & 86.91 & 95.11 & 93.00 & 97.59 & 91.86 & 97.11 \\
        
        \textbf{AA} & 87.72 & 93.17 & 97.08 & 98.11 & 98.96 & \textbf{99.27} & 92.04 & 94.42 & 97.88 & 98.79 & 93.11 & 94.88 & 38.62 & 7.68 & 89.89 & 93.58 & 70.58 & 78.55 & 89.61 & 93.34 & 89.97 & 93.14 \\ \hline 

      \end{tabular}}
    % \end{minipage}}
    \label{Tab4}
\end{table*}
%%%%%%%%%%%%%%%%%%%%%%%%%%
%%%%%%%%%%%%%%%%%%%%%%%%%%
\begin{figure*}[!t]
    \centering
	\begin{subfigure}{0.99\textwidth}
		\includegraphics[width=0.99\textwidth]{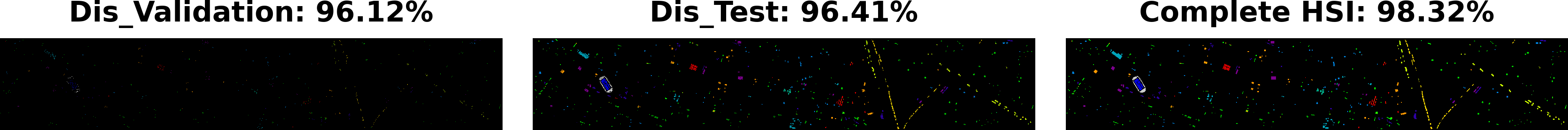}
		\caption{2D CNN} 
		\label{Fig21A}
	\end{subfigure}
	\begin{subfigure}{0.99\textwidth}
		\includegraphics[width=0.99\textwidth]{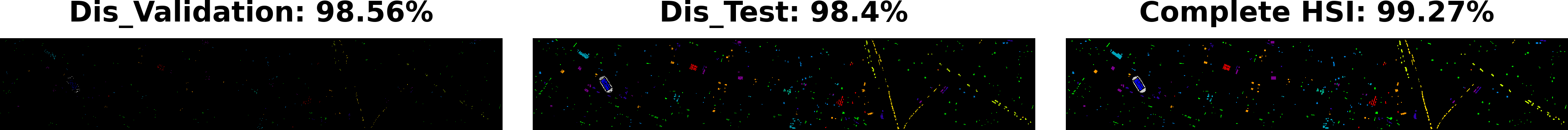}
		\caption{3D CNN}
		\label{Fig21B}
	\end{subfigure} 
	\begin{subfigure}{0.99\textwidth}
		\includegraphics[width=0.99\textwidth]{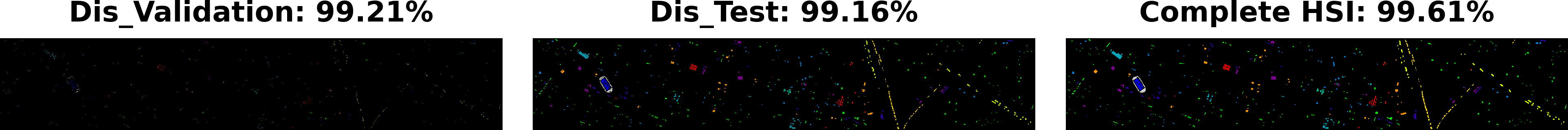}
		\caption{Hybrid CNN}
		\label{Fig21C}
	\end{subfigure} 
	\begin{subfigure}{0.99\textwidth}
		\includegraphics[width=0.99\textwidth]{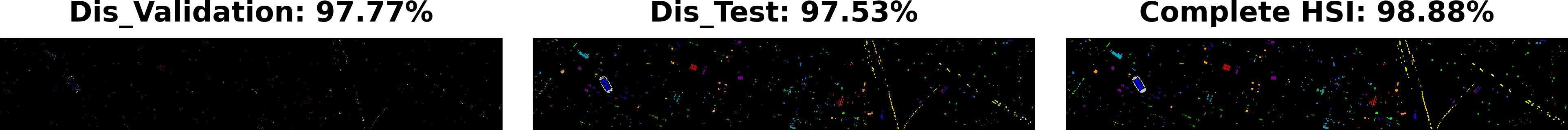}
		\caption{2D IN}
		\label{Fig21D}
	\end{subfigure} 
	\begin{subfigure}{0.99\textwidth}
		\includegraphics[width=0.99\textwidth]{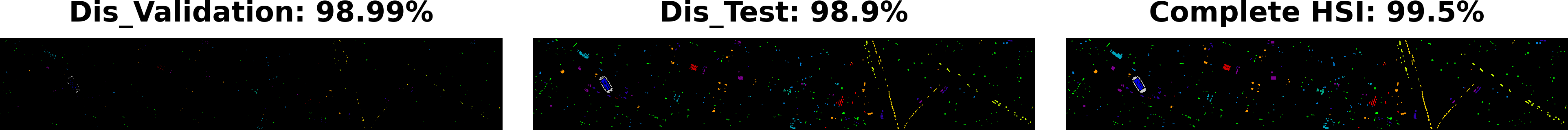}
		\caption{3D IN}
		\label{Fig21E}
	\end{subfigure} 
	\begin{subfigure}{0.99\textwidth}
		\includegraphics[width=0.99\textwidth]{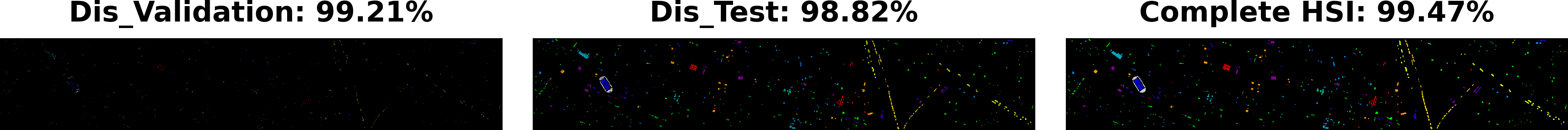}
		\caption{Hybrid IN}
		\label{Fig21F}
	\end{subfigure} 
	\begin{subfigure}{0.99\textwidth}
		\includegraphics[width=0.99\textwidth]{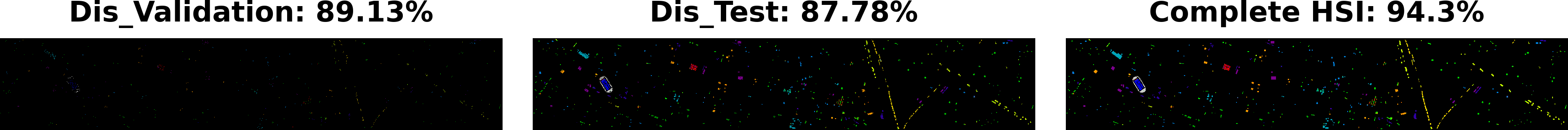}
		\caption{2D Exception Net}
		\label{Fig21G}
	\end{subfigure} 
	\begin{subfigure}{0.99\textwidth}
		\includegraphics[width=0.99\textwidth]{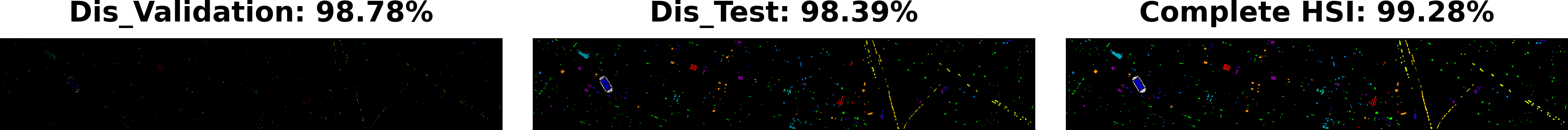}
		\caption{(2+1)D Exception Net}
		\label{Fig21H}
	\end{subfigure} 
    \begin{subfigure}{0.99\textwidth}
		\includegraphics[width=0.99\textwidth]{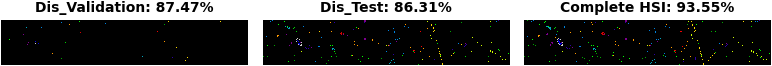}
		\caption{SCSNet}
		\label{Fig21I}
	\end{subfigure} 
	\begin{subfigure}{0.99\textwidth}
		\includegraphics[width=0.99\textwidth]{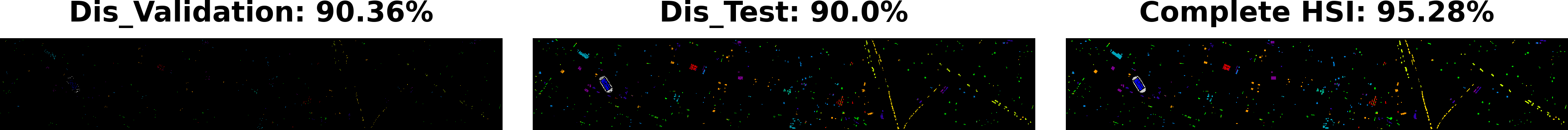}
		\caption{Attention Graph CNN}
		\label{Fig21J}
	\end{subfigure} 
	\begin{subfigure}{0.99\textwidth}
		\includegraphics[width=0.99\textwidth]{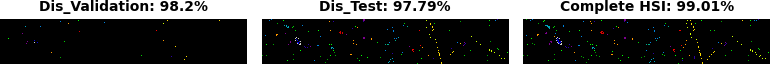}
		\caption{Spatial-Spectral Transformer}
		\label{Fig21K}
	\end{subfigure} 
\caption{\textbf{University of Houston Dataset:} Predicted land cover maps for disjoint validation, test, and the entire HSI used as a test set are provided. Comprehensive class-wise results can be found in Table \ref{Tab5}.}
\label{Fig21}
\end{figure*}
%%%%%%%%%%%%%%%%%%%%%%%%%%
\begin{table*}[htb]
  \centering
  % \rotatebox{90}{
  %   \begin{minipage}{0.99\textwidth}
      \centering
      \caption{Per-class classification results for \textbf{University of Houston} using various models with disjoint test set and the complete HSI as the test set with a patch size of $8\times 8$, 10\% allocated for training and validation samples, and the remaining 80\% for disjoint test samples.}
      \resizebox{\textwidth}{!}{\begin{tabular}{c|cc||cc||cc||cc||cc||cc||cc||cc||cc||cc||cc} \hline 
      \multirow{2}{*}{\textbf{Class}} & \multicolumn{2}{c|}{\textbf{2D CNN}} & \multicolumn{2}{c|}{\textbf{3D CNN}} & \multicolumn{2}{c|}{\textbf{Hybrid CNN}} & \multicolumn{2}{c|}{\textbf{2D Inception}} & \multicolumn{2}{c|}{\textbf{3D Inception}} & \multicolumn{2}{c|}{\textbf{Hybrid Inception}} & \multicolumn{2}{c|}{\textbf{2D Xception}} & \multicolumn{2}{c|}{\textbf{(2+1)D Xception}} & \multicolumn{2}{c|}{\textbf{SCSNet}} & \multicolumn{2}{c|}{\textbf{Attention Graph}} & \multicolumn{2}{c}{\textbf{Transformer}} \\ \cline{2-23}
        & \textbf{Test} & \textbf{HSI} & \textbf{Test} & \textbf{HSI} & \textbf{Test} & \textbf{HSI} & \textbf{Test} & \textbf{HSI} & \textbf{Test} & \textbf{HSI} & \textbf{Test} & \textbf{HSI} & \textbf{Test} & \textbf{HSI} & \textbf{Test} & \textbf{HSI} & \textbf{Test} & \textbf{HSI} & \textbf{Test} & \textbf{HSI} & \textbf{Test} & \textbf{HSI} \\ \hline
        
        Healthy grass & 98.10 & \textbf{99.99} & 98.60 & \textbf{99.99} & 98.10 & \textbf{99.99} & 98.10 & \textbf{99.99} & 98.80 & \textbf{99.99} & 98.80 & \textbf{99.99} & 96.80 & \textbf{99.99} & 98.60 & \textbf{99.99} & 95.10 & \textbf{99.99} & 94.10 & 99.98 & 98.10 & \textbf{99.99} \\
        
        Stressed grass & 97.70 & 98.08 & 97.80 & 98.16 & 96.91 & 97.28 & 98.00 & 98.40 & 98.90 & \textbf{99.12} & 98.50 & 98.80 & 97.80 & 98.00 & 98.70 & 98.96 & 94.32 & 94.81 & 97.41 & 97.60 & 97.70 & 98.08 \\
        
        Synthetic grass & 99.64 & \textbf{99.71} & 97.67 & 98.13 & 99.64 & \textbf{99.71} & 99.64 & \textbf{99.71} & 99.64 & \textbf{99.71} & 99.46 & 99.56 & 95.87 & 96.41 & 98.02 & 98.42 & 97.13 & 97.70 & 99.64 & \textbf{99.71} & 99.64 & \textbf{99.71} \\
        
        Trees & 99.89 & 99.91 & 99.69 & 99.67 & 99.69 & 99.75 & 98.69 & 98.87 & \textbf{100} & \textbf{100} & \textbf{100} & \textbf{100} & 98.99 & 99.11 & 99.89 & 99.91 & 96.48 & 96.54 & 97.08 & 97.34 & 99.49 & 99.59 \\
        
        Soil & \textbf{100} & \textbf{100} & 99.89 & 99.91 & \textbf{100} & \textbf{100} & 99.89 & 99.91 & \textbf{100} & \textbf{100} & \textbf{100} & \textbf{100} & 99.49 & 99.51 & \textbf{100} & \textbf{100} & 97.28 & 97.34 & 98.79 & 98.95 & \textbf{100} & \textbf{100} \\
        
        Water & 86.92 & 89.53 & 96.15 & 96.92 & 97.30 & \textbf{97.84} & 91.53 & 93.23 & 96.92 & 97.53 & 93.07 & 94.15 & 78.46 & 81.23 & 94.61 & 95.38 & 57.30 & 60.30 & 86.15 & 87.69 & 99.23 & 99.38 \\
        
        Residential & 95.66 & 95.74 & 96.84 & 97.16 & 98.42 & 98.65 & 97.04 & 97.08 & 98.12 & 98.34 & 98.91 & \textbf{98.97} & 90.54 & 91.71 & 96.35 & 96.68 & 94.87 & 95.50 & 88.07 & 89.11 & 95.96 & 96.45 \\
        
        Commercial & 94.67 & 95.01 & 97.48 & 97.74 & 97.89 & \textbf{98.07} & 97.38 & 97.66 & 97.18 & 97.50 & 97.89 & \textbf{98.07} & 86.94 & 88.58 & 97.69 & 97.90 & 86.44 & 87.78 & 94.97 & 95.25 & 97.59 & 97.82 \\
        
        Road & 96.00 & 96.48 & 98.70 & 98.88 & 98.60 & 98.72 & 98.90 & 99.04 & 99.30 & \textbf{99.44} & 98.30 & 98.40 & 76.64 & 77.55 & 98.60 & 98.72 & 80.73 & 82.10 & 83.83 & 85.14 & 98.90 & 98.96 \\
        
        Highway & 96.74 & 97.06 & 98.57 & 98.85 & 98.77 & 98.94 & 98.67 & 98.69 & 99.38 & \textbf{99.42} & 99.28 & 99.34 & 81.26 & 81.98 & 99.28 & 99.34 & 87.16 & 88.75 & 93.89 & 94.45 & 98.26 & 98.45 \\
        
        Railway & 98.58 & 98.70 & 98.88 & 99.02 & 98.68 & 98.94 & 98.58 & 98.86 & 98.58 & 98.86 & 99.19 & \textbf{99.35} & 91.49 & 92.22 & 95.74 & 96.19 & 96.65 & 96.84 & 88.76 & 89.95 & 99.69 & 99.75 \\
        
        Parking Lot 1 & 92.19 & 92.37 & 99.39 & 99.35 & 99.08 & 99.10 & 98.88 & 99.02 & 99.39 & 99.43 & 99.69 & 99.67 & 84.49 & 85.23 & 99.79 & \textbf{99.83} & 89.26 & 90.10 & 99.59 & 99.67 & 99.39 & 99.43 \\
        
        Parking Lot 2 & 93.88 & 94.88 & 93.08 & 94.24 & 97.60 & 97.65 & 84.04 & 85.28 & 94.41 & 94.88 & 98.13 & \textbf{98.50} & 40.95 & 47.33 & 94.68 & 95.52 & 72.87 & 73.98 & 91.22 & 92.53 & 84.57 & 86.56 \\
        
        Tennis Court & \textbf{100} & \textbf{100} & 98.83 & 99.06 & \textbf{100} & \textbf{100} & 98.83 & 99.06 & \textbf{100} & \textbf{100} & \textbf{100} & \textbf{100} & 93.87 & 93.69 & \textbf{100} & \textbf{100} & 83.38 & 84.81 & 80.46 & 82.94 & \textbf{100} & \textbf{100} \\
        
        Running Track & 98.48 & 98.48 & \textbf{100} & \textbf{100} & \textbf{100} & \textbf{100} & \textbf{100} & \textbf{100} & \textbf{100} & \textbf{100} & \textbf{100} & \textbf{100} & 95.07 & 95.75 & \textbf{100} & \textbf{100} & 97.72 & 97.72 & 98.29 & 98.63 & \textbf{100} & \textbf{100} \\ \hline 

        \textbf{Time(S)} & \textbf{1.15} & 274.67 & 1.43 & 285.60 & 41.12 & 2932.98 & 1.41 & 284.12 & 4.95 & 460.70 & 5.34 & 351.17 & 2.56 & 312.79 & 20.65 & 869.61 & 20.09 & 122.62 & 5.26 & 559.68 & 7.17 & 714.78 \\ 

        \textbf{Kappa} & 96.67 & 98.44 & 98.26 & 99.22 & 98.6 & 99.36 & 97.79 & 98.99 & 98.8 & 99.46 & 98.90 & \textbf{99.50} & 88.32 & 94.51 & 98.22 & 99.19 & 90.01 & 95.29 & 92.98 & 96.76 & 98.10 & 99.14 \\
        
        \textbf{OA} & 96.92 & 99.93 & 98.39 & 99.96 & 98.7 & 99.97 & 97.96 & 99.95 & 98.89 & \textbf{99.97} & 98.99 & \textbf{99.97} & 89.21 & 99.77 & 98.35 & 99.96 & 90.75 & 99.80 & 93.51 & 99.86 & 98.25 & 99.96 \\
        
        \textbf{AA} & 96.57 & 97.07 & 98.11 & 98.48 & 98.72 & 98.98 & 97.22 & 97.66 & 98.71 & 98.95 & 98.75 & \textbf{98.99} & 87.25 & 88.56 & 98.13 & 98.46 & 88.44 & 89.62 & 92.82 & 93.94 & 97.89 & 98.28 \\ \hline 

      \end{tabular}}
    % \end{minipage}}
    \label{Tab5}
\end{table*}
%%%%%%%%%%%%%%%%%%%%%%%%%%

To substantiate the outlined concepts in this survey and validate the claims, recent contributions encompass a variety of models, including 2D CNN, 3D CNN, Hyrbid CNN, 2D Inception Network, 3D Inception Net, Hybrid Inception Net, 2D Xception Net, (2+1)D Extreme Exception Net (EX Net), Attention Graph CNN, SCSNet: Sharpend Cosine Similarity-based Neural Network, and Spatial-Spectral Transformer \citep{ahmad2024importance}. A comparative analysis of experimental results has been conducted by considering representative works for each model as shown in Table \ref{Tab3}, Table \ref{Tab4}, Table\ref{Tab5} and Figure \ref{Fig19}, Figure \ref{Fig20}, and Figure \ref{Fig21}. To a certain degree, all the aforementioned studies rely on Convolutional Networks and undergo evaluation using three benchmark HSI datasets, namely IP, PU, and UH Scene. This survey specifically focuses on assessing the robustness of these models, taking into account the challenge of classifying HSI with a limited sample size in the training data for joint spatial-spectral classification. Comparative methods tend to misclassify samples with similar spatial structures, exemplified by the confusion between Meadows and Bare Soil classes in the Pavia University dataset, as illustrated in Table \ref{Tab3}, Table \ref{Tab4}, and Table\ref{Tab5} and Figure \ref{Fig19}, Figure \ref{Fig20}, and Figure \ref{Fig21}. Furthermore, the overall accuracy for the Grapes Untrained class is lower compared to other classes, attributed to the reasons mentioned earlier. In summary, it can be concluded that higher accuracy can be attained by augmenting the number of labeled training samples. Therefore, a larger set of labeled training samples has the potential to yield improved accuracies across all competing methods. 

In general, the Hybrid CNN and 3D Inception nets consistently outperform other comparative methods, particularly when dealing with a smaller number of labeled training samples. This suggests that these works exhibit stability and are less sensitive to variations in the number of training samples. While the accuracies of these methods improve with an increase in the number of training samples, it is noteworthy that other methods may outperform them. This trend persists even with a higher number of training samples. Consequently, one can infer that the works Attention Graph, SCSNet, and Transformers partially address the challenges posed by limited training sample availability, particularly when considering disjoint train/test samples. Furthermore, it can be deduced that models based on convolutions exhibit subpar performance compared to other models. Despite the advantage of the convolutional process, these models may fail to learn effectively in the absence of constraints. Additionally, the symmetric architecture can lead to an explosion of training parameters, escalating the training difficulty. While the works Attention Graph and SCSNet successfully address these challenges, the approach taken by Exception Net, which neglects the adoption of a greedy layer-wise strategy, results in inferior outcomes. This indicates a potential for enhancing methods in this category. 

In summary, Transformer and Graph-based classification yields significantly better results than convolution-based methods, especially when dealing with the limited availability of labeled training samples \citep{10399798, ahmad2024transformers}. While CNNs can capture the internal structure, the resulting feature representation may lack task-specific characteristics, explaining their comparatively lower performance compared to other models \citep{ahmad2021artifacts, ahmad2024pyramid, ahmad2024importance, BUTT2024103773}. 

%%%%%%%%%%%%%%%%%%%%%%%%%%
\section{Conclusions and Future Research Directions}
\label{Con}

This survey has charted the evolution of Hyperspectral Image Classification (HSC), covering traditional machine learning (TML), deep learning (DL), the latest Transformer-based architectures, and the Mamba model. Our goal was to provide a well-rounded understanding of the challenges and advancements within the HSC domain. Conventional TML approaches, while foundational, face challenges in handling high-dimensional HS data, prompting a shift towards Transformer and Mamba models with improved spectral-spatial feature handling and interoperability. Through comparative insights and experiments, we highlighted recent progress in Transformer models with attention mechanisms, self-supervised learning, and the use of diffusion models for denoising and feature extraction. The added section on Mamba-based models addresses state-of-the-art approaches, showing potential for integrating spatial-spectral information through architectures tailored to HSC. Key challenges like data scarcity, model interpretability, and real-time performance remain, and future research should focus on domain-specific advancements, enhancing model efficiency, and leveraging cross-domain transfer learning. By providing insights into readily available models, this survey aims to support future research with a comprehensive baseline, encouraging innovations that drive real-world applications in precision agriculture, environmental monitoring, and more.

%%%%%%%%%%%%%%%%%%%%%%%%%%
\bibliography{Sam}
%%%%%%%%%%%%%%%%%%%%%%%%%%
\end{document}